\documentclass[preprint,3p,times,twocolumn]{elsarticle}
\usepackage{graphicx}
\usepackage{times}  
\usepackage{helvet}  
\usepackage{courier}  
\usepackage{url}  
\usepackage{algorithm}
\usepackage{algpseudocode}
\usepackage{comment}
\usepackage{multirow}
\usepackage[table,xcdraw]{xcolor}
\usepackage{amsmath} 
\usepackage{amssymb}  
\usepackage{pifont}
\usepackage{todonotes}
\usepackage{hyperref}
\usepackage{balance}
\usepackage{svg}
\usepackage{subcaption}
\usepackage{graphicx}
\usepackage{caption}
\usepackage{subcaption}

\usepackage{pdfpages}
\usepackage{changepage}

\newcommand{\cmark}{\ding{51}}%
\newcommand{\xmark}{\ding{55}}%

\usepackage{tablefootnote}

\algnewcommand\algorithmicforeach{\textbf{for each}}
\algdef{S}[FOR]{ForEach}[1]{\algorithmicforeach\ #1\ \algorithmicdo}

\DeclareMathOperator*{\argmax}{argmax} 
\DeclareMathOperator{\atantwo}{atan2}

\journal{Robotics and Autonomous Systems}

\begin{document}

\begin{frontmatter}

\title{DWA-3D: A Reactive Planner for Robust and Efficient \\ Autonomous UAV Navigation in Confined Environments}

\author{Jorge Bes, Juan Dendarieta, Luis Riazuelo, Luis Montano.}

\address{
Instituto de Investigaci\'on en Ingenier\'ia de Arag\'on (I3A),
University of Zaragoza, Spain \\  montano@unizar.es
}

\begin{abstract}
Despite the growing impact of Unmanned Aerial Vehicles (UAVs) across various industries, most of current available solutions lack for a robust autonomous navigation system to deal with the appearance of obstacles safely. This work presents an approach to perform autonomous UAV planning and navigation in indoor or confined scenarios in which a safe and high maneuverability is required, due to the cluttered environment and the narrow rooms to move.
The system combines a RRT* global planner with a newly proposed reactive planner, DWA-3D, which is the extension of the well known \textit{DWA} method for 2D robots. We provide a theoretical-empirical method for adjusting the parameters of the objective function to optimize, easing the classical difficulty for tuning them. An onboard LiDAR provides a 3D point cloud, which is projected on an Octomap in which the planning and navigation decisions are made. There is not a prior map; the system builds and updates the map online, from the current and the past LiDAR information included in the Octomap. 
Extensive real-world experiments were conducted to validate the system and to obtain a fine tuning of the involved parameters. These experiments allowed us to provide a set of values that ensure safe operation across all the tested scenarios. Just by weighting two parameters, it is possible to prioritize either horizontal path alignment or vertical (height) tracking, resulting in enhancing  vertical or lateral avoidance, respectively. Additionally, our DWA-3D proposal is able to navigate successfully even in absence of a global planner or with one that does not consider the drone's size. Finally, the conducted experiments show that computation time with the proposed parameters is not only bounded but also remains stable around $40 ms$, regardless of the scenario complexity.

\end{abstract}


\end{frontmatter}

\section{Introduction} \label{sec:intro}

Unmanned Aerial Vehicles (UAVs) market is experiencing an outstanding growth \cite{kapustina2021global} in the last years driven by their versatility and ability to access difficult or dangerous locations. From mine inspection \cite{mining_drones_industry} and infrastructure maintenance \cite{10341871} to logistics \cite{roca2019logistic} and agriculture \cite{liu2022challenges} or even search-and-rescue missions \cite{surmann2022lessons}, UAVs are transforming industries.  While most applications thrive in open environments with neglectable collision risk, venturing into more complex areas demands robust autonomous navigation solutions due to the severe consequences that a potential crash would involve. Those needs became evident when the US Defense Advanced Research Projects Agency (DARPA) funded with $\$82$ million the \textit{DARPA Subterranean Challenge} \cite{montgomery2021pentagon}, which focused on navigation in hazardous scenarios like caves.

\begin{figure}[h!]
    \centering
    \begin{subfigure}{\columnwidth}
        \centering
        \includegraphics[width=\linewidth,keepaspectratio,trim = 2cm 2cm 2cm 2cm, clip]{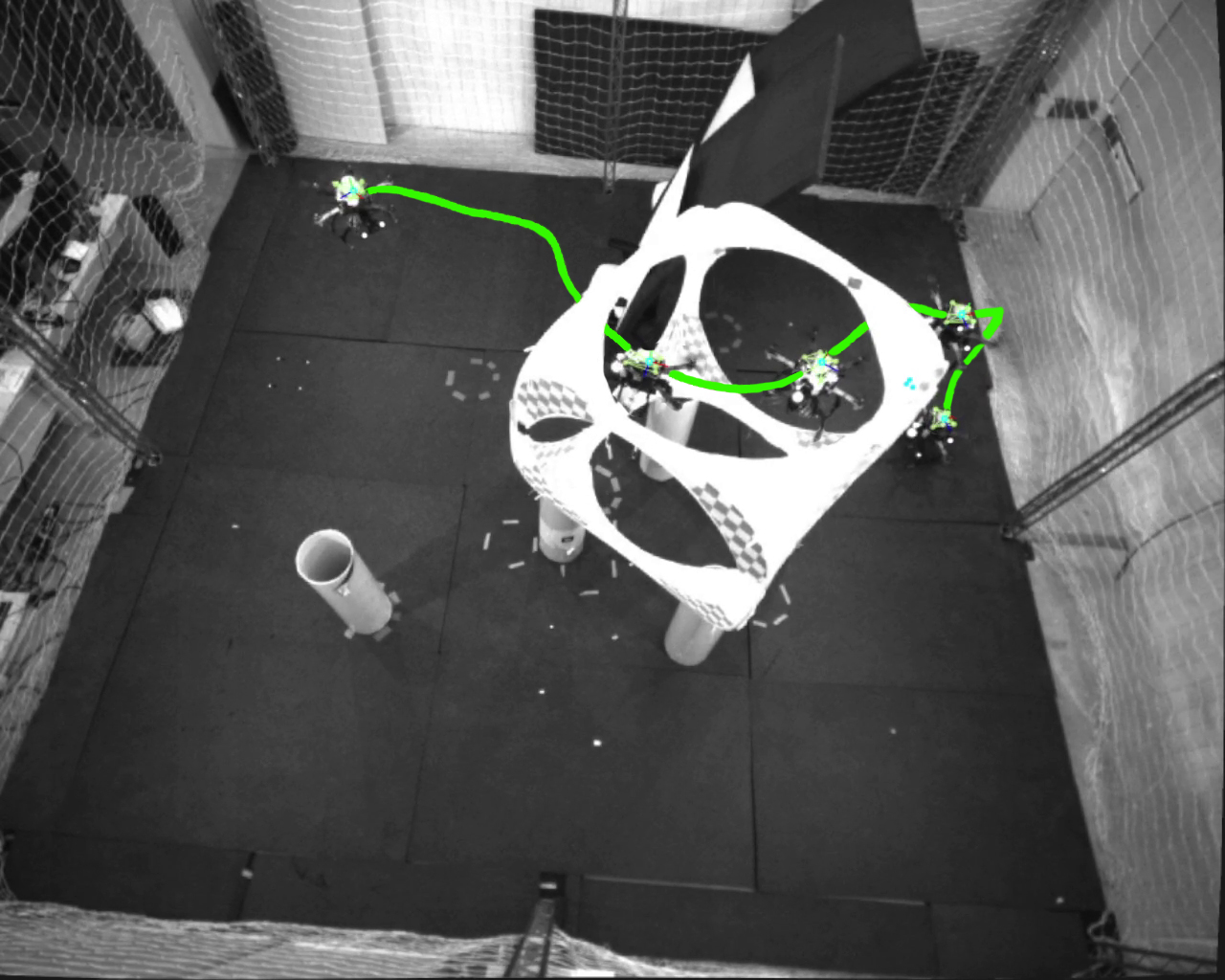}
        \caption{Maneuvers in the \textit{Rings Scenario}.}
    \end{subfigure}
    \hfill
    \begin{subfigure}{\columnwidth}
    \centering
        \includegraphics[width=\linewidth, keepaspectratio, trim = 5cm 3cm 0cm 0cm, clip ]{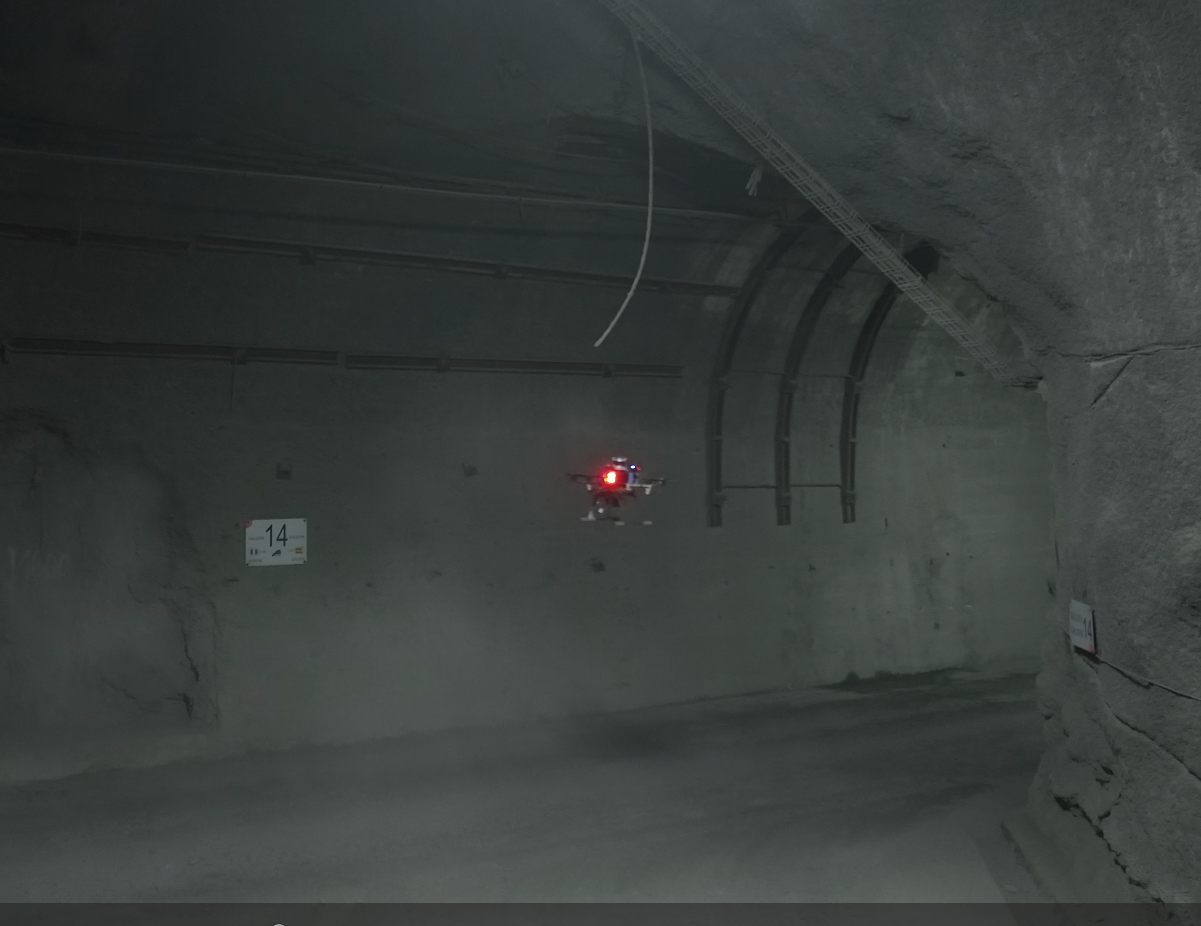}
        \caption{Inspection mission in the \textit{Somport Tunnel}, where poor light conditions, dust and changing winds are encountered.}
    \end{subfigure}
    \hfill
    \caption{Our hexarotor performing real autonomous flights. In our facilities (top), accurate localization is provided by a motion capture system. Outside our laboratory (bottom), we perform SLAM with the onboard 3D LiDAR (FLOAM \cite{FLOAM}).}
   \label{fig:Intro_image}
\end{figure}

Previous approaches to fully integrate an autonomous UAV navigation system have already been presented, like the one in \cite{perez2018architecture}, where the local planner changes the plan computed by the global when an obstacle interrupts the line of sight between waypoints. Although their results seem promising, no dynamic constraints are considered during the replanning and their real experiments lack of complexity. A NMPC perception-based reactive navigation method with real experimentation is proposed in \cite{lindqvist2021reactive} and integrated with a 3D Artificial Potential Field in \cite{lindqvist2022adaptive}, enabling also human-safe interactions. Although the 3D environment is taken into account, they do not exploit the vertical motion capabilities of the UAV during obstacle evasion, which may be mandatory in unstructured and uneven environments as caves.  
Two vision-based approaches with fast maneuver capabilities are presented in \cite{tordesillas2021panther} and \cite{tordesillas2021faster}, but due to depth cameras range limitations its application is neglected in critical low-visibility scenarios, hindering safe operation. In the latter, a JPS (Jump Point Search) is used as a global planner, and third-degree polynomial trajectories are computed by optimizing a constrained function and a triple integrator model  for moving the drone at high speed. The optimization algorithm iterates until convergence, therefore the computational time is not a priori bounded, and new subgoal recomputation is needed in these cases. Moreover, the method needs the global plan for re-computing feasible trajectories.  An obstacle avoidance method based on RMS (Riemannian Motion Policies), feasible for both volumetric maps and raw 3D LiDAR scans is shown in \cite{eth_reactive_LIDAR}, where high control rates are attained thanks to GPU raycasting operations and concurrency. It exploits the massive parallel policies combination and raycasting in a voxel-based map. As a local planner, the method also falls sometimes in local minima, because it there is not integrated with a global planner. However, complex scenarios are only tested in simulation, and only a simpler scenario for a real experiment is achieved. The work in \cite{lu2019laser} combines a collision detector with a RRT* global planner for drone navigation but both, the method  and the simulations are developed for a 2D horizontal navigation. \cite{elmokadem2021hybrid} integrates a RRT-Connect global planner with a collision detector that uses the distance to the obstacle and a sliding mode based reactive controller to avoid the closest obstacle, returning to the path planned when the obstacle is far from that distance. 
\cite{ohradzansky2022lidar} addresses a LiDAR-based drone navigation for underground tunnels, using a bio-inspired nearness detection and a reactive proportional control for the velocities and the $Yaw$ angle. The work does not consider a true  avoidance of obstacles in the way, only the walls of the tunnel. \cite{dirckx2023optimal} combines an optimal control solved as a nonlinear problem for obstacle avoidance with a MPC and a low-level PID controller for navigating in environments with many obstacles. The paper neither provides the computation time to evaluate the real-time performance to be applied in real experiments, nor drone trajectories among the obstacles.
These methods have been evaluated in simulation and considering the map and the obstacle location are completely known beforehand. We compare our approach and the previous ones regarding the main features needed for safe autonomous navigation in the potential target environments in \autoref{tab:SOTA_Comp}. In \autoref{sec:ComparacionFASTER}, we compare our approach with \cite{tordesillas2021faster}, the work that we considered the most promising one given its results.

\begin{table*}[h!]
    \centering
    \begin{tabular}{|*{7}{c|}}
        \hline
        \multirow{2}{1.8em}{\textbf{Ref.}} & \multirow{2}{1.75em}{\textbf{Code}} & \textbf{Previous} & \multirow{2}{4.9em}{\textbf{Real-world}} & \multirow{2}{5em}{\textbf{Perception}} & \multirow{2}{5.5em}{\textbf{Localization}} & \textbf{Confined}  \\
        & & \textbf{Map} &  & &  & \textbf{Scenarios} \\
        \hline
        \cite{perez2018architecture} & \xmark & \cmark & Only Lab & RBG-D camera & MonteCarlo & \xmark  \\
        \hline
        \cite{lindqvist2021reactive} & \xmark & \xmark & Only Lab & 2D LiDAR & Motion Capture & \xmark \\
        \hline
        \cite{lindqvist2022adaptive} & \xmark & \xmark & \cmark & 3D LiDAR & SLAM\hyperref[foot:VIO]{$^1$}  & \cmark \\
        \hline
        \cite{tordesillas2021panther} & \cmark & \xmark & \cmark & RGB-D camera & VIO\hyperref[foot:VIO]{$^2$} & \xmark \\
        \hline
        \cite{tordesillas2021faster} & \cmark & \xmark & Only Lab & RGB-D camera & Motion Capture & \xmark   \\
        \hline
        \cite{eth_reactive_LIDAR} & \cmark & \xmark & Only Lab & 3D LiDAR & \xmark & \xmark \\
        \hline
        \cite{lu2019laser} & \xmark & \xmark & Simulation & 2D LiDAR &  Simulated GT & \cmark  \\
        \hline
        \cite{elmokadem2021hybrid} & \xmark & \xmark & Simulation & \xmark & Simulated GT & \xmark  \\
        \hline
        \cite{ohradzansky2022lidar} & \xmark & \xmark & Simulation & Unreal 3D LiDAR &\xmark & \cmark  \\
        \hline
        \cite{dirckx2023optimal} & \cmark & \xmark & Simulation  & \xmark &  Simulated GT & \xmark   \\
        \hline
        \textbf{Ours} & \cmark & \xmark & \cmark & 3D LiDAR & SLAM & \cmark  \\
        \hline
    \end{tabular}
    \caption{Qualitative comparison with previous approaches. SLAM$^1$: Simultaneous Localization And Mapping. VIO\label{foot:VIO}$^2$: Visual Inertial Odometry.}
    \label{tab:SOTA_Comp}
\end{table*}

The Dynamic Window Approach (DWA) \cite{dwa_1997} is a well known solution to the 2D reactive navigation problem that has been widely used since it was proposed in 1997. Despite being a reliable and reference method, it still has not been proposed to extend it to the 3D search space for UAVs, as far as the authors are concerned. Two proposals for submarine vehicles are presented in \cite{DWA_submarine} and \cite{DWA_submarine2} where only simulated experiments are performed. In \cite{MULTIUAV_DWA_VFH} the original 2D DWA is combined with 3DVFH+ to perform multi-UAV obstacle avoidance. Nevertheless, their DWA is limited to the 2D plane, computing the velocity in Z axis out of the 2D velocity optimization achieved by DWA. The work lacks of real experimentation, being the simulation scenarios with obstacles very simple.  Moreover, the authors do not provide hints about how to tune the values of the multiple parameters the method includes.

Our work focuses on the development of a new local planner, DWA-3D, for scenarios in which a drone has to maneuver among nearby obstacles distributed in the environment. The local planner is integrated  with a state of the art global planner, RRT* \cite{RRT*}. Planning and navigation make decisions on an occupancy map (Octomap \cite{octomap}) built from the environment real time information provided by an onboard 3D LiDAR. Real-world experiments have been performed with a custom hexarotor (see \autoref{fig:Intro_image}) equipped with a LiDAR Ouster OS032. First, the proposal has been fine-tuned and evaluated in an indoor arena, where a motion capture system gives an accurate estimation of the drone pose. Later, more complex environments where no ground-truth localization was available have been faced successfully. 

The main contributions are:

\begin{itemize}
    \item We have developed a novel technique, DWA-3D, solved as an optimization in a  velocity space. The dynamic constraints, in terms of capability of acceleration-deceleration, are implicit in the method, therefore only feasible commands are computed every control period. The drone can maneuver in scenarios with lateral, top or bottom obstacles, making the best (optimal for the objective function) avoidance maneuver depending on two decision parameters.

    \item Unlike in the classical 2D method, a non uniform distance computation to obstacles has been proposed, to enhance them as a function of the relative position and orientation with respect to the drone forward motion. The dense point cloud LiDAR information is condensed in a voxel-map representation, which allows to reduce strongly the computation time in the optimization, without loosing relevant obstacle information. Moreover, this approach allows the reactive planner to consider the past observed information to make decisions, instead of directly using the point cloud in the current sampling period for a pure reactive behavior.
   
    \item An ample set of exhaustive real-world experiments in different kind of scenarios have been performed, which allowed to qualitatively evaluate the method. From this evaluation,  we provide a set of default parameters, whose constraints have been analytically established.

\end{itemize}

       The computational time is around 40ms despite the scenario complexity, allowing to perform real-time control of the UAV and react to unexpected obstacles in the path. During real-world experiments, the proposed DWA-3D reactive planner was integrated alongside three global planner variants to address the challenges posed by frequent recomputations in partially observed or dynamic environments. These variants included a naive planner, a drone-size-agnostic planner, and a safer drone-size-aware planner. The capability of DWA-3D local planner of making the last and safer decisions about the best velocity command to apply, has been evaluated in the three cases, concluding its robustness in all the scenarios evaluated. DWA-3D computes commands to follow or not the global planned path, according with the new information incoming from the sensors  every control period and the feasibility and safety of the resulting motion estimated in the optimization. The architecture has been implemented in ROS \cite{ros}. The authors commit to release a well-tested and configurable code both for simulation and real experimentation once the publication is accepted.

\section{System overview} \label{sec:overview}


The proposed UAV architecture relies on 3D pointclouds (from a 3D LiDAR or a depth camera) to build a map, enabling fast and efficient queries about occupancy or distance to obstacles (Octomap \cite{octomap} was chosen for this purpose).

Initial experimental verification of the architecture was conducted in a controlled environment equipped with a motion capture system, providing precise drone localization. 
This allows to decouple the navigation problem, the main objective of this work, from the localization one. Once that safe autonomous navigation was guaranteed, the system was deployed in real-world scenarios, relying on a LiDAR based SLAM method (F-LOAM \cite{FLOAM}) self-localization. 

\autoref{fig:SystemOverview} presents the integrated navigation system and its components.
\begin{figure}[h!]
    \centering
    \includegraphics[width=1\columnwidth,keepaspectratio]{img/System_overview/GlobalScheme_V8.pdf}
    \caption{System General Scheme.}
    \label{fig:SystemOverview}
\end{figure}

First, path planning (also known as global planning) computes a set of waypoints that guide the robot to the target location. As the most computationally expensive component it operates at the lowest rate. Furthermore, its results remains valid for a longer period of time than the others. 

Then, the local planner is in charge of aligning the UAV with the global path, ensuring not only that the desired route is followed as precisely as possible, but also that the dangers and the sudden changes along it are avoided in a reactive way. One of the key objectives of the work is to evaluate the local planner's capabilities to make optimal motion decisions at each sampling time, even with a very simple global planner or a low-frequency one, or in scenarios where the global path might intersect with previously unknown or dynamic obstacles. This part of the system must work in real time and will compute the actions needed to perform the trajectory. Thinking in many of the drone missions in which a continuous frontal observation of the environment is desired, the method will enforce forward motion and avoid pure lateral movements. Therefore, only motions in the robocentric reference $X$, and $Z$ axes, and the rotation around $Z$ axis (Yaw) are allowed as control actions.

The low-level onboard control obeys the linear and angular velocity commands sent by the reactive navigator, moving the drone and allowing the sensors to perceive a new fragment of the world and update the map.

System modularity, flexibility and accessibility are key requirements, minimizing dependencies between components while enhancing their substitution if needed. Thus, the architecture has been implemented in ROS, using standard messages types for inter-component communication.

\section{Localization}

The system has initially been tested in controlled scenarios within an indoor arena equipped with a motion capture system (Optitrack \cite{Optitrack}). Having access to this precise drone localization in real time allows us to decouple the localization and navigation problems and focus on the latter. 

\begin{figure}[h]
    \centering
    \begin{subfigure}{\columnwidth}
    \centering
        \includegraphics[width=0.9\linewidth, trim= 0cm 0cm 0cm 4.75cm, clip]{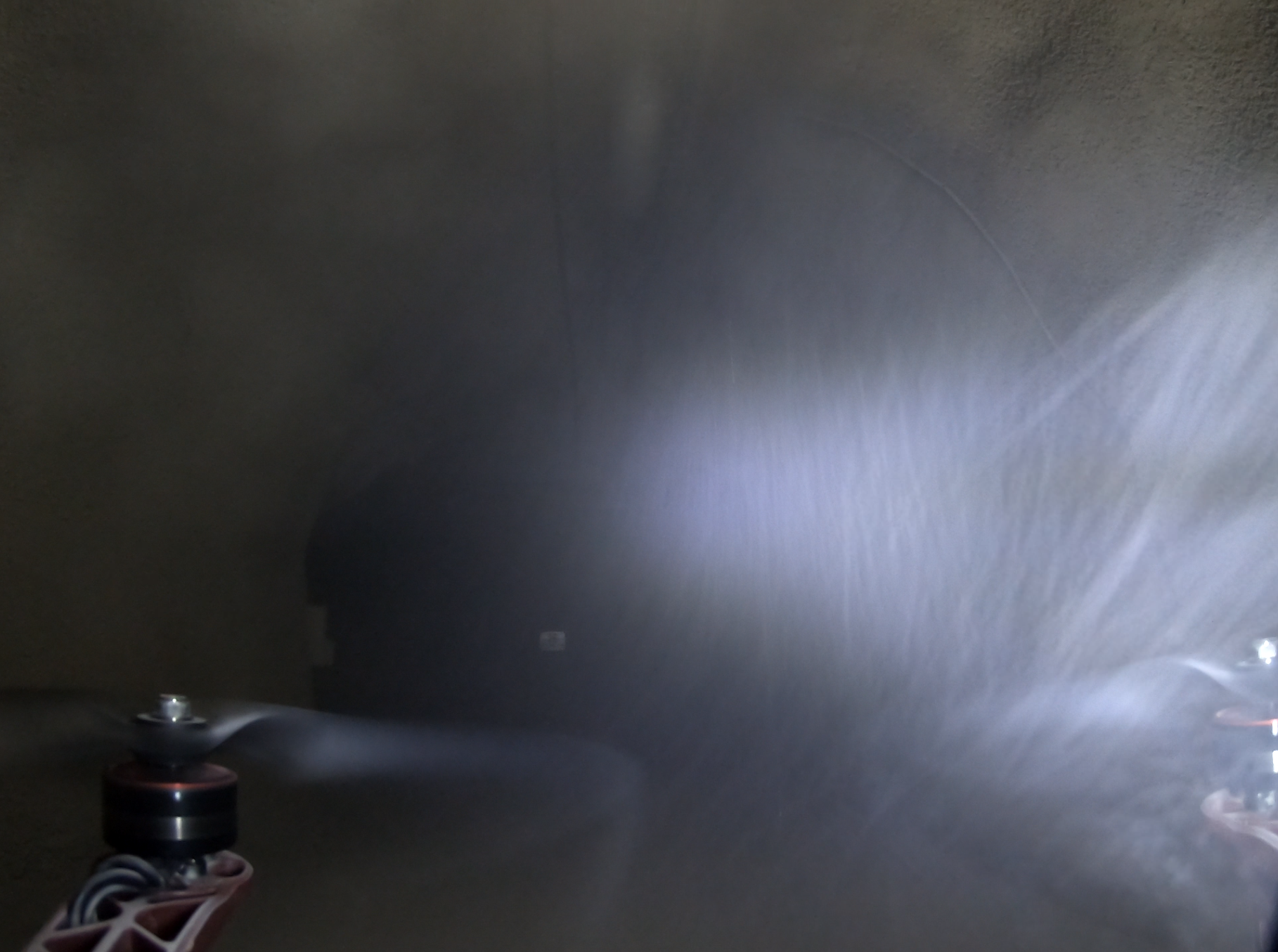}
        \caption{Drone's point of view inside a dust cloud while flying in a dark tunnel.}
    \end{subfigure}
    \hfill
    \begin{subfigure}{\columnwidth}
    \centering
        \includegraphics[width=0.9\linewidth,trim = 0cm 3.5cm 0cm 6.25cm, clip ]{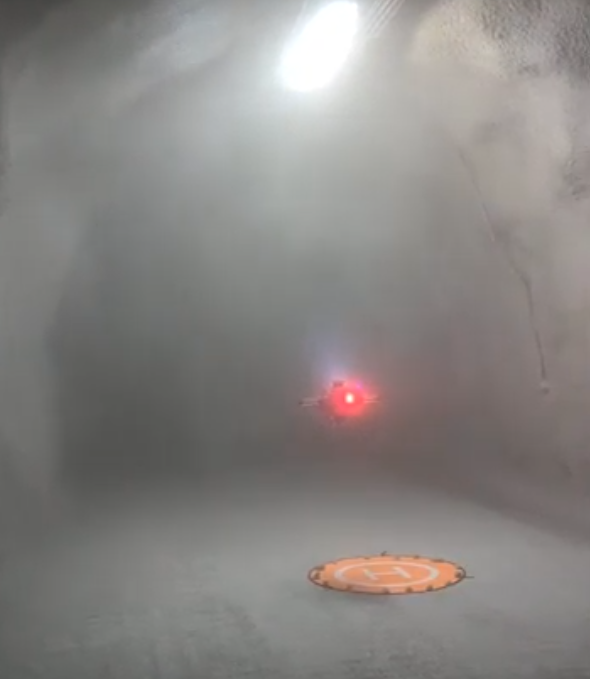}
        \caption{UAV flying through a dense dust cloud in a tunnel with poor visibility.}
    \end{subfigure}
    \hfill
    \caption{Potential scenarios in which the drone could be deployed. The adverse visibility conditions would make a visual-based localization method fail. Despite the dense dust and lack of light, F-LOAM (\cite{FLOAM}) was able to localize the UAV during this real flight.}
    \label{fig:Visibilidad_mala}
\end{figure}

Outside the arena another localization solution is needed. Given that 3D pointclouds are available, a LiDAR Odometry and Mapping method, F-LOAM (\cite{FLOAM}), has been chosen. This decision has been reinforced by the poor visibility conditions for a vision-based localization system that the drone would face in some of the scenarios in which it could be deployed, \autoref{fig:Visibilidad_mala}.  

\begin{figure}[h]
    \centering
    \begin{subfigure}{\columnwidth}
        \centering
        \includegraphics[width = 0.9\linewidth, keepaspectratio]{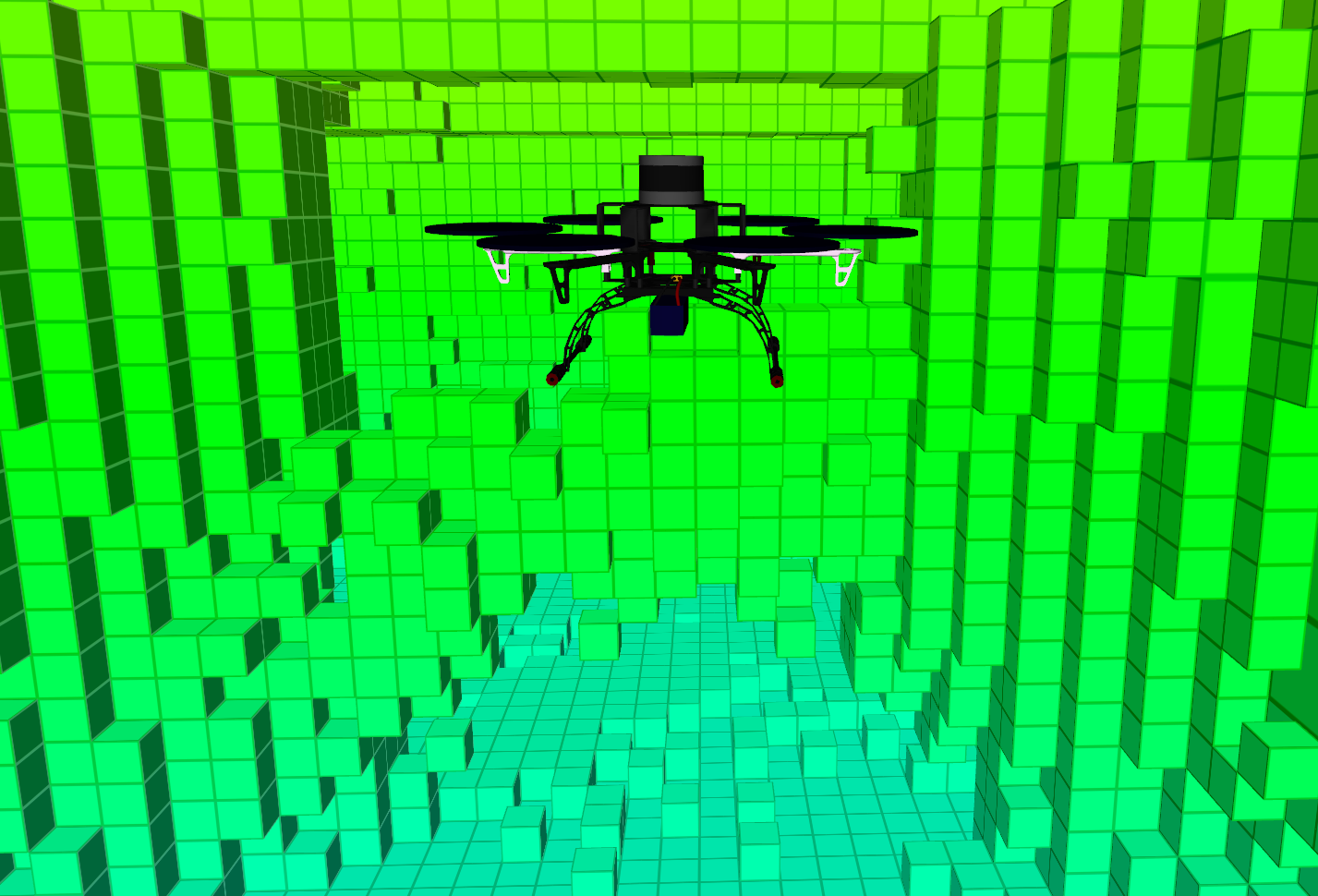}
        \caption{Octomap view with the blocked door.}
    \end{subfigure}
    \begin{subfigure}{\columnwidth}
        \centering
        \includegraphics[width = 0.9\linewidth, trim=0cm 3.5cm 0cm 0cm , clip,keepaspectratio]{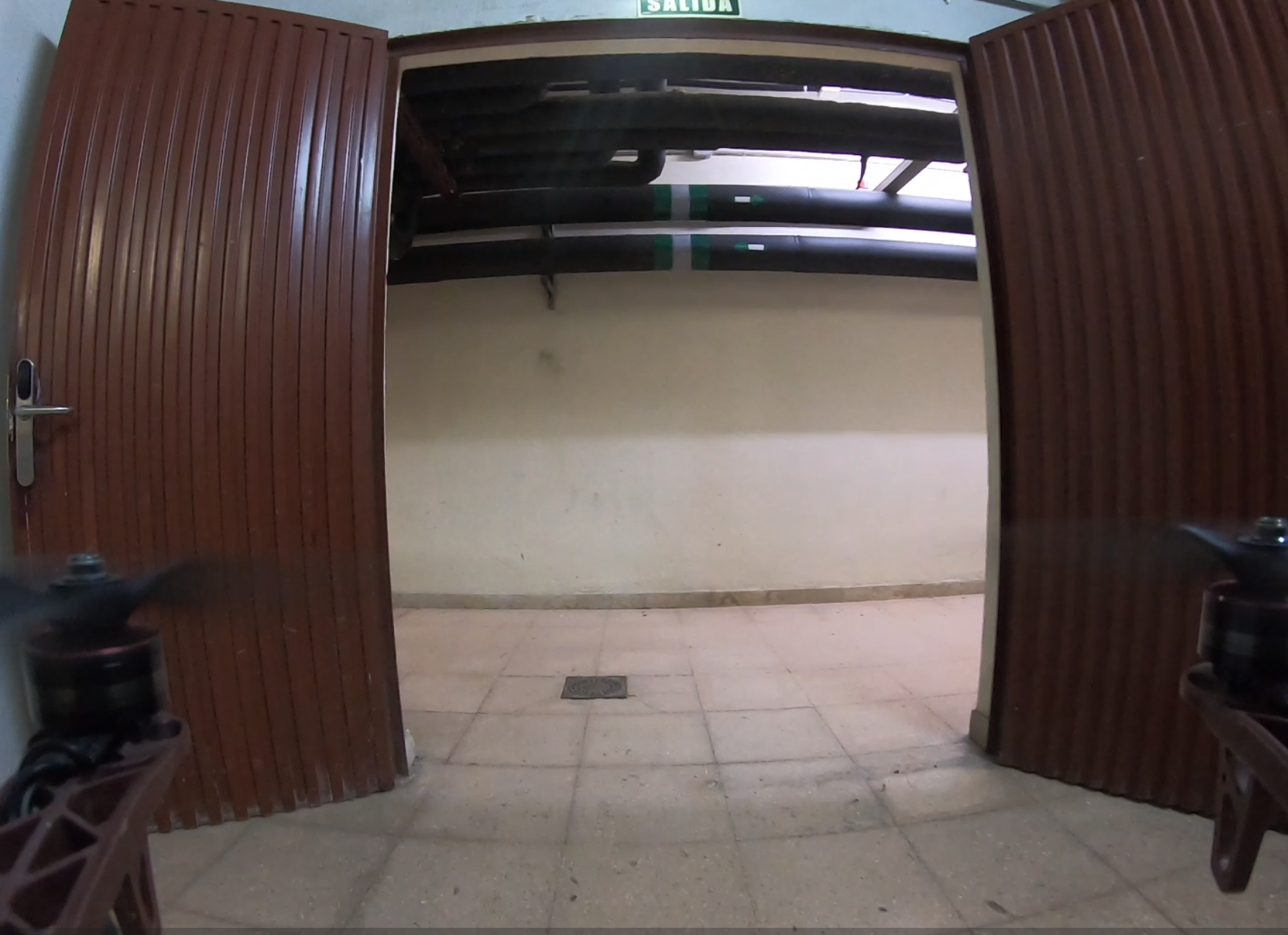}
        \caption{Drone's point of view when the door becomes blocked in the Octomap.}
    \end{subfigure}
    \hfill
    \caption{Attitude estimation drift situation in which a door gets blocked in the Octomap.}
    \label{fig:BlockedDoor}
\end{figure}

During F-LOAM integration within the architecture, considerable attitude estimation drift was observed. Unlike in ground robot applications, where such drift primarily affects the map quality, for a UAV, inaccurate attitude can have severe consequences, including collisions with the ceiling or floor. Additionally, this $Z$-coordinate error led to situations in which the drone got blocked by querying about surroundings occupancy in wrong locations and updating incorrectly certain Octomap regions. In \autoref{fig:BlockedDoor} the UAV got trapped because F-LOAM estimated that it was at $2m$ instead of $1m$ over the floor, so DWA-3D was checking if the door's upper frame was occupied instead of the door gap. In addition to that, the floor section that the 3-D LiDAR was perceiving was wrongly added to the Octomap's region corresponding to the door gap, blocking the only way that the UAV had to keep advancing. 

We tackled and solved this issue with a downward-facing 1D-LiDAR range sensor. The main idea is fusing both the attitude estimation coming from F-LOAM and the 1D-LiDAR measurement. Rather than directly fusing absolute values, the proposed solution fuses at each iteration $k$ the variations of both F-LOAM attitude estimation ($z^{k}_{FLOAM}$) and the 1D LiDAR measurement ($z^{k}_{1D}$) with respect to their previous iteration,
\begin{equation}
    \hat{z}^{k} = \hat{z}^{k-1} + w (z^{k}_{FLOAM} - z^{k-1}_{FLOAM}) + (1 -w) (z^{k}_{1D} - z^{k-1}_{1D})
\end{equation} 
where $w$ is a fusion weighting parameter. Additionally, the fused value is also fed to F-LOAM so its attitude drift gets bounded in the next estimation. This solution enabled reducing considerably $Z$-coordinate drift. A further analysis of the error committed with F-LOAM is performed in \autoref{sec:ErrorLocalization}.
\section{Map Representation}
\begin{figure*}
    \centering
    \hfill
    \begin{subfigure}{0.3\textwidth}
        \centering
        \includegraphics[height= 4.5cm, trim=25cm 0cm 0cm 0cm, clip,keepaspectratio]{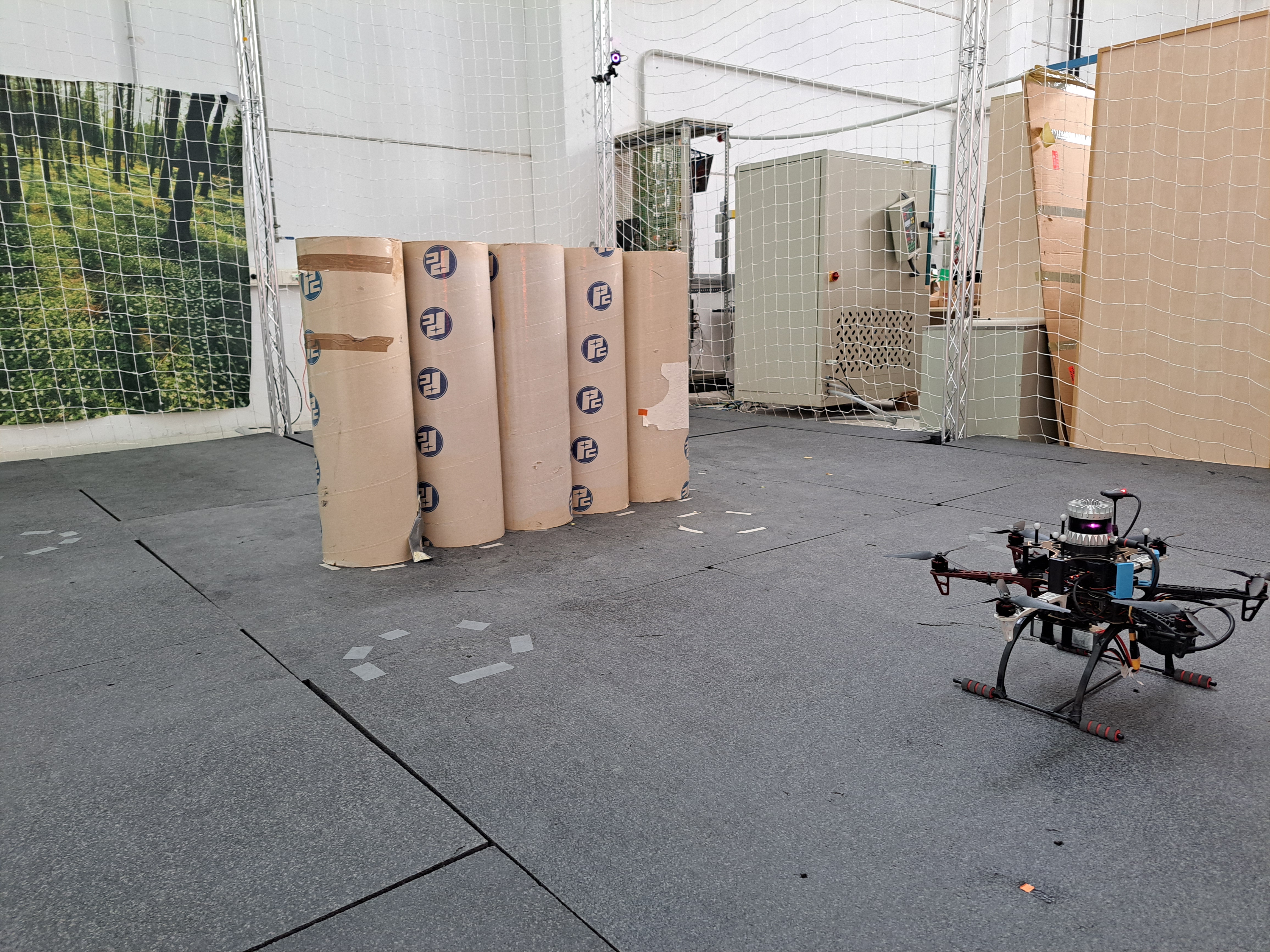}
        \caption{Real scenario image.}
        \label{fig:RealWorldExample}
    \end{subfigure}
    \hfill
    \begin{subfigure}{0.32\textwidth}
        \centering
        \includegraphics[height= 4.5cm, trim=5cm 0cm 10cm 0cm , clip,keepaspectratio]{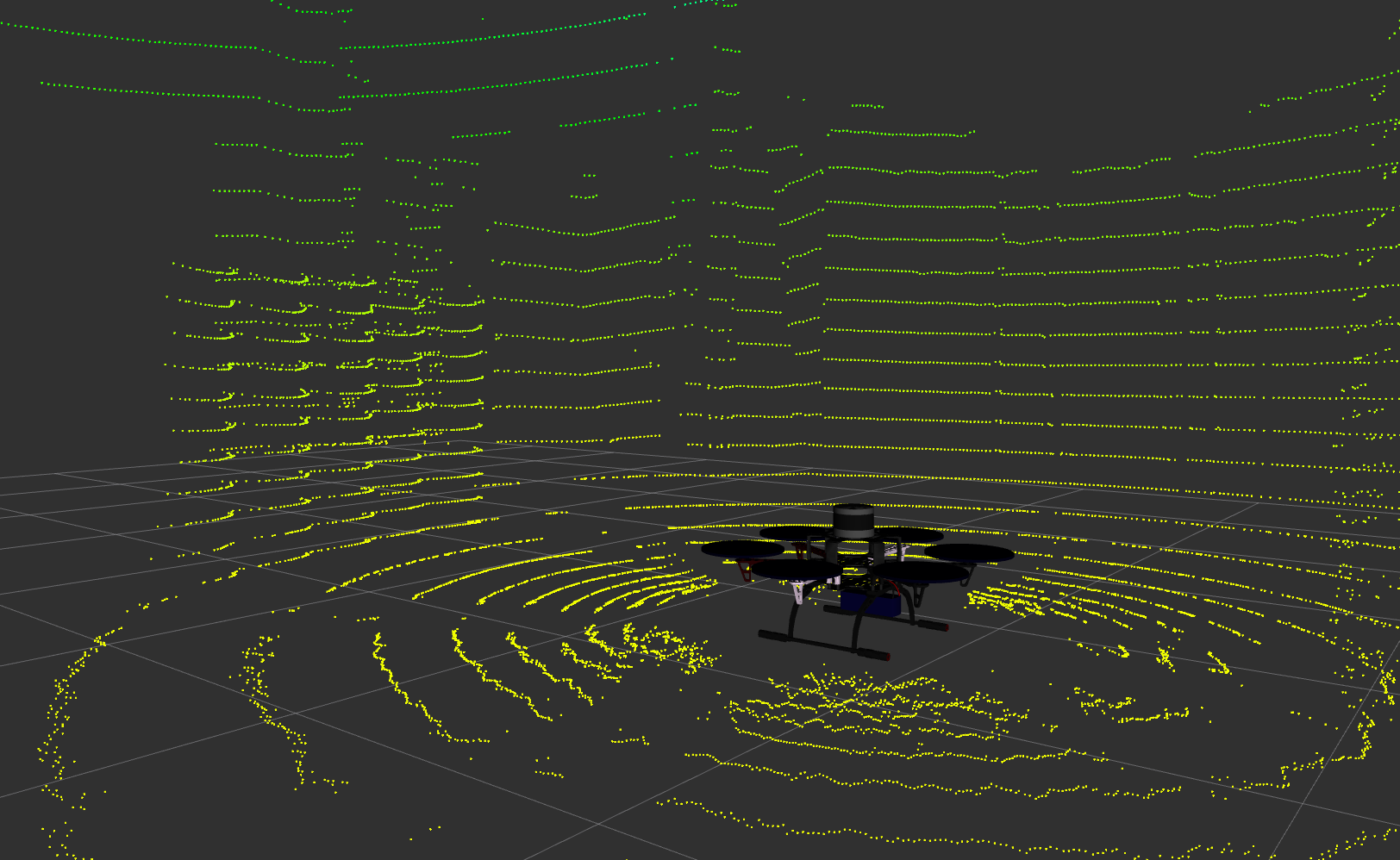}
        \caption{Captured Point Cloud.}
        \label{fig:PointCLoudExample}
    \end{subfigure}
    \hfill
    \begin{subfigure}{0.3\textwidth}
        \centering
        \includegraphics[height= 4.5cm,trim=10cm 0cm 10cm 0cm, clip, keepaspectratio]{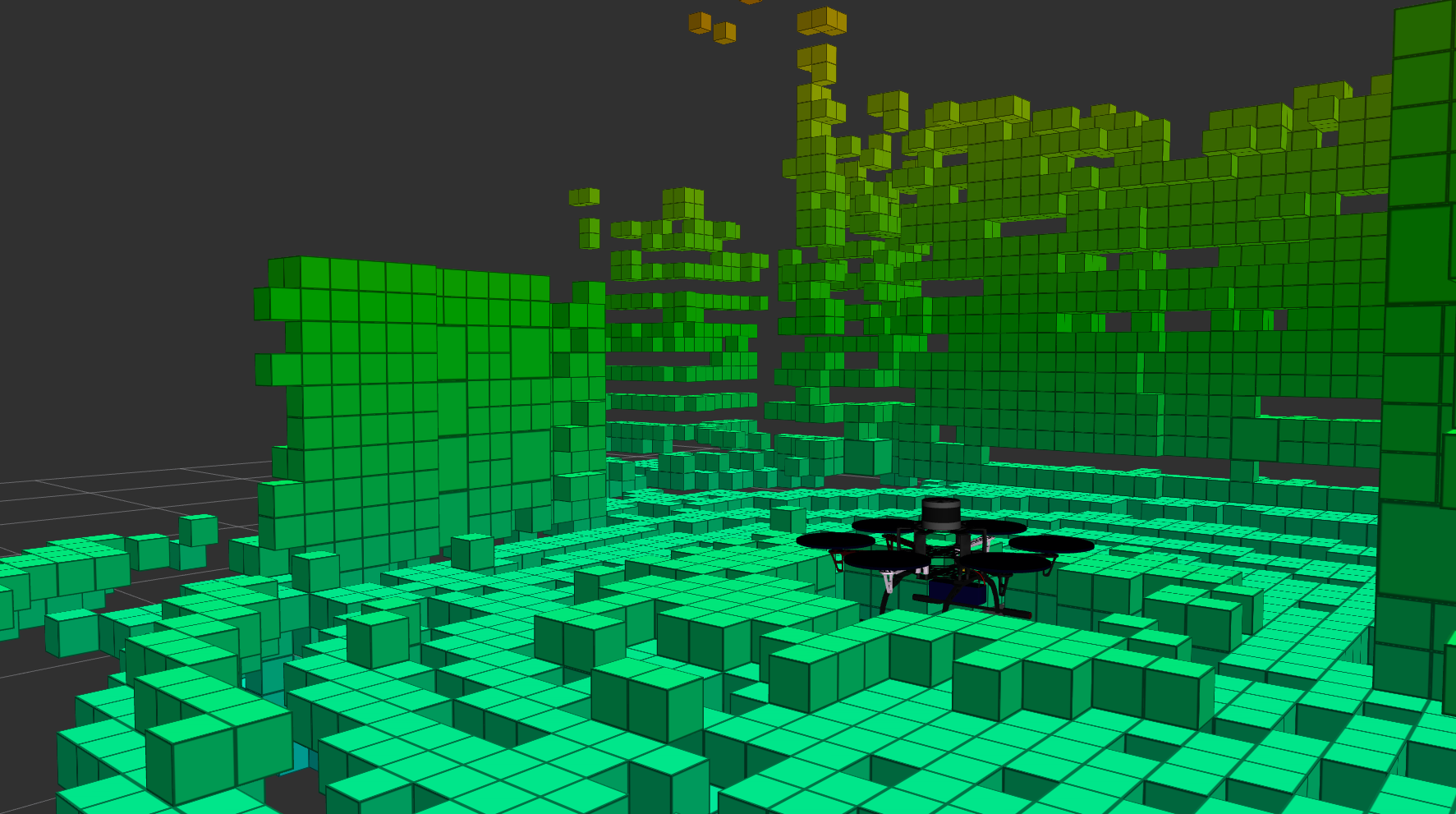}
        \caption{Octomap generated from the Point Cloud.}
        \label{fig:OctomapExample}
    \end{subfigure}
    \hfill
    \caption{Steps from the real world to the Octomap environment representation.}
    \label{fig:Map_example}
\end{figure*}

The environment around the robot must be stored in order to use it both in global and local planning. As 3D LiDAR pointclouds tend to be too dense, it is advisable to compress the information before giving it to both planners. The selected representation system has been Octomap \cite{octomap}, a 3D occupancy grid map that can be built online ($7 Hz$). Moreover, it updates already known areas if there is any change, allowing reactive navigation even with moving obstacles. \autoref{fig:Map_example} shows an example of the transition from the real world environment to the captured pointcloud and the resulting Octomap representation. A voxel size of $10 cm$ has been chosen, balancing computation load and information loss; reducing the map density from 32768 points per each LiDAR scan to just 1000 voxels per $m^3$. Furthermore, this data structure allows for efficient occupancy queries along the space. 

\section{DWA-3D (Reactive Planner)}
\label{sec:DWA}
One of our main proposals is recovering the basic ideas of the Dynamic Window Approach (DWA) and extending it to a UAV for 3D navigation, by controlling advance and ascent speeds along with the rotation velocity in the horizontal plane. Additionally, the avoiding behavior can be configured, being able to selected between vertical or horizontal preference.
This section details the main features of DWA-3D, the Dynamic Window Approach extended to 3D space: the search space with the velocities available and the evaluation function used to select the optimal combination of velocities.

\subsection{Problem formulation}
Given an UAV that must follow a trajectory $\mathcal{P}$ composed by a set of $n$ waypoints, let $\textbf{x}$ be its position in the $3D$ space and $\textbf{x}_{g_i}$ the position of the current subgoal that is being tracked, both represented in the world reference $W$.  
\begin{equation}
    \ \mathcal{P} = \{ \mathbf{g_1}, ..., \mathbf{g_n}\}
\end{equation}
\begin{equation}
    \ \textbf{x} = 
    \begin{bmatrix}
        x, y, z, \phi, \theta,\psi \
    \end{bmatrix}    
\end{equation}
\begin{equation}
    \ \textbf{x}_{g_i} = 
    \begin{bmatrix}
        x_{g_i}, y_{g_i}, z_{g_i} \ 
    \end{bmatrix} 
\end{equation}
with $\phi$, $\theta$, $\psi$ the drone's Roll, Pitch, Yaw angles and . The velocity vector of the drone in the dronecentric $3D$ space, $\dot{\textbf{x}}$, can be expressed as
\begin{equation}
    \ \dot{\textbf{x}} = \begin{bmatrix}
        v_x, v_y, v_z, \omega_{x}, \omega_{y},\omega_{z} \
    \end{bmatrix}
\end{equation}
Although in our setup a 3D LiDAR is available, this is not the case for most of the aerial robots, which usually count with RGB or RGBD cameras pointing forward with a limited field of view. Thus, the navigation system takes only into account $v_x$, $v_z$ and $\omega_{z}$ to perform the control, neglecting $v_y$, $\omega_x$ and $\omega_y$, therefore avoiding to enforce lateral motions. Then, the control vector $\textbf{v}$ in the dronecentric reference can be defined as
\begin{equation}
    \ \textbf{v} = 
    \begin{bmatrix}
        v_x, v_z, \omega_{z} \
    \end{bmatrix}    
\end{equation}
An action $\textbf{v}$ is selected at each iteration, being $T$ the control period. Note that in the following subsections, the tracked waypoint will be expressed as $\mathbf{g}$ instead of $\mathbf{g_i}$ for the shake of readability.
\subsection{Search space}
The $v_x$, $v_z$ and $\omega_z$ combinations must be selected between a discrete set of candidates, known as \textit{Velocities Search Space}, \autoref{fig:DWA_example}. This search space is limited and includes only the velocities that fulfill the drone's dynamics and kinematics constraints while ensuring its physical integrity, given the current robot status and the surroundings information. 

\subsubsection{Maximum velocities space $V_s$}

First of all, the search space should be restricted by the UAV's minimum and maximum velocities. These limits can be imposed by hardware capabilities or safety considerations according with the environment density. Additionally, $v_x^{min} = 0$ is established to avoid backward movements, considering the desired forward-facing orientation of drone sensors. This search space, $V_s$, is known as \textit{maximum velocities space}.

\begin{multline}
    V_s = \{\textbf{v} | v_x \in [0, v_{x}^{max}] \\\land v_z \in [-v_{z}^{max}, v_{z}^{max}] \\\land \omega_z \in [-\omega_{z}^{max}, \omega_{z}^{max}]\}    
\end{multline}

\subsubsection{Dynamic Window $V_d$}
Then, the search space is further refined by considering the UAV's current velocity $\textbf{v}_c = \left[ v_{c_x}, v_{c_z}, \omega_{c_z} \right]$ and its maximum linear and angular accelerations ($\dot{v}_{x}^{max}$, $\dot{v}_{z}^{max}$ and $\dot{\omega}^{max}$). This way, only reachable velocities from the actual situation after a prediction time-step $\Delta t$ ($\Delta t \neq T$) are taken into account during the objective function optimization process. These velocities compose the \textit{Dynamic window} $V_d$, which is crucial for optimizing the objective function as it represents realistic options during the next control cycles.  

\begin{multline}
         V_d = \{\textbf{v}| v_x \in [v_{c_x} - \dot{v}_x^{max} \cdot \Delta t, v_{c_x} + \dot{v}_x^{max} \cdot \Delta t] \\\land v_z \in [v_{c_z} - \dot{v}_z^{max} \cdot \Delta t, v_{c_z} + \dot{v}_z^{max} \cdot \Delta t] \\\land \omega_{z} \in [\omega_{c_z} - \dot{\omega}_z^{max} \cdot \Delta t, \omega_{c_z} + \dot{\omega}_z^{max} \cdot \Delta t]\}
\end{multline}

\subsubsection{Admissible velocities space $V_a$}
Finally, the remaining candidate velocities are evaluated to determine their collision risk. This evaluation involves predicting the drone's future positions for each candidate velocity $\textbf{v}$ applied during a prediction time-step $\Delta t$. Any of those virtual locations where the distance to the closest obstacle falls below the drone's braking distance is discarded due to collision risk. The remaining collision-free velocities form the \textit{admissible velocities space} $V_a$,

\begin{equation}
    V_a = \{(v_x, v_z)| {v}_{xz} \leq \sqrt{2 \cdot d_{col} \cdot a_{max}}\}
\end{equation}

where $v_{xz}=\sqrt{v^2_x + v^2_z}$, $d_{col}$ is the euclidean distance from the predicted drone position to its closest obstacle and $a_{max}$ is the maximum deceleration. The straight distance $d_{col}$ instead of the curved one is used as a good approach, since the time-step is very short. Furthermore, it will always be smaller, thus it lowers the speed upper bound, which results in a safer navigation.

\begin{figure}[h]
        \centering
        \includegraphics[width=\columnwidth,keepaspectratio]{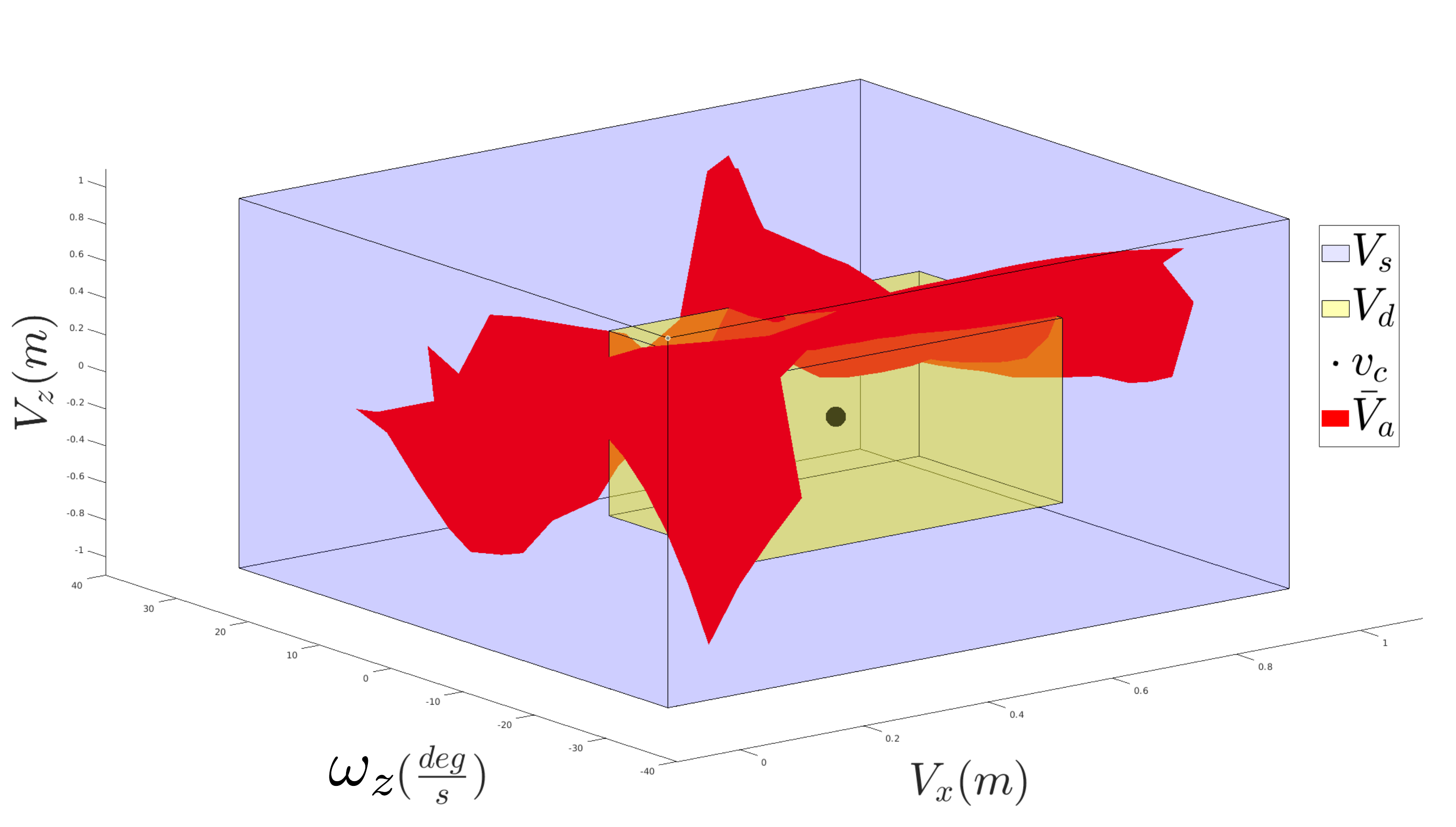}
    \caption{\textit{Search Velocities Space} $V_r = V_s \cap V_a \cap V_d$. The blue volume contains the \textit{Maximum velocities} ($V_s$), the yellow region are the reachable velocities (\textit{Dynamic Window}, $V_d$) given the current ones (black dot, $v_c$), and in red the non-admissible ones ($\bar{V}_a$).}
    \label{fig:DWA_example}
\end{figure}

The final search space where the optimal velocity is searched in the intersection between the three spaces: $V_r = V_s \cap V_a \cap V_d$ (\autoref{fig:DWA_example}). Note that, for efficiency, although the $V_a$ space is represented in the figure for the entire $V_s$ space, at each control time it only needs to be computed within the $V_d$ space, the set of velocities reachable from the current situation.

\subsection{Objective function \ G(\textbf{v})}
Once the discretized velocity search space $V_r$ has been determined, the algorithm seeks the optimal velocity combination $\mathbf{v^*}$ within it. This optimal velocity is chosen by maximizing the following objective function: 

\begin{subequations}
    \begin{equation}
    \label{eq:G_compressed}
    \ G(\textbf{v}) = \alpha \cdot Heading(\textbf{v}) + \beta \cdot Dist(\textbf{v}) + \gamma \cdot Vel(\textbf{v})
    \end{equation}
    \begin{equation}
    \label{eq:v*}
    \ \mathbf{v}^*=\argmax_{\textbf{v}\in V_r}{G(\textbf{v})}
\end{equation}
\end{subequations}

where $Heading$ favours the motion towards the goal, $Dist$ tries to maximize the distance to obstacles, and $Vel$ the high velocities. The $Heading$ term can be expanded into two terms, $Head_z$ and $Head_{\psi}$, splitting the progress towards the goal in alignment in height and in orientation in the horizontal plane. Then, \autoref{eq:G_compressed} becomes
\begin{multline}
\label{eq:G_expanded}
    G(\textbf{v}) = \alpha \cdot (K_{\psi} \cdot Head_{\psi}(\textbf{v}) + K_z \cdot Head_z(\textbf{v})) \\ + \beta \cdot Dist(\textbf{v}) + \gamma \cdot Vel(\textbf{v})
\end{multline} 

This way, it is possible to tune whether staying aligned in the horizontal plane with the objective or at the same height is prioritized. As a direct consequence, if an obstacle appears between a waypoint and the UAV, it will tend to avoid it laterally if $K_z > K_{\psi}$ because $Head_z$ value will be weighted stronger, enforcing the drone to keep the height of the goal. On the other hand, if $K_{\psi} > K_z$, vertical avoidance will tend to be chosen, as it will mean being alligned with the objective, thus keeping a greater value of $Head_{\psi}$. Each term and the weights in \autoref{eq:G_expanded} are normalized between 0 and 1, and therefore the function $G(\textbf{v})$; enabling intuitive configuration of the reactive planner. Thus, the following constraints must be fulfilled, 

\begin{equation}
    \label{eq:W_contraint}
    \begin{cases}
    \alpha + \beta + \gamma = 1 \\
     K_{\psi} + K_z =1
    \end{cases}
\end{equation}

\begin{figure}
    \centering
    \includegraphics[width=0.8\columnwidth,keepaspectratio]{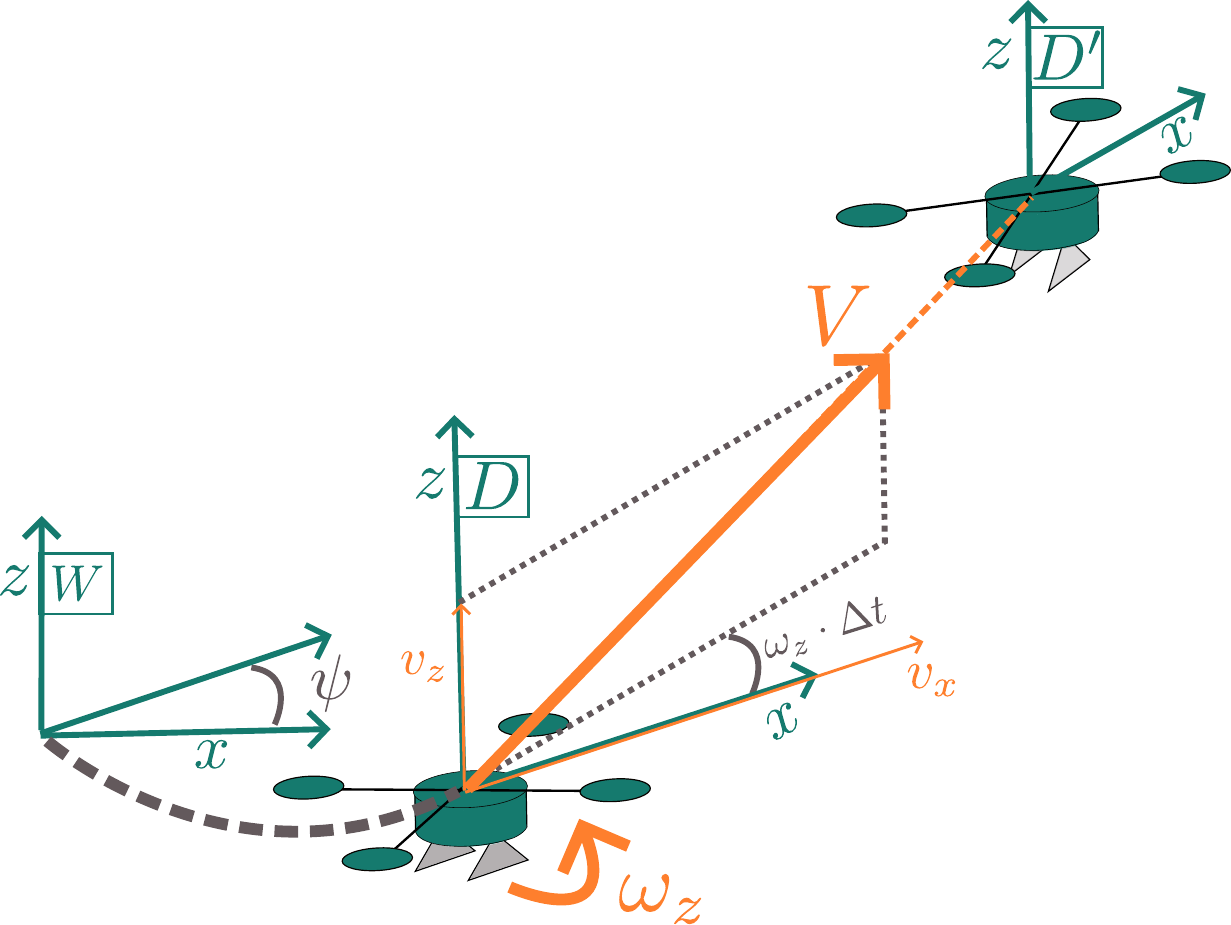}
    \caption{Predicted state of the drone after applying the selected velocities $(v_x, v_z, \omega_z)$.}
    \label{fig:trajectory}
\end{figure}

These functions work with the predicted position of the drone $\mathbf{x}'=(x',y',z')$ after applying $\mathbf{v}$ during a prediction time-step $\Delta t$ (\autoref{fig:trajectory}). As this prediction must be computed for each set of velocities in $V_r$, it is mandatory to speed up the computation as much as possible while preserving its accuracy, so neither accelerations nor the effect of roll and pitch angles are taken into account. Also, the linear and angular motions are approximated as uniform ones every sampling time $\Delta t$,

\begin{equation}
\label{eq:PosePredicted}
    \begin{cases}
    \psi' = \psi + \omega_z \cdot \Delta t\\
    x' = x + v_x \cdot \Delta t \cdot \cos{\psi'}\\
    y' = y + v_x \cdot \Delta t \cdot \sin{\psi'}\\
    z' = z + v_z \cdot \Delta t
\end{cases}
\end{equation}

being $x, y, z$ and $\psi$ the current values of the position and orientation of the drone pose and $x', y', z'$ and $\psi'$ their predicted values respectively.

The objective function $G(\textbf{v})$ is discretized for computing its maximum value and its corresponding optimal linear and angular velocities $\textbf{v}^*$
(see \autoref{eq:v*}). Different discretization steps are used for the three controlled velocities in order to keep a balance between the computation time and the velocity jumps. Against other state of the art local planners that achieve  non linear optimizations, the computation time is bounded, only depending on the discretization parameters. Moreover, the method directly computes optimal linear and angular velocity commands providing feasible motions under the kinodynamic constraints, according with the objective function, without the necessity of calculating complex trajectories between subgoals.

\subsubsection{Heading}
The $Heading$ term is the responsible of guiding the drone towards the goal by encouraging the alignment between them, both in the XY plane (orientation) and in height, which is a key difference from the original 2D method. Additionally, two more weights, $K_{\psi}$ and $K_z$, are introduced. To keep with the normalization criteria, they must also add up to 1, allowing to prioritize one or the other. By tuning these weights, the preference in the avoiding behavior can be chosen. A greater value of $K_{\psi}$ will prioritize staying oriented with the current subgoal, thus vertical avoidance will be preferred. In contrast, a higher value of $K_z$ will try to keep the UAV at the same height than the current waypoint, giving more importance to lateral obstacle avoidance. 

\begin{itemize}
    \item \textbf{$Head_{\psi}$}:
It rewards velocities that help staying oriented towards the goal (\autoref{fig:heading_yaw}). First the angle between the drone and the goal locations, $\psi^{rel}$, is computed,

\begin{subequations}
    \begin{equation}
    \psi^{rel} = \atantwo(\Delta y, \Delta x)
    \end{equation}
    \begin{equation}
    \ \Delta x = x_g - x',  \hspace{0.75cm} \Delta y = y_g - y'
    \end{equation}
\end{subequations}

being $x_g, y_g$ the goal coordinates in the world reference. Finally, the orientation of the drone is considered to compute the normalized term (\autoref{fig:heading_yaw}).
\begin{equation}
    Head_{\psi} = 1 - \frac{\lvert \psi^{rel} - \psi' \lvert}{\pi} ; \hspace{0.35cm} \psi^{rel} - \psi' \in [-\pi,\pi]
\end{equation}

\item\textbf{$Head_z$}:
In this case, the difference in height with respect to the tracked waypoint wants to be minimized,
\begin{equation}
    \ \Delta z_i = \lvert z_g - z'_i\lvert
\end{equation}
The normalization of this term is performed by finding its maximum value of those computed at the current iteration and dividing by it,
\begin{equation}
    \ Head_z = 1 - \frac{\Delta z_i}{\underset{i}{\max{}} \Delta z_i}
\end{equation}

\begin{figure}[h]
    \centering
    \begin{subfigure}{0.45\columnwidth}
        \includegraphics[width=\linewidth]{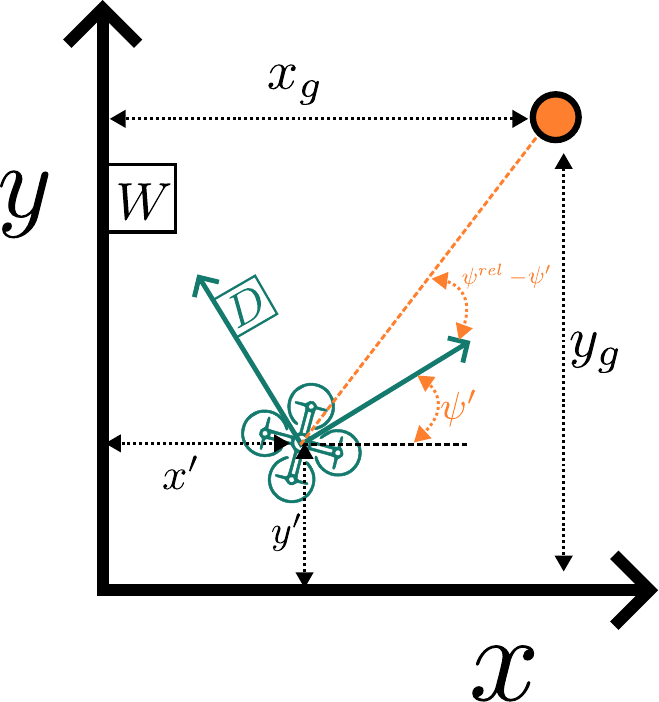}
        \caption{$Head_{\psi}$.}
        \label{fig:heading_yaw}
    \end{subfigure}
    \begin{subfigure}{0.45\columnwidth}
        \includegraphics[width=\linewidth]{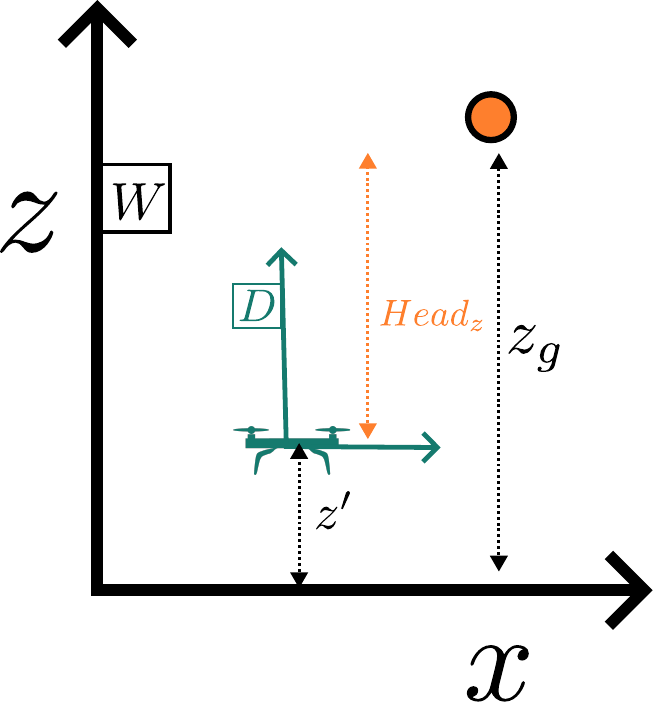}
        \caption{$Head_z$.}
        \label{fig:heading_z}
    \end{subfigure}
    \hfill
    \caption{Graphic $Head_{\psi}$ and $Head_z$ computation for a predicted pose $\{x', y', z'\}$. The orange point represents the tracked waypoint.}
    \label{fig:Heading_computation}
\end{figure}

\end{itemize}

\subsubsection{Distance}
\begin{figure}[h!]
    \centering
    \includegraphics[width=\columnwidth,keepaspectratio]{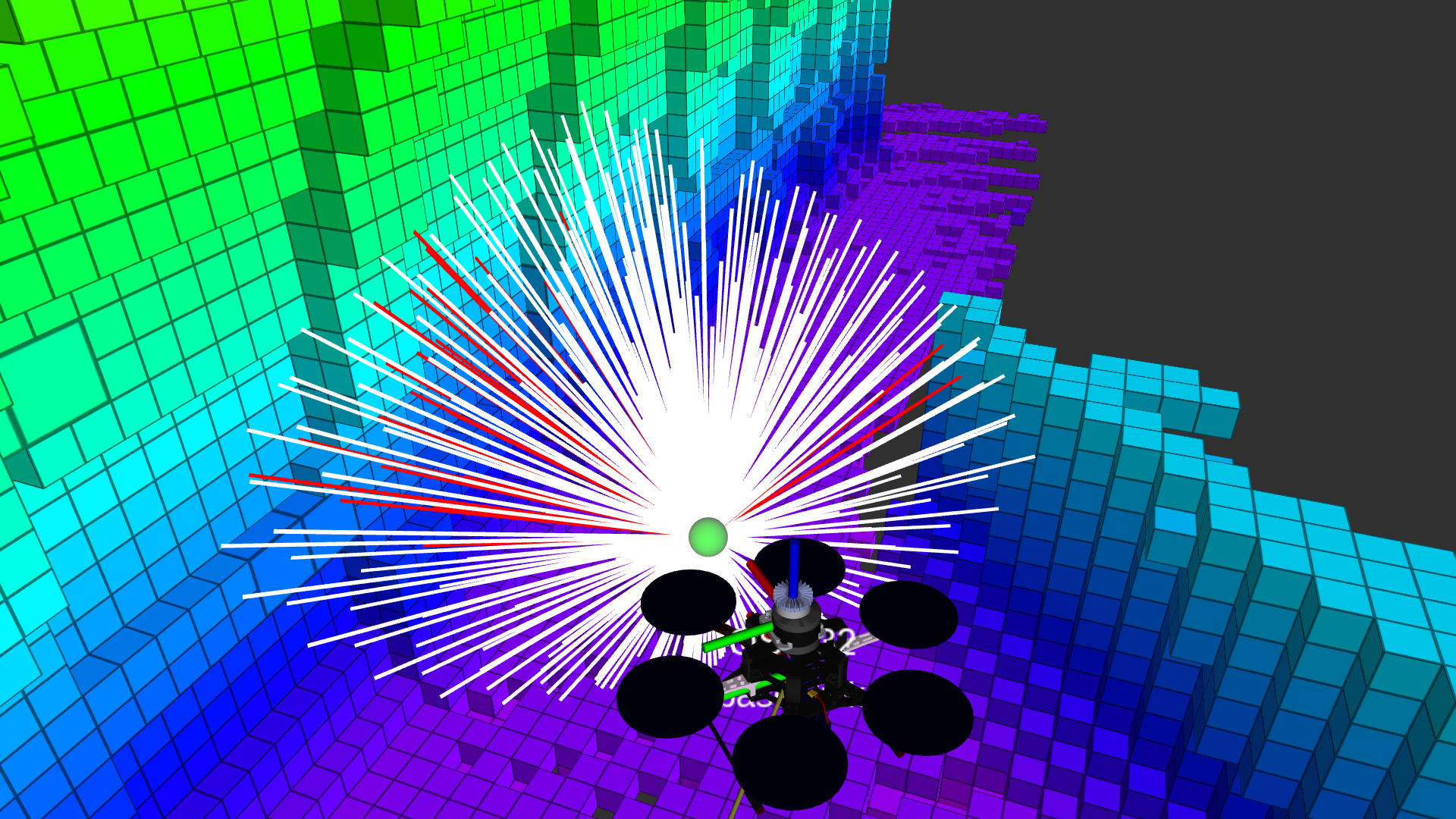}
    \caption{Example of the rays casted (white and red) from the predicted pose (green) during the distance term computation of DWA-3D. 
    Note that the further they are from the motion direction, the shorter they are, rewarding changing the motion direction during avoidance. The red rays represent those that have found an obstacle. These shortest of the red rays is the one used for computing the Distance term in \autoref{eq:G_compressed}.}
    \label{fig:Rays_EXAMPLE}
\end{figure}

\begin{figure}[h!]
    \centering
    \begin{subfigure}{0.45\columnwidth}
        \includegraphics[width=\linewidth, trim = 2cm 8cm 2cm 7cm, clip]{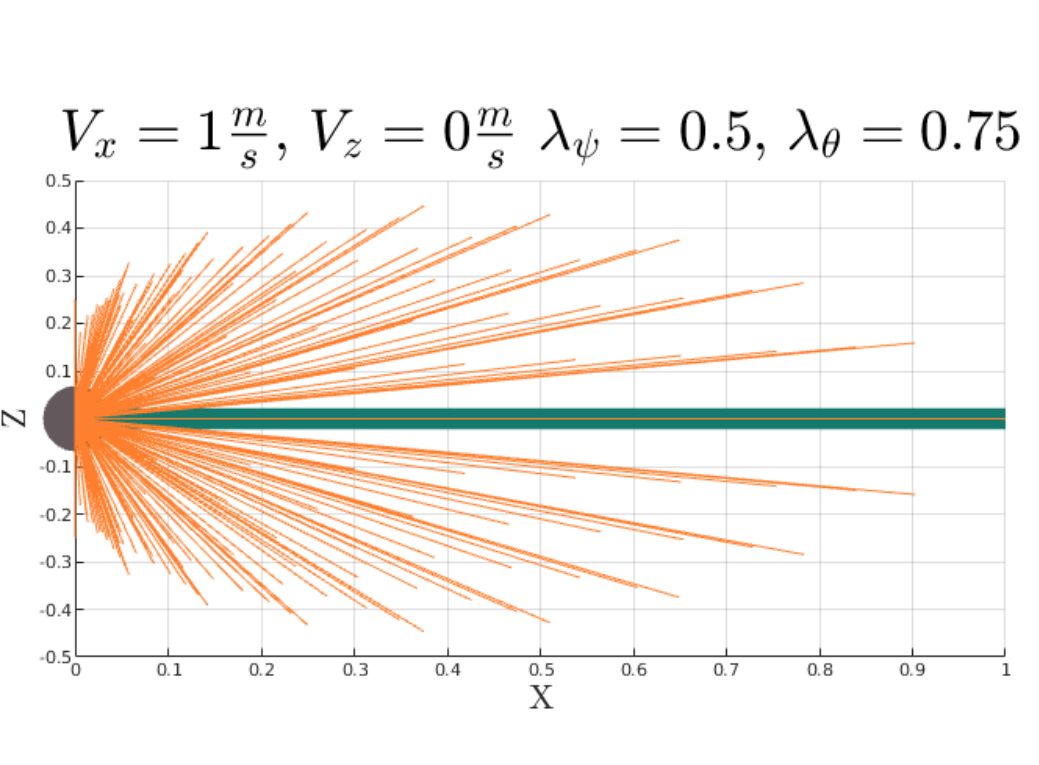}
        \caption{$v_x = 1 \frac{m}{s}$, $v_z = 0 \frac{m}{s}$.}
    \end{subfigure}
    \begin{subfigure}{0.45\columnwidth}
        \includegraphics[width=\linewidth, trim = 2cm 8cm 2cm 7cm, clip]{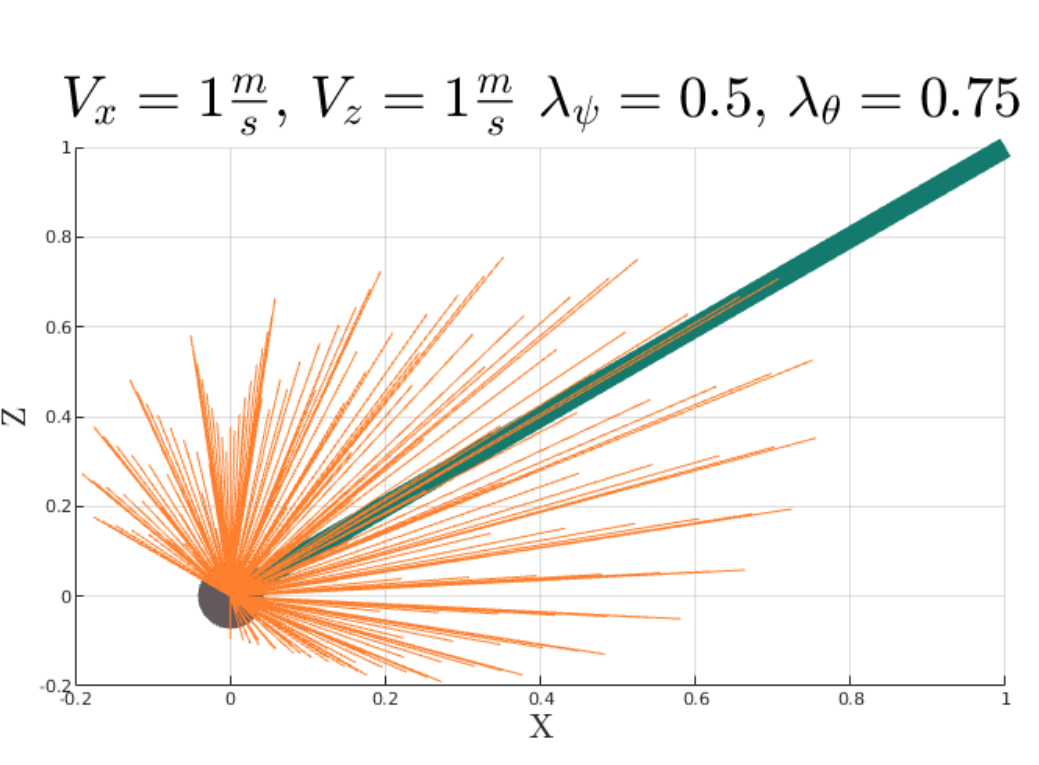}
        \caption{$v_x = 1 \frac{m}{s}$, $v_z = 1 \frac{m}{s}$.}
    \end{subfigure}
    \begin{subfigure}{0.45\columnwidth}
        \includegraphics[width=\linewidth, trim = 2cm 8cm 2cm 7cm, clip]{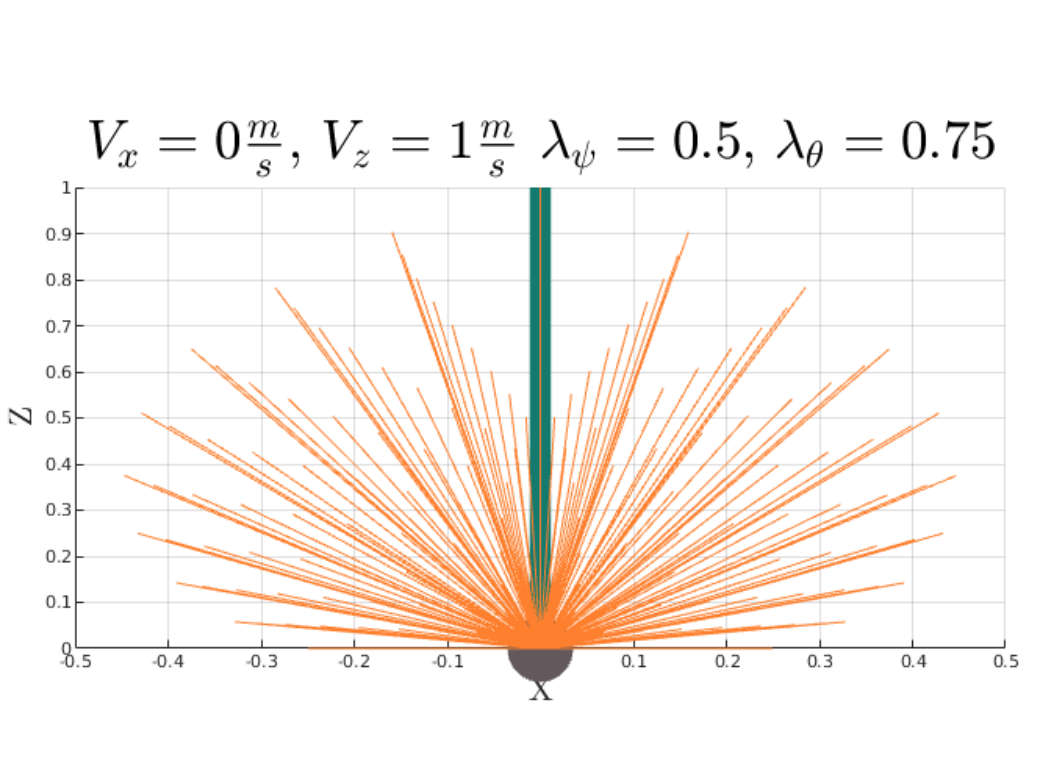}
        \caption{$v_x = 0 \frac{m}{s}$, $v_z = 1 \frac{m}{s}$.}
    \end{subfigure}
    \hfill
    \caption{Ray beam (orange) reorientation according to the angle of $\overrightarrow{V}$ (green) during the distance term computation with different $v_x$ and $v_z$ combinations, with $r_{search} = 1m$, $\lambda_{\psi} = 0.5$ and $\lambda_{\theta} = 0.75$.}
    \label{fig:Beam_heading}
\end{figure}

This term considers the obstacles that could cause a collision, penalizing those sets of velocities that are more dangerous. It is evaluated in the predicted positions by casting rays against the map up to certain distances $L_{ij}$ (\autoref{eq:Lij}) according to the direction of the ray with respect to the robocentric $X$ axis in $D$ reference, let $r_{search}$ be the length of the ray casted in that direction (robocentric $X$ axis): 
\begin{equation}
    \label{eq:Lij}
    \ L_{ij} = r_{search} \cdot \rho_{\psi}^{i} \cdot \rho_{\theta}^{j}
\end{equation}
\begin{equation}    
    \label{eq:rho_dist}
    \ \rho_{\psi}^{i} = 1 - \lambda_{\psi} \cdot \frac{|\psi_{i}|}{\psi_{max}^{beam}};  \hspace{0.75cm} \rho_{\theta}^{j} = 1 - \lambda_{\theta} \cdot \frac{|\theta_{j}|}{\theta_{max}^{beam}}
\end{equation}

where $\psi_{i}$ and $\theta_{j}$ are the angles that the ray forms with the $X$ axis of the drone in the $XY$ and the $XZ$ planes respectively, $\lambda_{\psi}$ and $\lambda_{\theta}$ parameters allow to adjust the lateral and vertical safe distances, while $\psi_{max}^{beam}$ and $\theta_{max}^{beam}$ are the maximum angles in which a ray is casted. An example of this operation can be observed in \autoref{fig:Rays_EXAMPLE}. $L_{ij}$ increases in the motion direction, enforcing a greater change of orientation or flying height when an obstacle is encountered near this direction, being smaller when an obstacle is detected angularly farther of that motion direction.  
Additionally, the beam is reoriented according to the motion direction by adding the argument of $\overrightarrow{V} = \{v_x, 0, v_z\}$ to it, see \autoref{fig:Beam_heading}. 

Once all the rays have been cast, the one that found an obstacle (or an unknown voxel) at the closest distance is considered for the computation of the normalized $Distance$ term:
\begin{equation}
    \label{eq:Dist_Normalized}
    \ Distance(\textbf{v}) = \max(0, \frac{dist_{min} - R_{drone}}{r_{search} - R_{drone}})
\end{equation}
where $R_{drone}$ is the radius of the UAV and must be taken into account to avoid collisions; as the minimum distance in the predicted position could be smaller than $R_{drone}$ the subtraction is saturated at 0. 

Casting rays against the map instead of consulting the raw pointcloud offers not only a reduction of the elements that have to be analysed but also a more efficient way of accessing them. Additionally, the map's persistence, allows incorporating previously sensed regions, even if outside the sensor field of view, avoiding sudden dangerous reactions and oscillatory movements.

\begin{algorithm}[h]
    \caption{Distance term evaluation.}
    \label{alg:Dist_algo}
    \begin{algorithmic}[1]
        \State $\textbf{Input: } (x', y', z', \psi', v_x, v_z, Map)$  
        \State $\textbf{Params: } (r_{search}, \psi_{max}^{beam}, \theta_{max}^{beam}, \lambda_{\psi}, \lambda_{\theta}, \delta_{\psi}, \delta_{\theta}, R_{drone})$
        \State $\theta_{beam} \gets \atantwo(v_z,v_x)$
        \State $dist_{min} \gets r_{search}$ 
        \For{$\psi_{i}$ \textbf{from} $-\psi_{max}^{beam}$ \textbf{to} $\psi_{max}^{beam}$ \textbf{with step} $\delta_{\psi}$}
            \State $\rho_{\psi} \gets r_{search} \cdot (1 - \lambda_{\psi} \cdot \frac{|\psi_{i}|}{\psi_{max}^{beam}})$
            \For{$\theta_{j}$ \textbf{from} $-\theta_{max}^{beam}$ \textbf{to} $\theta_{max}^{beam}$ \textbf{with step} $\delta_{\theta}$}
                \State \hskip-0.5em $\rho_{\theta} \gets r_{search} \cdot (1 - \lambda_{\theta} \cdot \frac{|\theta_{j}|}{\theta_{max}^{beam}})$
                \State \hskip-0.5em $L_{ij} \gets r_{search} \cdot \rho_{\psi} \cdot \rho_{\theta}$
                \State \hskip-0.5em $x_r \gets \cos(\psi_{i}+\psi') \cdot \cos(\theta_{j}+\theta_{beam})$
                \State \hskip-0.5em $y_r \gets \sin(\psi_{i}+\psi') \cdot \cos(\theta_{j}+\theta_{beam})$ 
                \State \hskip-0.5em $z_r \gets \sin(\theta_{j}+\theta_{beam})$
                \State \hskip-0.5em $ray \gets (x_r,y_r,z_r)$
                \State  \hskip-0.5em $dist_{ij} \gets $\textbf{Cast} $ray$ \textbf{From }$(x',y',z')$ \textbf{UpTo }$L_{ij}$
                \State \hskip-0.5em $dist_{min} \gets \min(dist_{min}, dist_{ij})$
            \hskip-0.5em \EndFor
        \EndFor \\
        $Dist=\frac{dist_{min} - R_{drone}}{r_{search} - R_{drone}}$ \\
        \Return $\max(0, Dist)$
    \end{algorithmic}
\end{algorithm}

Finally, the term $Dist(\textbf{v})$ in \autoref{eq:G_expanded} is obtained from the output of \autoref{alg:Dist_algo}.

\subsubsection{Velocity}
The velocity term makes higher speeds preferred so that the drone is forced to move, avoiding staying still when it is aligned with the goal, additionally reaching the target in the shortest possible time. Whereas high $v_z$ values are already promoted  by $Head_z$ when there are significant differences between the UAV height and the waypoint one, this is not the case of $v_x$. According to the $Heading$ preference chosen (orientation or height) the advance speed should be enhanced under certain situations: 
\begin{itemize}
    \item $K_z > K_{\psi}$. Always, as it encourages the drone to move forward while maintaining its current height during lateral obstacle avoidance maneuvers.
    \item $K_{\psi} > K_z$. Only when the waypoint is in front of the UAV,  promoting forward progress while allowing to fly over and under obstacles if they are between the robot and the tracked point. 
\end{itemize} 

\begin{equation}
\centering
    Vel(\textbf{v}) = \begin{cases}
        \frac{v_x}{v_x^{max}} & \text{if } K_z > K_{\psi} \text{ or } \\ 
        \ & (Head_{\psi} > 0.5  \text{ and } K_z < K_{\psi})\\
        \ 0  & \text{otherwise} 
    \end{cases}
\end{equation}

\section{Global Planner}
\begin{figure}[t]
    \centering
    \includegraphics[width=\columnwidth, trim = 6cm 5cm 18cm 5cm, clip,keepaspectratio]{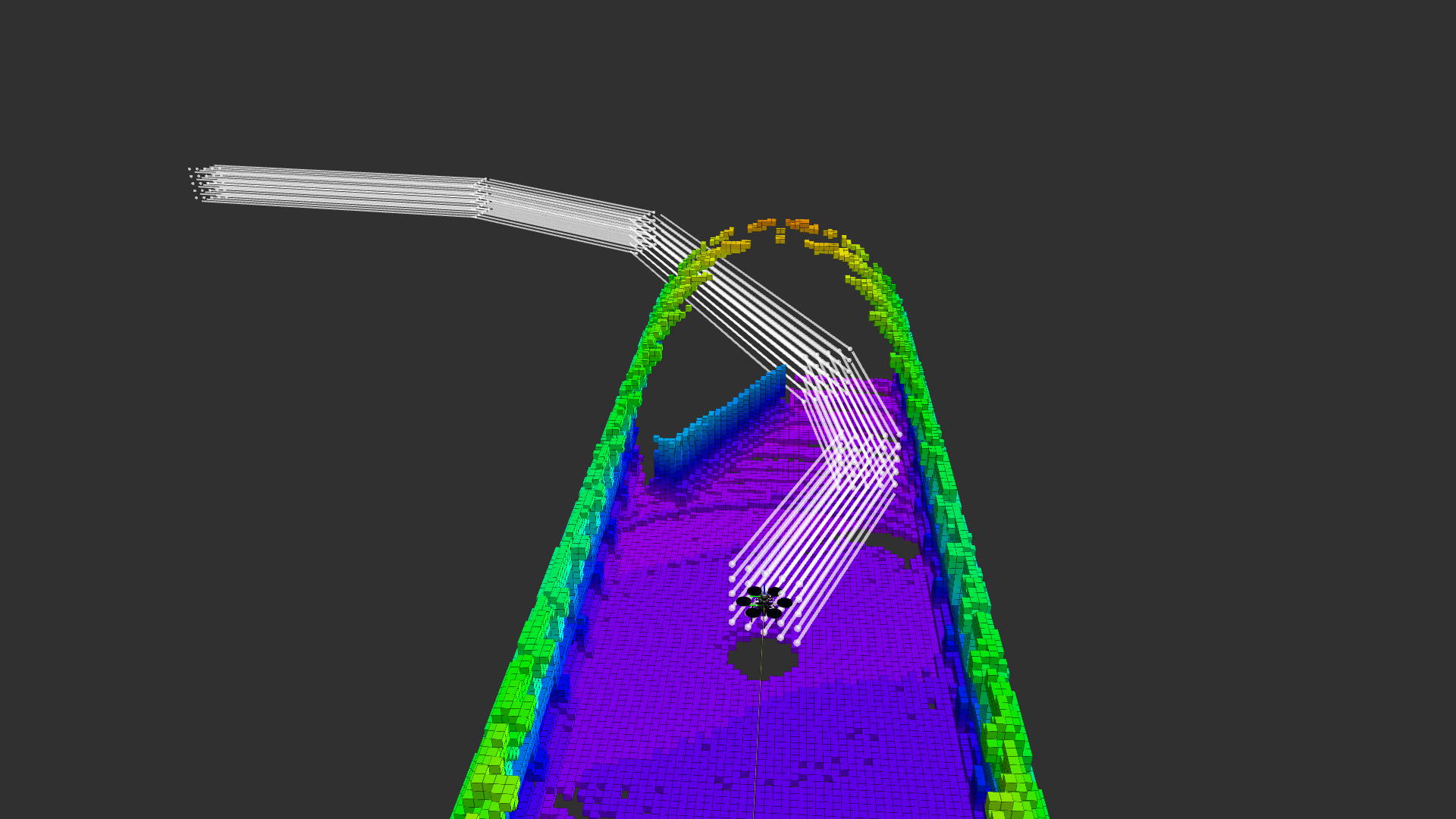}
    \caption{Example of a path (white) computed by RRT* with the size-aware approach. On the one hand, the safety distance is checked around in the waypoints (spheres). On the other hand, a parallelepiped (lines) is casted between consecutive waypoint's safe regions. }
    \label{fig:RRT_Paralelepipedo}
\end{figure}

The Global Planner computes a preliminary plan to follow, given the whole information known about the environment, according to a certain optimization criteria that could be imposed to the system. Because of UAVs' reduced battery autonomy, the desired behavior would be taking the shortest path possible, thus minimizing the sum of the euclidean distance between waypoints; while being aware of not approaching too much to the obstacles taking into account the drone size (size-aware planner). That is why a safety distance can be imposed to the waypoints computed by the global planner, both by checking its surroundings and the parallelepiped that connects two consecutive waypoint's safe regions, \autoref{fig:RRT_Paralelepipedo}. Additionally, as in the local planner, it would be desirable to choose if the trajectory prefers moving laterally or vertically, so a the different global path can be computed. The following expression, is minimized during global plan computation:
\begin{adjustwidth}{-0.5cm}{0cm}
\begin{multline}
   \ D_{wp} = K_{length} \sum_{i = 1}^{n-1} d(\mathbf{g_i}, \mathbf{g_{i+1}}) + K_{height} \sum_{i = 1}^{n-1} |z_n -z_i|
\end{multline}
\label{eq:Optimization_global}
\end{adjustwidth}
where $n$ is the number of waypoints $\mathbf{g_i}$ that conform a feasible solution with states sampled by the RRT*, $d()$ is the 3D euclidean distance between two points, and $z_n$ is the Z coordinate of the goal position. $K_{length}$ and $K_{height}$ allow to balance between shortening the path and being at the same height than the destination. Note that although distance to obstacles is not included in \autoref{eq:Optimization_global}, it is considered during the optimization to discard waypoints if they involve a collision or path segments if they intersect one obstacle.  

\begin{table*}[t]
    \centering
    \begin{subtable}[h]{0.6\textwidth}
    \begin{tabular}{|*{4}{c|}}
        \hline
        \multicolumn{4}{|c|}{General Parameters} \\
        \hline
        \textbf{Param.} & \textbf{Range} & \textbf{Rec. Value} & \textbf{Restrictions} \\
        \hline
         $\alpha$ & [0, 1] & 0.3 & \multirow{3}{5em}{$\alpha + \beta + \gamma = 1$} \\
         $\beta$ & [0, 1] & 0.6 & \\
         $\gamma$ & [0, 1] & 0.1  & \\
         \hline
         $T$ & X & 100ms  & \multirow{2}{1em}{}\\
         $\Delta t$ & $> T$ & $10\times T = 1s$ & \\
         \hline
         
        \multicolumn{4}{|c|}{Avoidance Behavior Selection} \\
        \hline
        \textbf{Param.} & \textbf{Range} & \textbf{Rec. Value} & \textbf{Restrictions} \\
        \hline
         \multirow{2}{1em}{$K_{z}$} & \multirow{2}{2.5em}{[0, 1]}  & 0.8 ($\leftrightarrow$ avoid) & \multirow{4}{5em}{$K_{\psi} + K_z = 1$} \\
         & & 0.2 ($\updownarrow$ avoid) & \\
         \multirow{2}{1em}{$K_{\psi}$} & \multirow{2}{2.5em}{[0, 1]}  & 0.2 ($\leftrightarrow$ avoid) &  \\
         & & 0.8 ($\updownarrow$ avoid)& \\
         \hline
         
        \multicolumn{4}{|c|}{Risk Tolerance in Obstacle Avoidance} \\
        \hline
        \textbf{Param.} & \textbf{Range} & \textbf{Rec. Value} & \textbf{Restrictions} \\
        \hline
         $r_{search}$ & $> R_{drone}$ & $1m \sim 1.5m$ & \multirow{3}{8.8em}{$r_{search} (1 - \lambda_{\psi}) > R_{drone}$ $r_{search} (1 - \lambda_{\theta}) > H_{drone}$} \\
         $\lambda_{\psi}$ & [0, 1] & 0.5 & \\
         $\lambda_{\theta}$ & [0, 1] & 0.75 & \\
         \hline
         $\psi_{max}^{beam}$ & [0, 180$^{\circ}$] & 90$^{\circ}$ & \multirow{2}{1em}{} \\
         $\theta_{max}^{beam}$ & [0, 180$^{\circ}$] & 90$^{\circ}$ & \\
         \hline
    \end{tabular}
    \end{subtable}
    \begin{subtable}[h]{0.3\textwidth}
    \begin{tabular}{|*{4}{c|}}
        \hline
        \multicolumn{4}{|c|}{Dynamic Window Discretization} \\
        \hline
        \textbf{Param.} & \multicolumn{2}{|c|}{\textbf{Value}} & \textbf{Restrictions} \\
        \hline
         $v^{step}_{x}$ & \multicolumn{2}{|c|}{$0.05 m/s$
         } & \multirow{3}{6em}{Computation time and gross discretization}\\
         $v^{step}_{z}$ & \multicolumn{2}{|c|}{$0.05 m/s$
         } & \\
         $\omega_{z}^{step}$ & \multicolumn{2}{|c|}{$2.5  ^{\circ}/s$} & \\
         \hline
         
        \multicolumn{4}{|c|}{Hexarotor Physical Parameters} \\
        \hline
        \textbf{Param.} & \multicolumn{3}{|c|}{\textbf{Value}} \\
        \hline
        $m$ & \multicolumn{3}{|c|}{$3.65 kg$} \\
        $R$ & \multicolumn{3}{|c|}{$0.4 m$} \\
        $H$ & \multicolumn{3}{|c|}{$0.3m$} \\
        \hline
        \multicolumn{4}{|c|}{Hexarotor Dynamic Parameters} \\
        \hline
        \textbf{Param.} & \multicolumn{2}{|c|}{\textbf{Value}} & \textbf{Restrictions} \\
        \hline
         $v^{max}_{x}$ &  \multicolumn{2}{|c|}{$0.3 m/s$
         } & \multirow{6}{6em}{Environment size and motors limits}\\
         $v^{max}_{z}$ &  \multicolumn{2}{|c|}{$0.3 m/s$
         } & \\
         $\omega_{z}^{max}$ &  \multicolumn{2}{|c|}{$45 ^{\circ}/s$ } & \\
         $\dot{v}^{max}_{x}$ &  \multicolumn{2}{|c|}{$1 m/s^2$
         } & \\
         $\dot{v}^{max}_{z}$ &  \multicolumn{2}{|c|}{$1 m/s^2$
         } & \\
         $\dot{\omega}_{z}^{max}$ &  \multicolumn{2}{|c|}{$100 ^{\circ}/s^2$}  & \\
         \hline
    \end{tabular}
    \end{subtable}
    \caption{Main DWA parameters summary, with their admissible ranges and recommended values.}
    
    \label{tab:DWA_parameters}
\end{table*}

Moreover, to test the DWA-3D maneuvering capabilities under complex situations as, for instance, paths that drive the drone to move in narrow rooms, global planner's size awareness can be disabled. When deactivated, only the line of sight between waypoints is verified (not size-aware planner). In \autoref{sec:LabExperiments}, the integration with three variants of the global planner is described: Naive (direct straight line to the goal), Not size-aware (the drone size is not considered), and Size-aware (the drone size is taken into account).  

To enhance the flexibility of the system, the OMPL library \cite{OMPL} has been used as to implement the core of this subsystem, offering the chance to select between a wide number of SOTA global planning algorithms. During the real experiments RRT* \cite{RRT*} has been chosen. Note that the main scope of this work is to present the local planner, DWA-3D. Thus, global planning has only been a tool to show the reactive planner's capabilities. Global planner refinement is out of the scope of project. 

\section{Experimental Results}

In this section several experiments have been addressed for evaluating the performance of the proposed method in different situations. The main issue we want to evaluate is the capability of DWA-3D jointly with a global planner for avoiding the obstacles, following the global path planned if possible.  The motion has to comply the desired behavior of the avoidance, laterally or vertically, depending on the scenario, whilst always ensuring the drone safety, being DWA-3D the one that makes the final motion decision.

We decided to split the experimental validation of the method in two phases. The first one aims to evaluate DWA-3D capabilities, compare the effect of the global planner selection and verify that safe autonomous navigation is achieved. These initial experiments were carried out in a controlled environment, where a motion capture system provides reliable localization, allowing us to decouple the localization and navigation problems. The second phase involves facing field scenarios out of the controlled one, where no external localization source is available. Here, the architecture demonstrates its capabilities in field scenarios. In \autoref{sec:LabExperiments} we present the real-world experiments in controlled environments inside our laboratory, whereas \autoref{sec:FieldExperiments} contains the field ones.

\subsection{Setup}
\label{sec:setup}
\begin{figure}[h]
    \centering
    \includegraphics[width= \linewidth,keepaspectratio, trim = 0.5cm 1.5cm 0cm 2cm, clip]{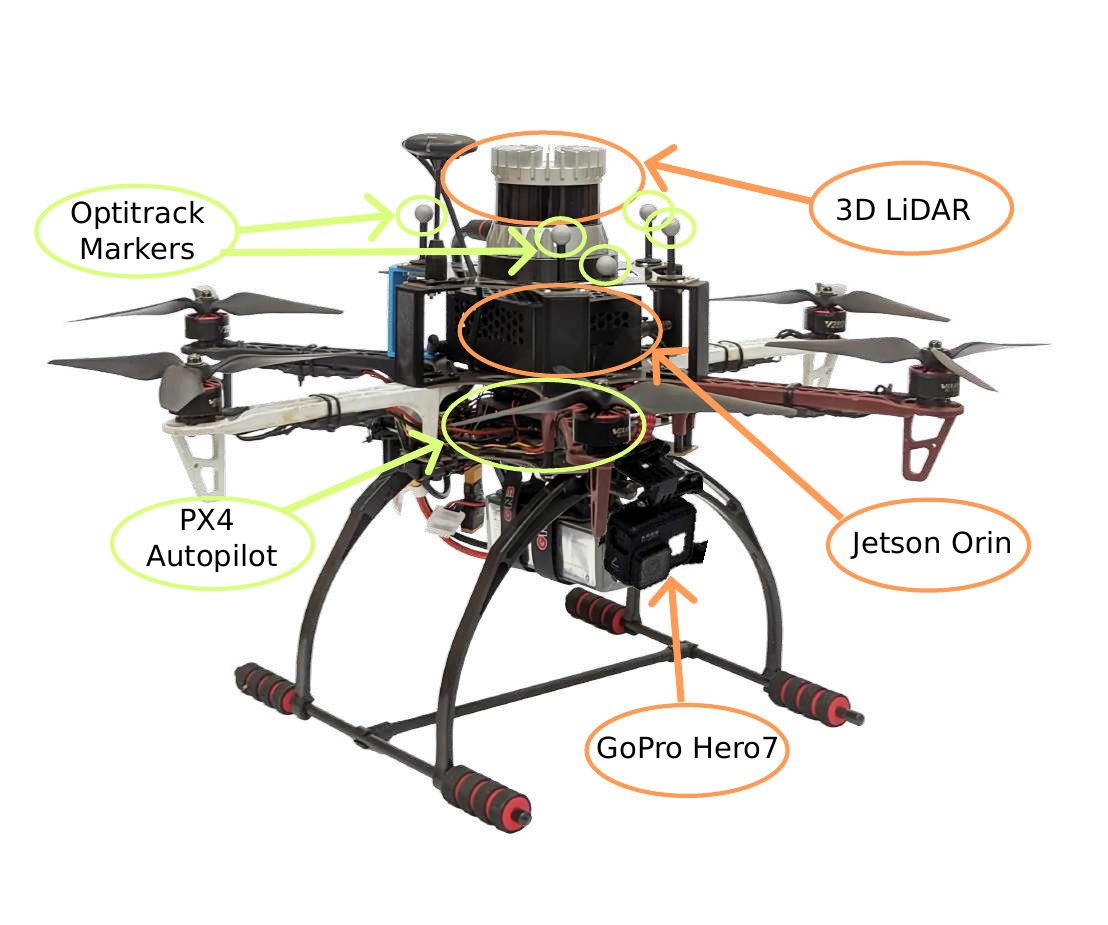}
    \caption{Custom hexarotor used to perform the real experiments, equipped with a OS0-32 3D LiDAR, a Jetson Orin Nano Devkit, a PX4 flight controller and GoPro Hero 7 for visual inspection.}
    \label{fig:tyrion_image}
\end{figure}

The real experiments have been performed with a custom hexarotor (\autoref{fig:tyrion_image}) built with a DJI F550 frame. It has 0.8 m of diameter. The described architecture is fully implemented in an on-board Jetson Orin Nano Devkit, while a Pixhawk 4 is in charge of the flight control. The sensing is performed with an Ouster OS0-32 3D LiDAR, which offers a vertical resolution of 32 planes of 1024 points each and a field of view of $360^{\circ}\times90^{\circ}$. A GoPro is mounted aiming forward, enabling FPV (First Person View), for performing, for instance, inspection tasks. 

\begin{figure*}
    \centering
    \begin{subfigure}{0.35\textwidth}
        \centering
      \includegraphics[height=4.7cm,keepaspectratio]{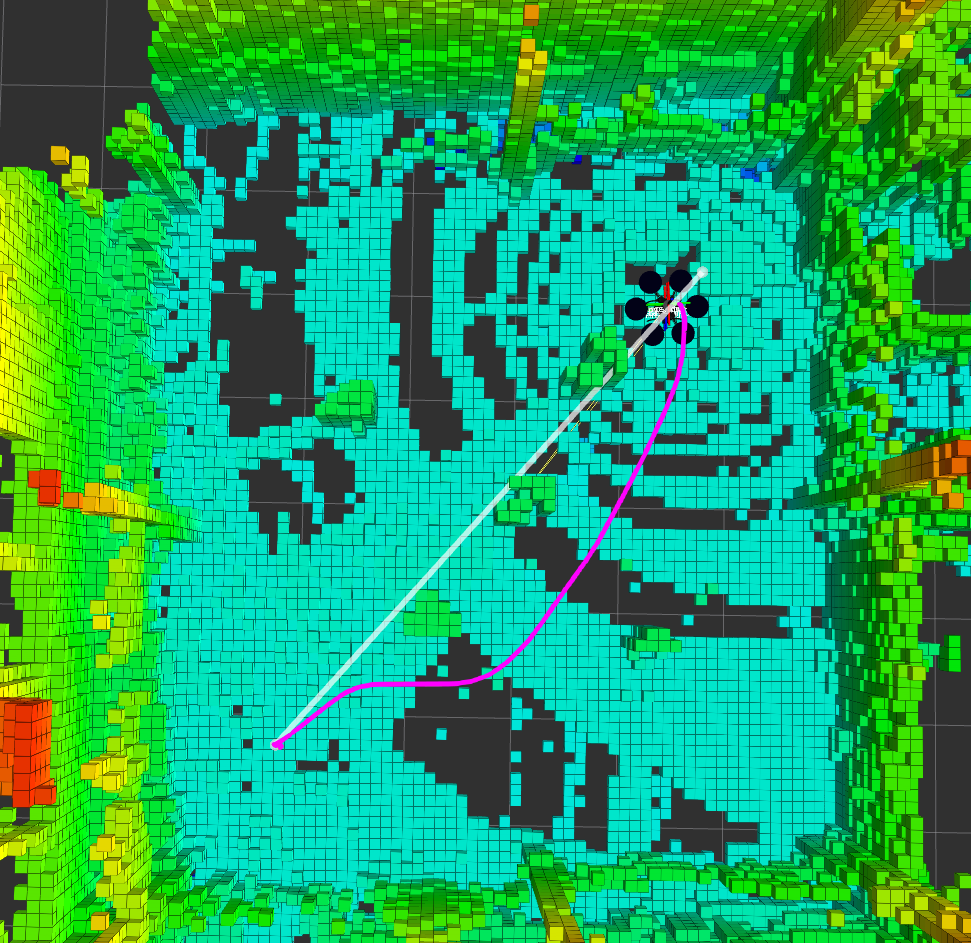}
        \caption{Naive path (straight line); drone does not follow it, passing by the middle of two obstacles.}
        \label{fig:PATH_NO_GLOBAL}
    \end{subfigure}
    \begin{subfigure}{0.3\textwidth}
        \centering
      \includegraphics[height=4.7cm,keepaspectratio]{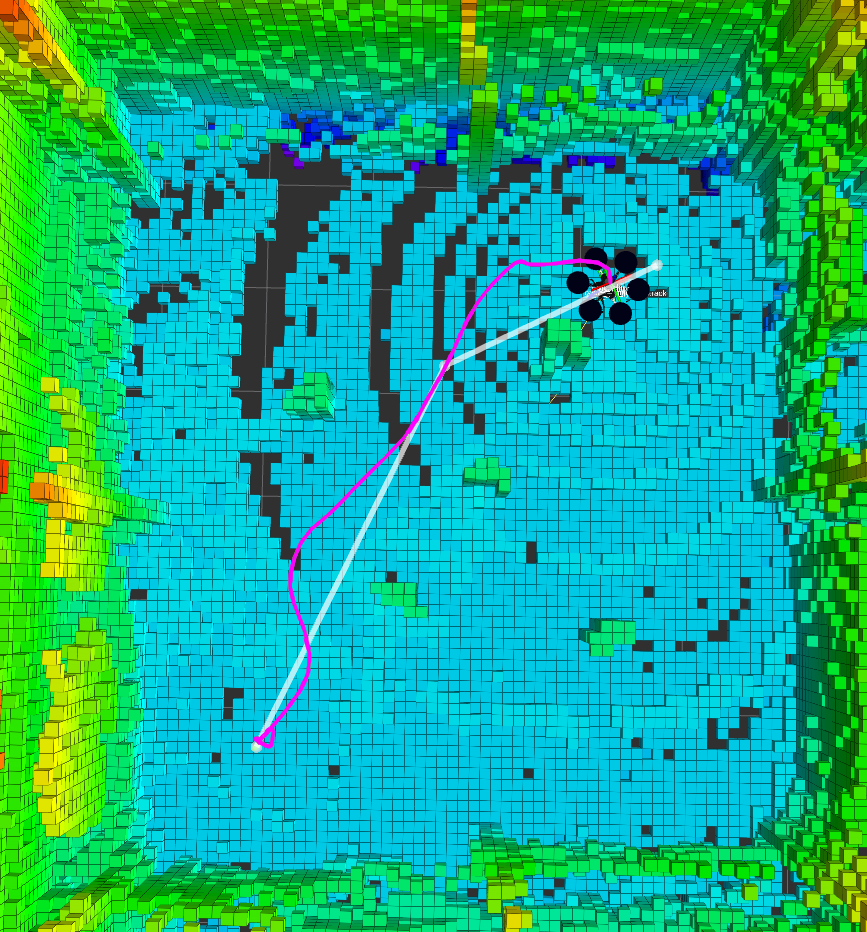}
        \caption{RRT* NOT size aware path passing too near \\ to obstacles; DWA-3D keeps the safety distance.}
        \label{fig:PATH_GLOBAL_NO_SIZE}
    \end{subfigure}
    \begin{subfigure}{0.3\textwidth}
        \centering
       \includegraphics[height=4.7cm,keepaspectratio]{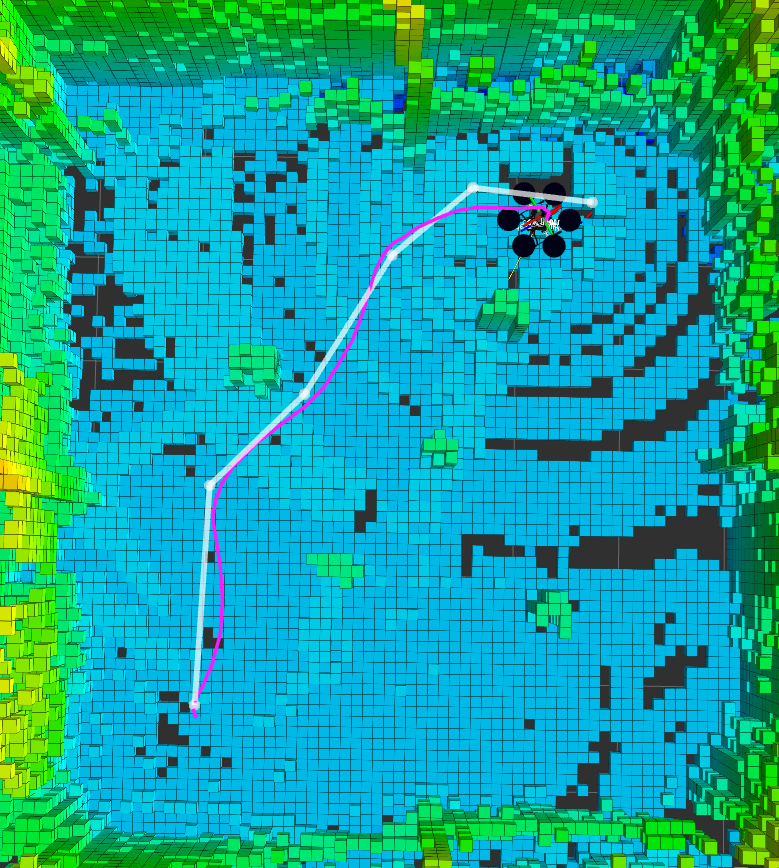}
        \caption{RRT* size aware path avoiding the areas where the drone does not fit; it follows the planned path.}
        \label{fig:PATH_GLOBAL_SIZE}
    \end{subfigure}
    \hfill
    \caption{Global planning situations under which the DWA local planner has been tested. Planned path (white) and performed trajectory (purple) with the real-time updated Octomap after the flight. More intense green colour represents obstacles. Note that the map is not known beforehand, thus the path may pass near obstacles that initially were partially or fully occluded.}
    \label{fig:Planners_example}
\end{figure*}

\subsection{Parameters Configuration}
\label{sec:ParamsConfig}
As it has been shown along \autoref{sec:DWA}, several parameters are involved in the DWA-3D computations. \autoref{tab:DWA_parameters} shows a summary of the main parameters, suggested values and the ones used in the experiments. This section presents the main equations that sustented the parameter selection. For a further mathematical analysis, please refer to \hyperref[sec:AppendixA]{Appendix A}. The maximum velocities were restricted to those values according with the nature of the scenarios and for ensuring safe maneuvers when the obstacles are very close. They could be increased in the cases of larger environments and more clearance among the obstacles. On the other hand, the RRT* safety distance (when used) has been established to $\frac{r_{search}}{2}$, being $r_{search}$ the  frontal safety distance used for the local planner, see \autoref{eq:Lij}. 

Regarding DWA-3D, its parameters must comply the following relationships deduced from \hyperref[sec:AppendixA]{Appendix A}, 
\begin{equation}
    \beta > \alpha
    \label{eq:b>alpha}
\end{equation}
\begin{equation}
            \beta \cdot \lambda_{\psi} > \alpha \cdot \frac{\omega_{z}^{max} \cdot \Delta t}{\pi}
            \label{eq:b*lambda>alpha}
\end{equation}
\begin{equation}
    \ \beta > \gamma
    \label{eq:b>gamma}
\end{equation}
\begin{equation}
    \ \alpha \cdot max(K_z, K_{\psi}) > \gamma
    \label{eq:alpha>gamma}
\end{equation}

We applied those constraints during parameter selection, resulting in a safe autonomous system as we will show in the remaining of this section.

\subsection{Laboratory Experiments}
\label{sec:LabExperiments}
The autonomous navigation of the drone has been tested first in our "Unizar Drone Arena", a $6 \times 6 \times 6$m safe area, equipped with a the motion capture system. It has provided a reliable and accurate UAV localization, allowing us to focus on verifying the reactive navigation algorithm. 
Several configurations with obstacles have been proposed under different circumstances and preferences. As one of our main contributions is the DWA-3D reactive navigator, three different navigation circumstances have been proposed in order to be able to analyze its performance and effect.
\begin{enumerate}
    \item \textbf{Without global planner (Naive)}(\autoref{fig:PATH_NO_GLOBAL}): Under this configuration, the path that is given to the DWA-3D local planner is a straight line from the starting point of the drone to the desired final position, thus it must find the way to reach the goal only using reactive navigation.
    \item \textbf{With NOT size-aware global planner} (\autoref{fig:PATH_GLOBAL_NO_SIZE}): A first approach to introduce a global planner has been without taking the drone size into account, which leads to paths that guide it towards the goal through short paths, but that could involve traversing dangerous areas that pass too near to obstacles. 
    \item \textbf{With size-aware global planner} (\autoref{fig:PATH_GLOBAL_SIZE}): Finally, a safer global planner has been integrated by leaving a safety distance between the planned path and the obstacles, thus gaps that are smaller than the drone size are avoided. Nevertheless, the DWA-3D reactive planner is still needed as the map is not known beforehand. The global path is computed using the initial UAV surroundings, so hidden obstacles would still represent a hazard if navigation only depended on the global planner. 
\end{enumerate}

\begin{figure*}
    \centering
    \begin{subfigure}{\textwidth}
        \centering
        \includegraphics[width=0.75\linewidth,keepaspectratio]{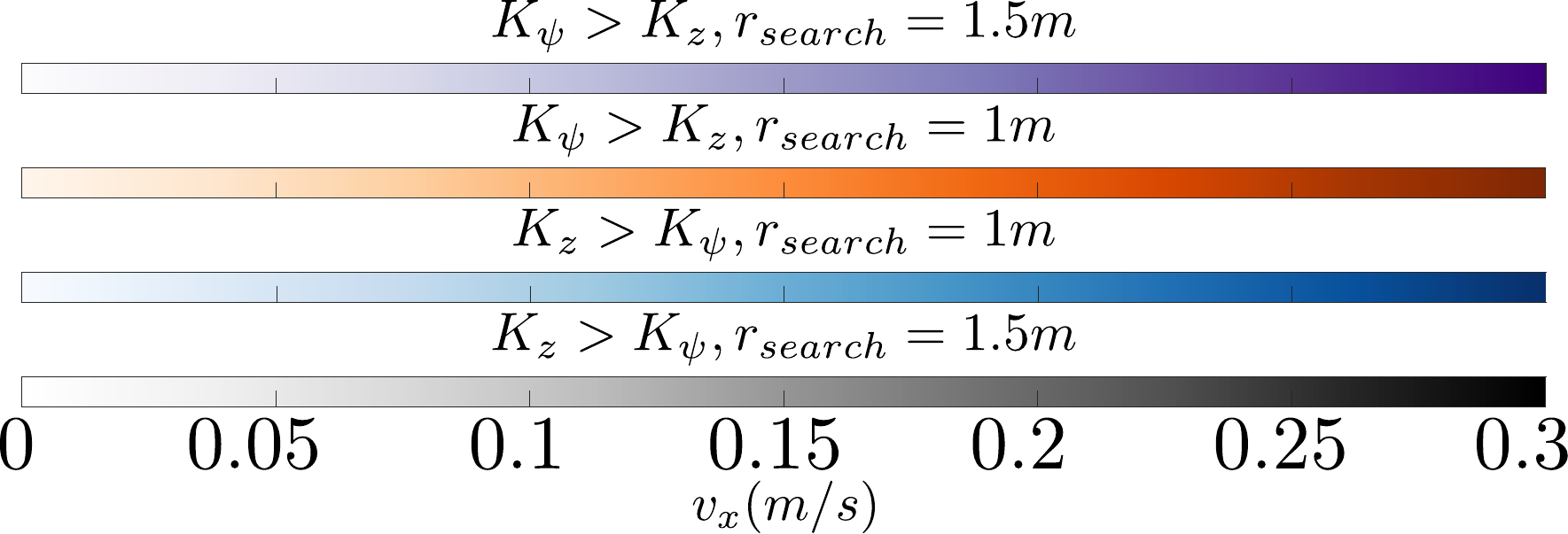}
        \includegraphics[width=8cm,keepaspectratio, trim = 3cm 0cm 6cm 0cm, clip]{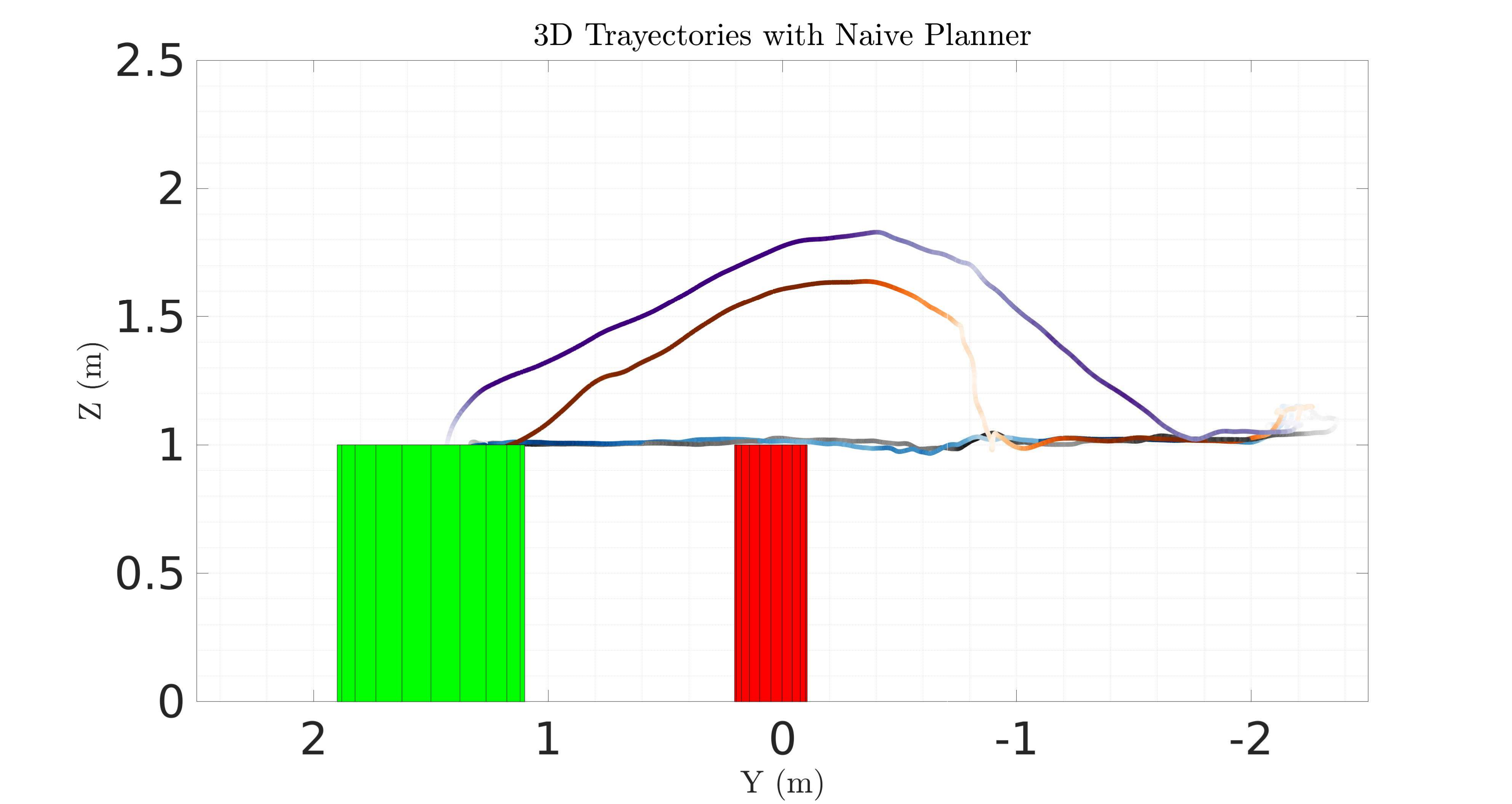}
        \includegraphics[width=8cm,keepaspectratio, trim = 4cm 0cm 5cm 0cm, clip]{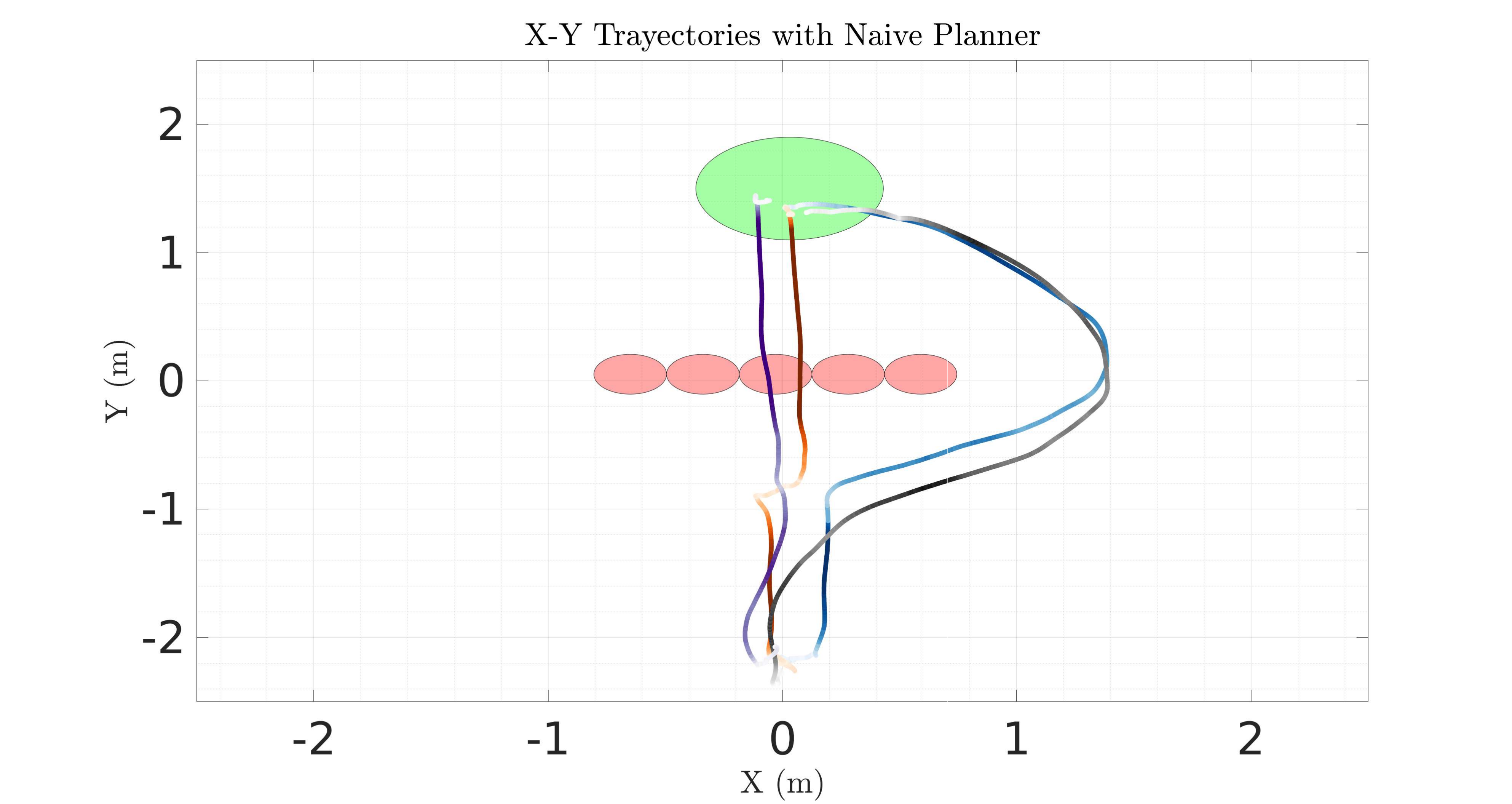}
        \caption{Without global planner (Naive planner).}
        \label{fig:Wall_Naive}
    \end{subfigure}
    \begin{subfigure}{\textwidth}
        \centering
        \includegraphics[width=8cm,keepaspectratio, trim = 3cm 0cm 6cm 0cm, clip]{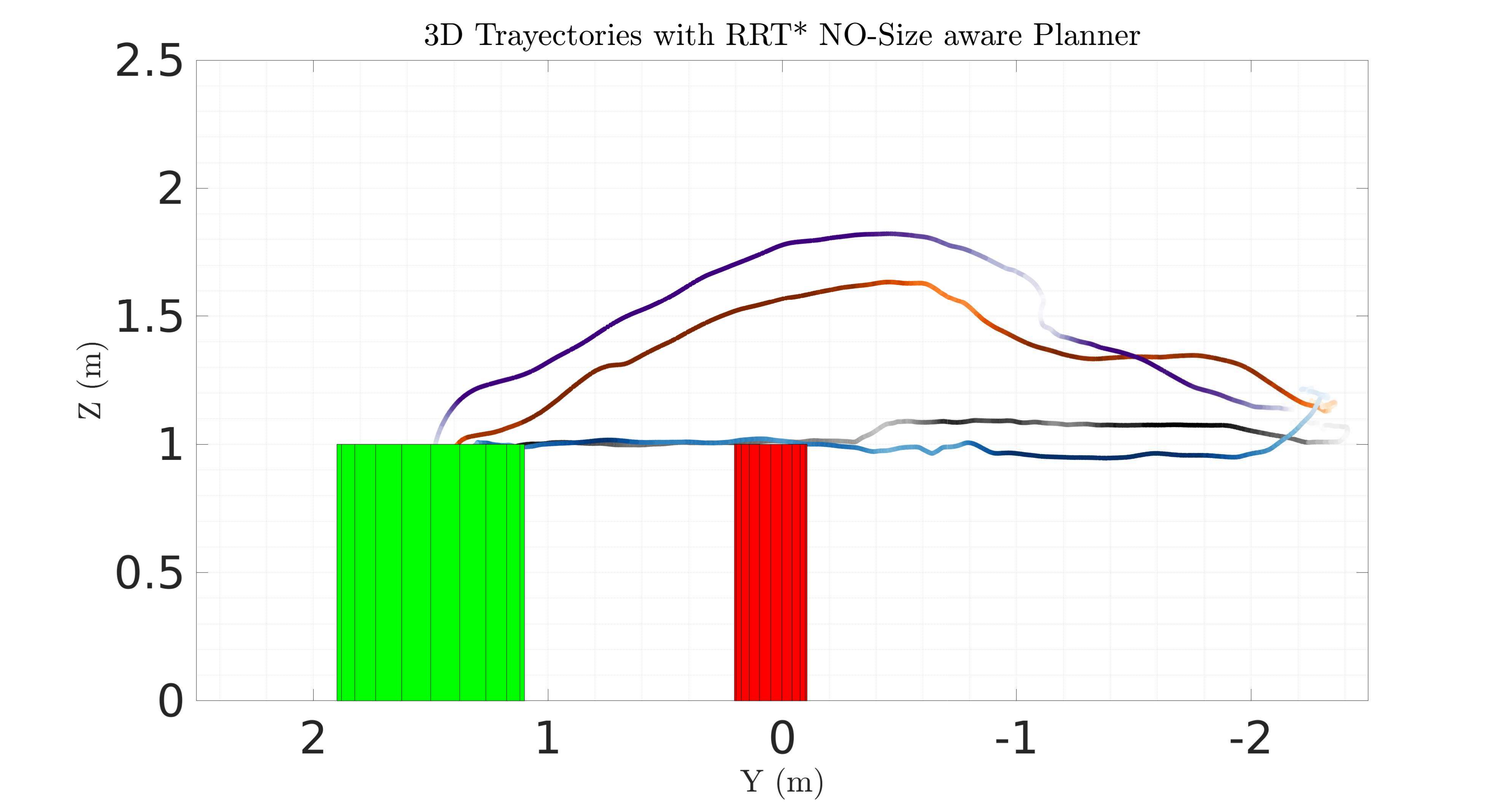}
        \includegraphics[width=8cm,keepaspectratio, trim = 4cm 0cm 5cm 0cm, clip]{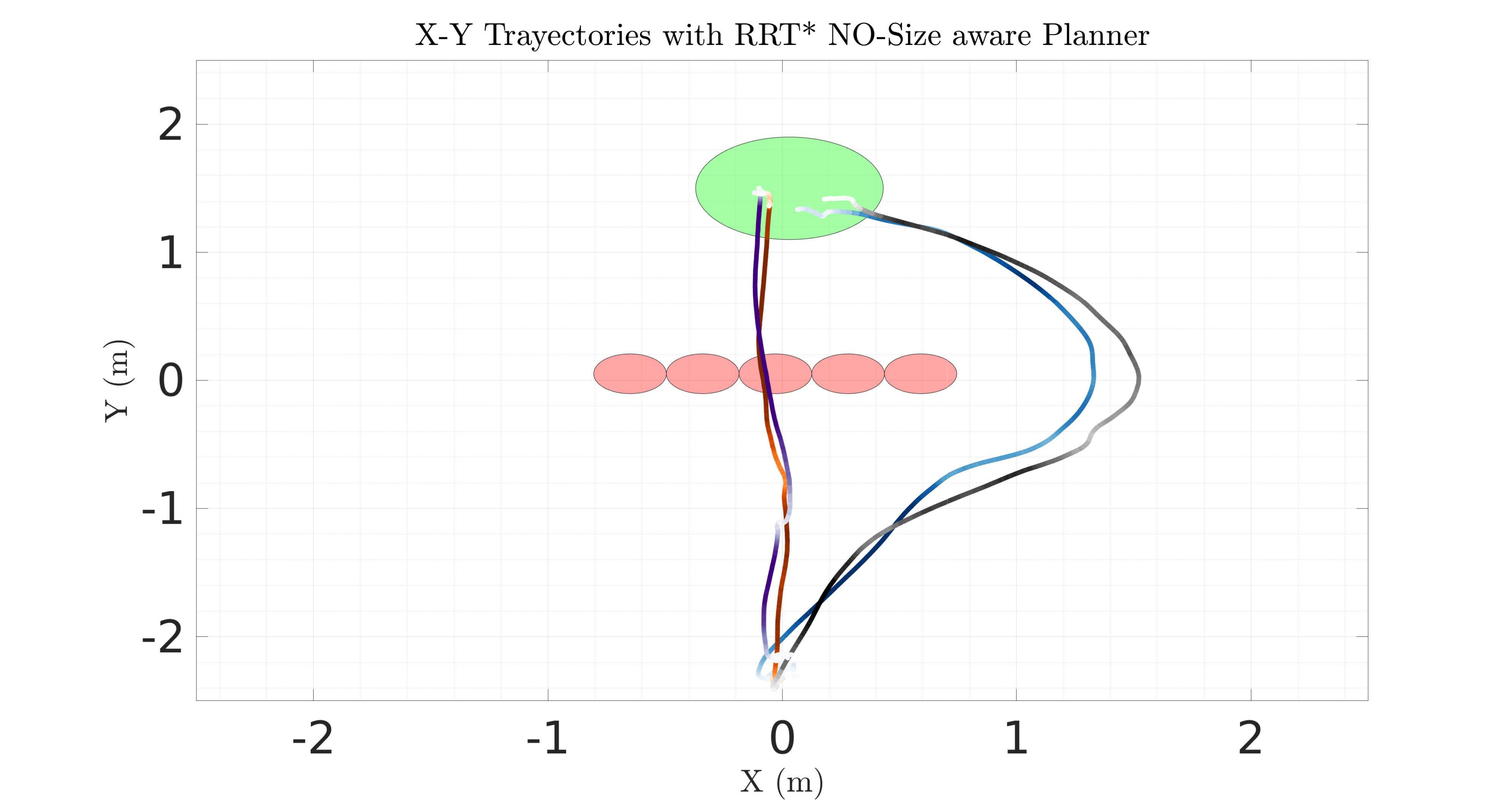}
        \caption{RRT* NOT size aware.}
        \label{fig:Wall_RRT_NO_SIZE}
    \end{subfigure}
    \hfill
    \begin{subfigure}{\textwidth}
        \centering
        \includegraphics[width=8cm,keepaspectratio, trim = 3cm 0cm 6cm 0cm, clip]{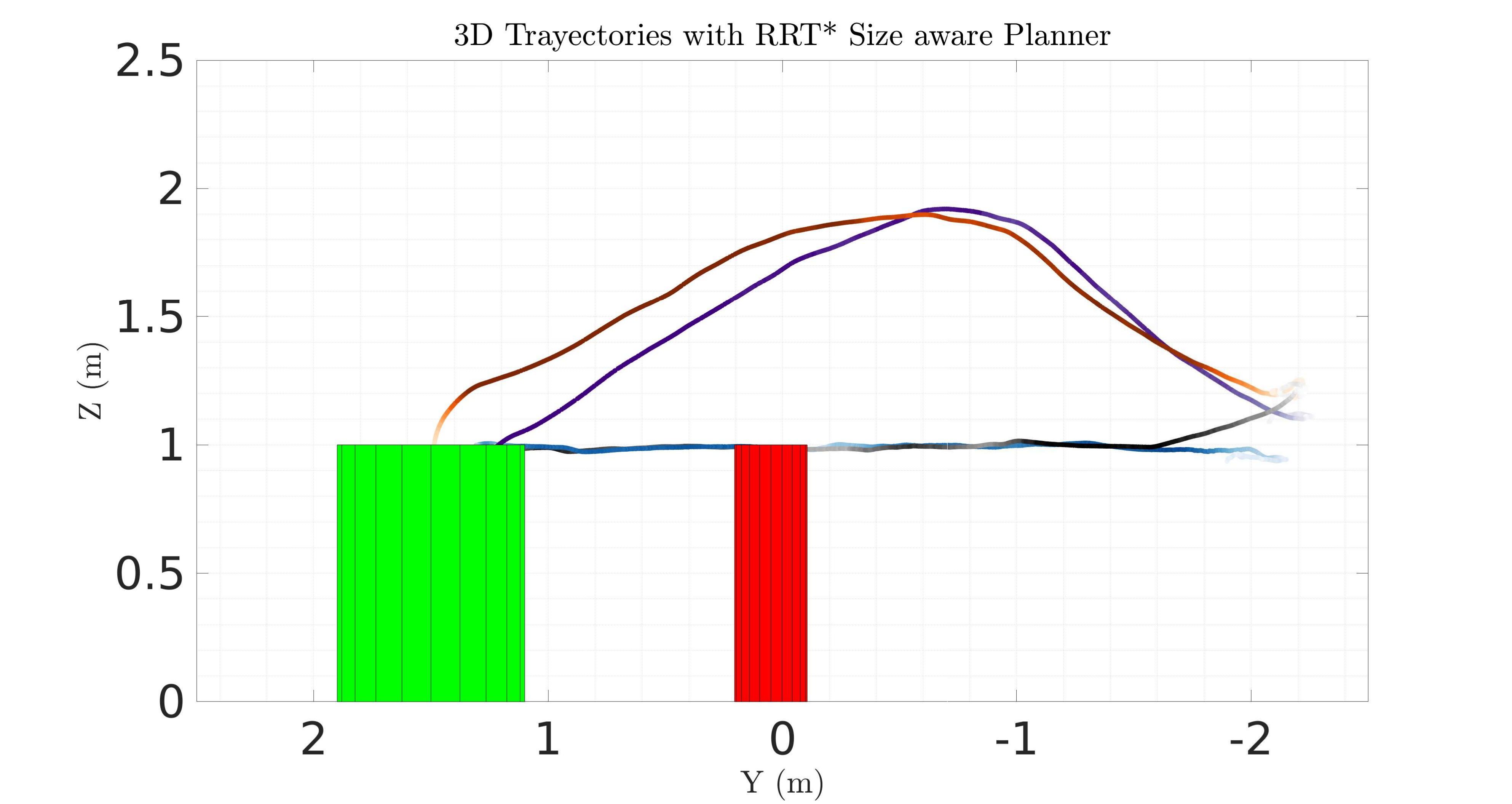}
        \includegraphics[width=8cm,keepaspectratio, trim = 4cm 0cm 5cm 0cm, clip]{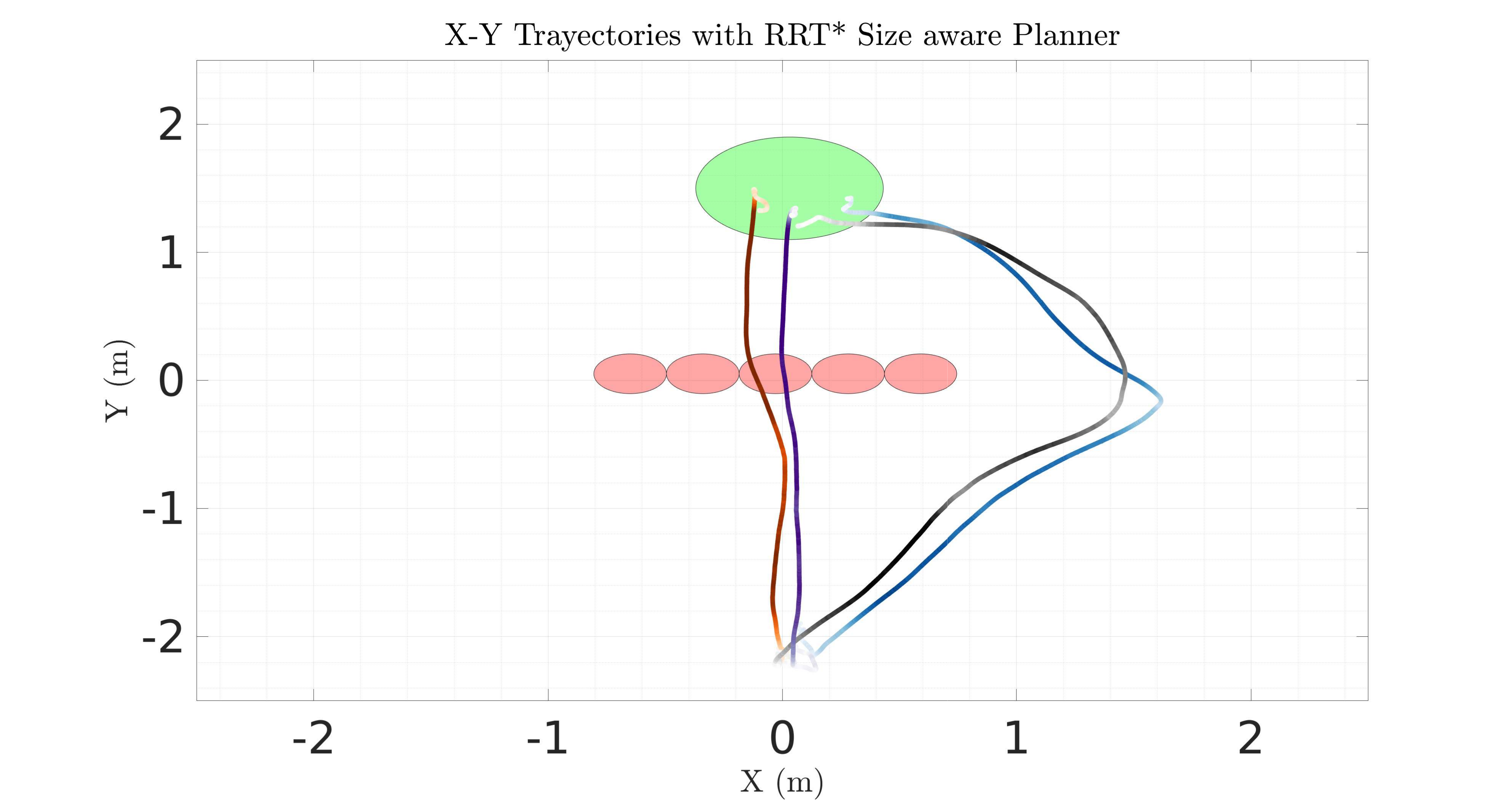}
        \caption{RRT* size aware.}
        \label{fig:Wall_RRT_SIZE}
    \end{subfigure}
    \caption{Lateral and Top views for the trajectories performed in the \textit{Wall Scenario} under $r_{search}$, $K_{\psi}$ and $K_z$ combinations proposed in \autoref{tab:DWA_parameters}. Left, $K_{\psi}=0.8, K_z=0.2$ (purple and orange), keeping orientation is prioritized, therefore a vertical avoidance is executed; right, $K_{\psi}=0.2, K_z=0.8$ (blue and black), keeping orientation is relaxed, therefore a lateral avoidance is allowed. The color intensity of each trace corresponds to the forward velocity $v_x$. Red cylinders are obstacles and the green one is the goal. The faster the UAV goes the darker the trace is.}
\end{figure*}

In every situation, DWA-3D makes the best final decision for maneuvering around the obstacles, following or not the global path, according with the safety criterium represented in the maximized objective function.

Those configurations have faced scenarios with specific challenges to remark DWA-3D performance. Additionally, we checked the capability of changing the avoidance preferences by tunning the parameters and its impact on the navigation. For the experiments carried out in the "Drone Arena", due to its relatively reduced dimensions we decided to not allow the global planner to replan once it has provided the first solution. 

\subsubsection{Wall Scenario}
The goal and the drone are separated by a wall of $1m \times 1.5m \times 0.3m$ $(H \times L \times W)$ that can be avoided by flying over it or by passing by any of its sides. Although this scenario may not involve a challenge for an autonomous navigation system, it is the perfect situation to show the changes produced by variations in parameters. 
\autoref{fig:Wall_Naive} contains 4 trajectories (safer or riskier; lateral or vertical avoidance) following the naive plan while changing $K_z$,  $K_{\psi}$ and $r_{search}$ according to \autoref{tab:DWA_parameters}. It can be seen that for $Kz > K_{\psi}$, lateral avoidance is preferred because orientation to the goal is enforced to be kept; otherwise, vertical is. Moreover, it is manifested that the lower $r_{search}$ is, the later is produced the reaction to the obstacle, as it is considered at a lower distance. 

Then, the same flights are performed with a global planner that does not considers the drone's size, \autoref{fig:Wall_RRT_NO_SIZE}. It can be seen that after adding the global planner, the obstacle evasion starts earlier than before, no matter the value of $r_{search}$. Nevertheless, as those plans tend to lead to lateral collision (UAV's dimensions were not considered), DWA-3D must react and avoid them. That phenomena is reflected as a \textit{two-steps} avoidance, the first and inaccurate one performed by the global planner and the one that ensures the safe operation due to DWA-3D intervention. Is in this second step were the effect of the $r_{search}$ value influences.

Finally, the global planner is configured to consider the drone's size, computing safer paths, \autoref{fig:Wall_RRT_SIZE}. Although now the difference of selecting more conservative or more riskier $r_{search}$ value is subtle, we can still observe two phases during obstacle avoidance and that the second one is influenced by it. While the global path is within the known region, the UAV just have to follow it, as it is safe. Once it arrives to a region that was unknown when the path was computed, the effect $r_{search}$ value is translated as a faster and less conservative or slower and more careful approach to the goal.

\subsubsection{Narrow Gaps Scenario}
\begin{figure}[h!]
    \centering
    \begin{subfigure}{\columnwidth}
        \centering
        \includegraphics[width=\linewidth,keepaspectratio, trim=14cm 6cm 17cm 4cm, clip]{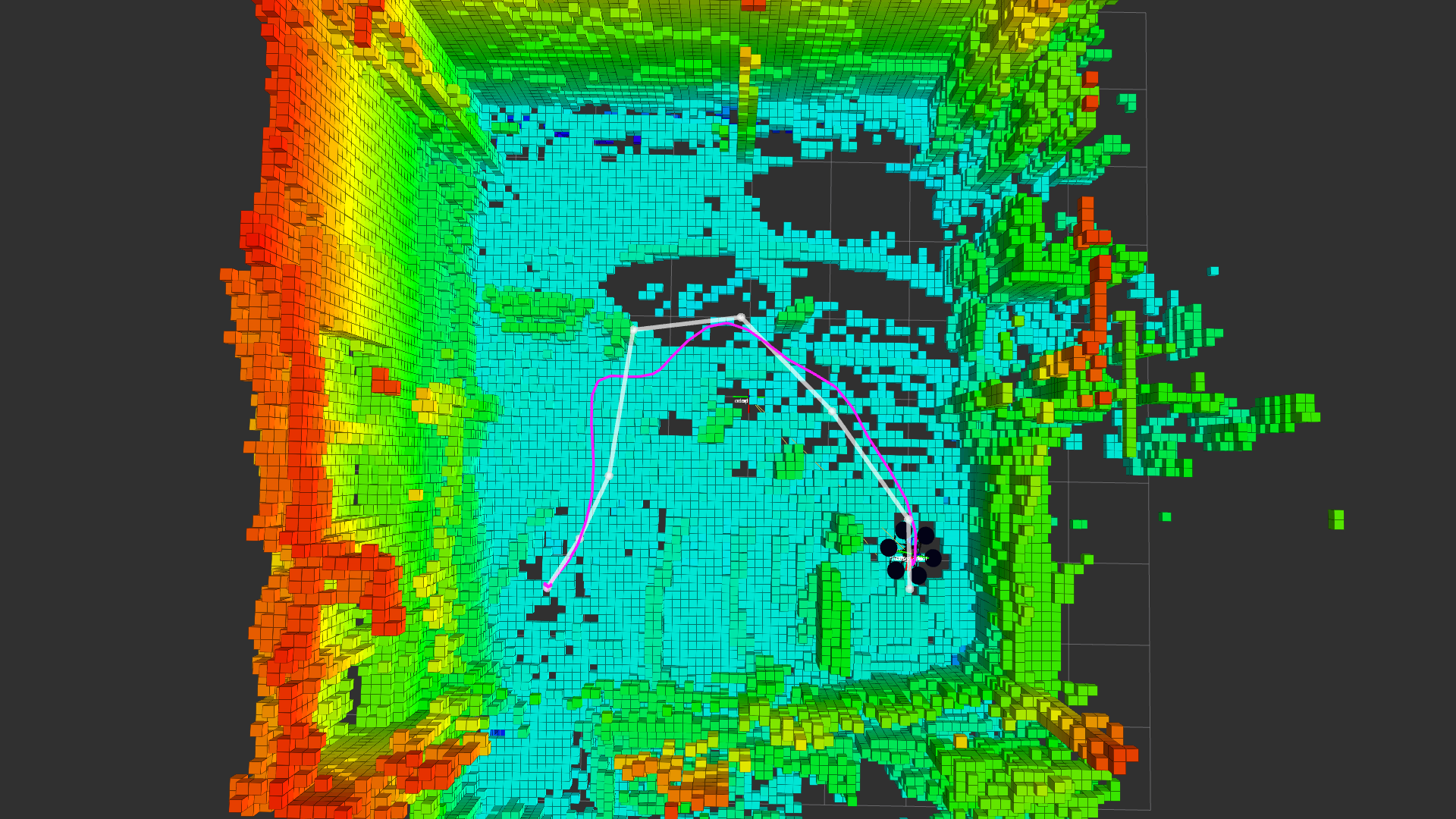}
        \caption{Performed trajectory (purple) and global plan (white). Note in the top figure the Octomap had unknown areas when the global plan was computed at the beginning and it does not change (replan disabled).}
        \label{fig:NarrowGaps1}
    \end{subfigure}

    \begin{subfigure}{\columnwidth}
        \centering
        \includegraphics[width=\linewidth,keepaspectratio, trim = 5cm 0cm 6cm 0.25cm, clip]{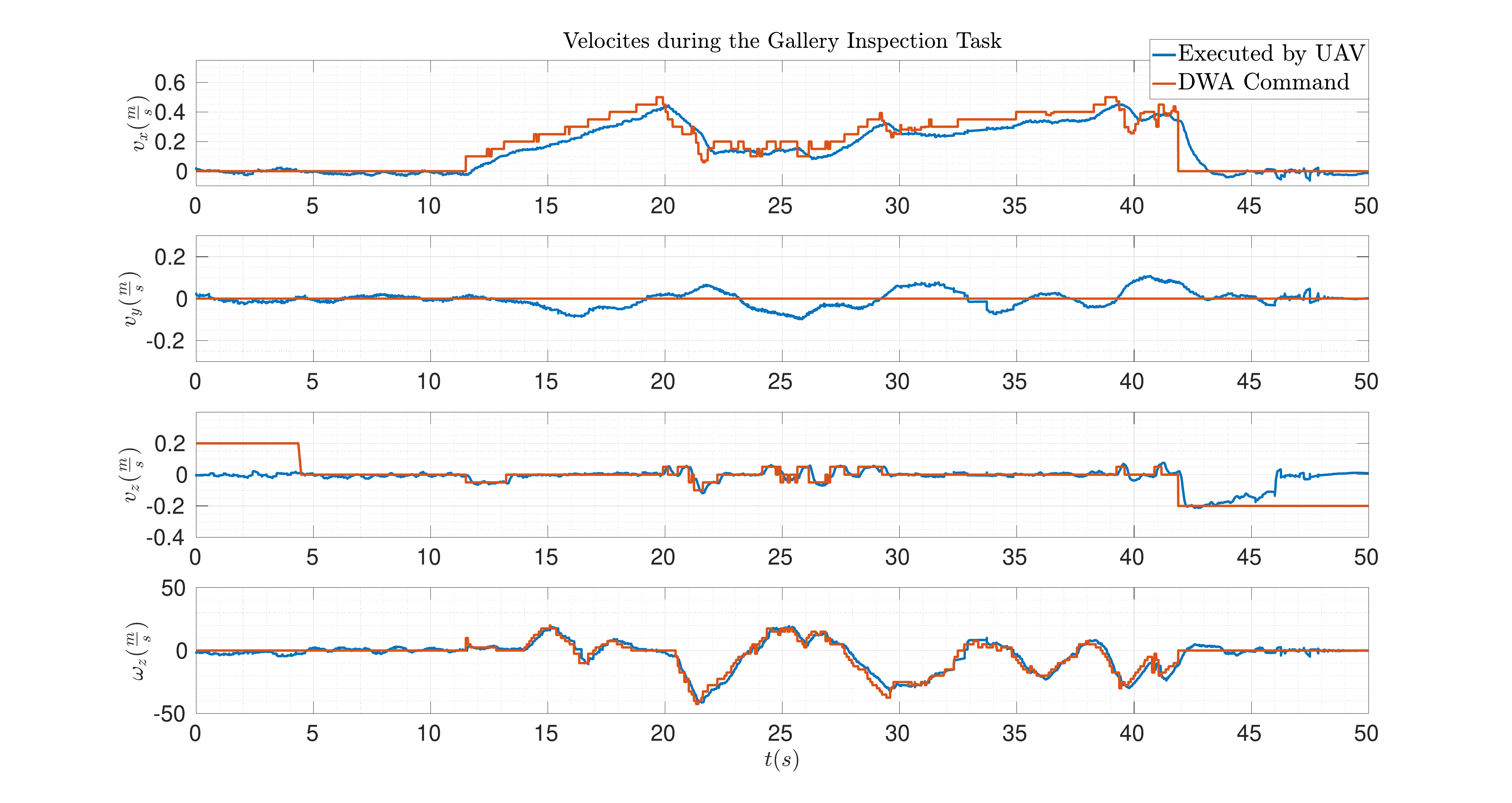}
        \caption{Velocities profiles during a flight in the \textit{Narrow Gaps Scenario} with $v_{x}^{max} = 0.75 \frac{m}{s}$. In red, the one commanded by the local planner (DWA-3D); in blue the ones executed by the UAV.}
        \label{fig:NarrowGapsVelocities}
    \end{subfigure}
    \caption{Flight in the \textit{Narrow Gaps Scenario} (top) and velocities (desired and executed) during it (bottom). {\color{blue}\href{https://www.youtube.com/watch?v=6-wjs-jtENo}{VIDEO}}}
    \label{fig:NarrowGaps}
\end{figure}

In this scenario, the hexarotor must fly while overcoming sparse but nearby obstacles, by traversing through narrow gaps ($1.25m \sim 1.35m$), having about 25 cm of clearance  on each side in the worst case. Additionally, the maximum allowed $v_x$ have been increased to $0.75 \frac{m}{s}$. 
The trajectory executed (purple in trace \autoref{fig:NarrowGaps1}) compared with the global path (white) leverages DWA-3D's capabilities, avoiding an obstacle that had not been observed when computing the plan. The dodging maneuver is reflected in the velocities profiles (\autoref{fig:NarrowGapsVelocities}) with a strong right rotation coupled with a deceleration around $t \sim 21s$. Later, the UAV performs additional but softer avoidance maneuvers while traversing narrow gaps towards the final destination. During these last maneuvers the drone does not need to slow down, it even accelerates up to $0.4 \frac{m}{s}$. 

\subsubsection{Rings}
\begin{figure}[h!]
    \centering
    \begin{subfigure}{\columnwidth}
    \centering
        \includegraphics[width=\linewidth, keepaspectratio, trim= 0cm 0cm 0cm 3cm, clip]{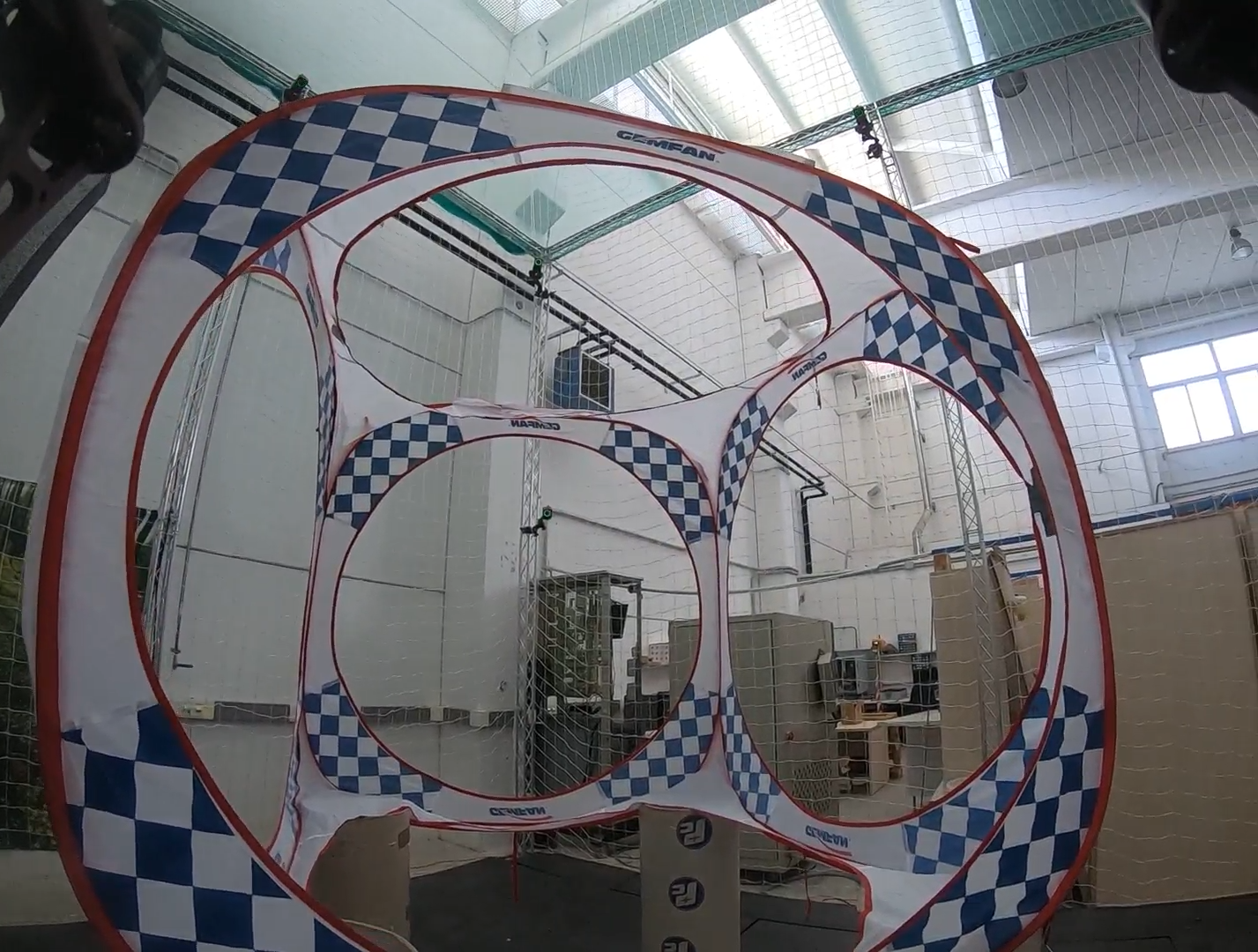}
        \caption{Drone's point of view while traversing the rings.}
    \vspace{0.13cm}
        \label{fig:Ring_real}
    \end{subfigure}
    \begin{subfigure}{0.4\columnwidth}
        \centering
        \includegraphics[height=3.75cm,keepaspectratio]{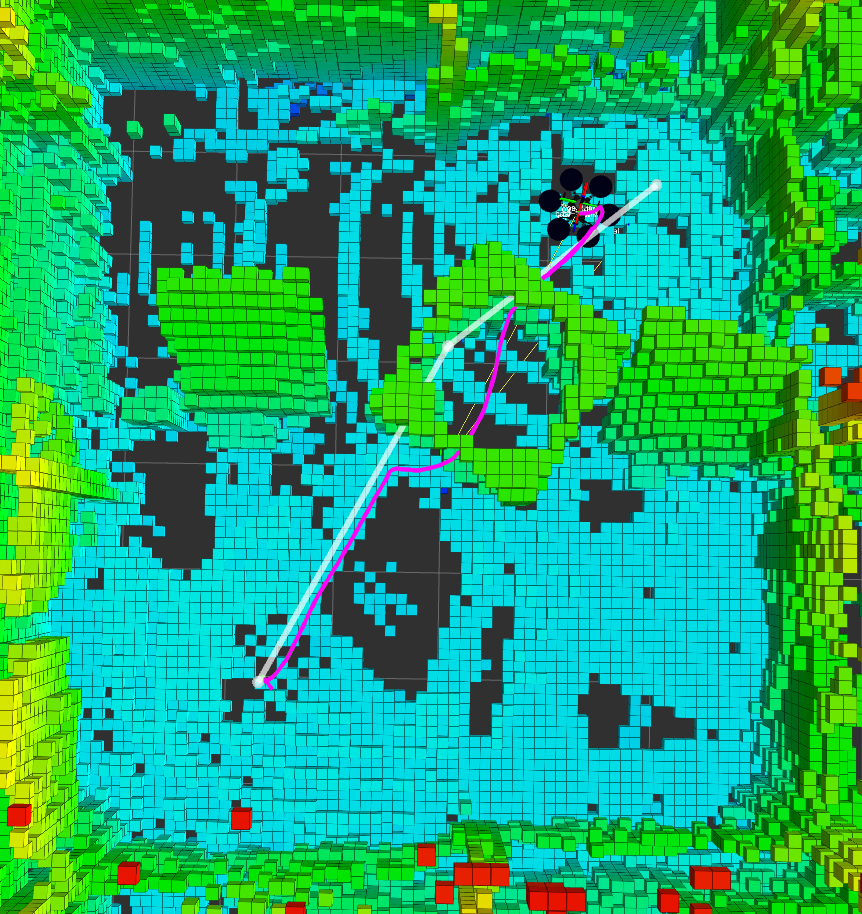}
        \caption{Pass-through flight, traversing two parallel rings. {\color{blue}\href{https://www.youtube.com/watch?v=Q2bgwkcf2P0}{VIDEO}}.}
        \label{fig:Rings1}
    \end{subfigure}
    \hfill
    \begin{subfigure}{0.5\columnwidth}
        \centering
        \includegraphics[height=3.75cm,keepaspectratio]{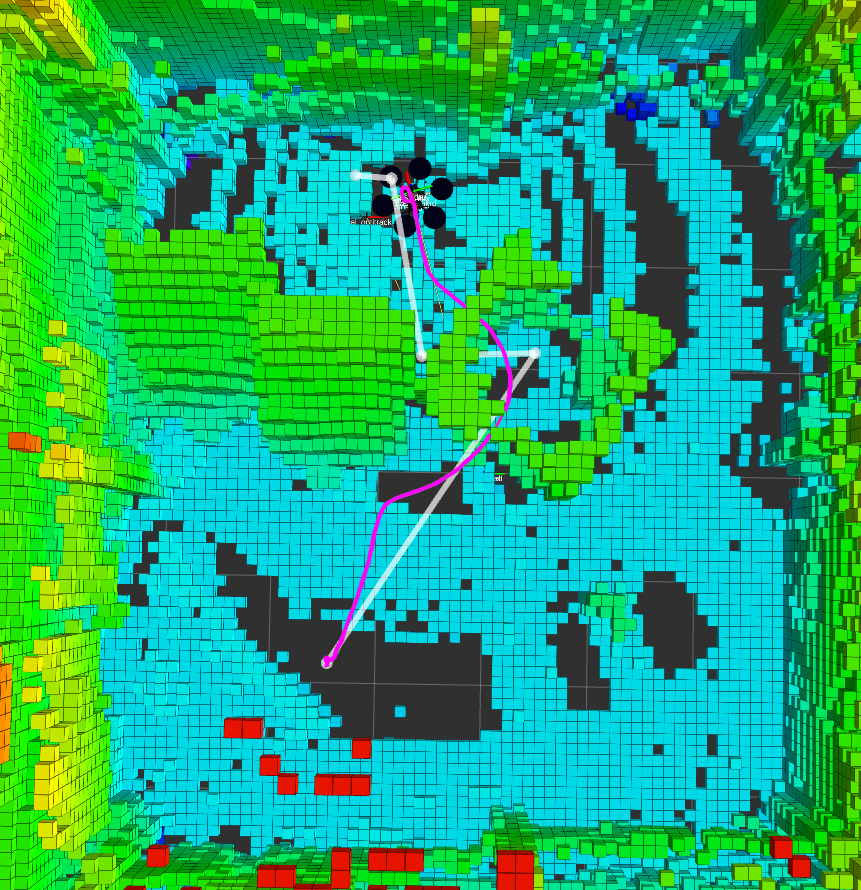}
        \caption{$90^{\circ}$ between rings flight, leaving through the ring at the left. {\color{blue}\href{https://www.youtube.com/watch?v=OsXvbI0YvC0}{VIDEO}}.}
        \label{fig:Rings2}
    \end{subfigure}
    \hfill
    \caption{Top: UAV in the \textit{Rings Scenario}. Bottom: Trajectories executed (purple) and global planner path (white) in rings scenario, lateral avoidance selected and $r_{search} = 1m$.}
    \label{fig:Rings}
\end{figure}

In order to exploit the drone maneuvering capabilities, a more complex navigation scenario through rings is proposed, see \autoref{fig:Ring_real}. Two variations were tested: a straight pass-through (\autoref{fig:Rings1}) and a side exit through a lateral ring (\autoref{fig:Rings2}). 
Here RRT* safety distance has been reduced to $0.2m$ instead of $0.5m$ to enhance paths passing inside the ring as tolerance here is tighter. The drone follows the planned paths between subgoals when possible, keeping the safety distance.

\subsubsection{Moving obstacle}
\begin{figure}
    \centering
    \begin{subfigure}{0.4\columnwidth}
        \centering
        \includegraphics[height=3.85cm,keepaspectratio]{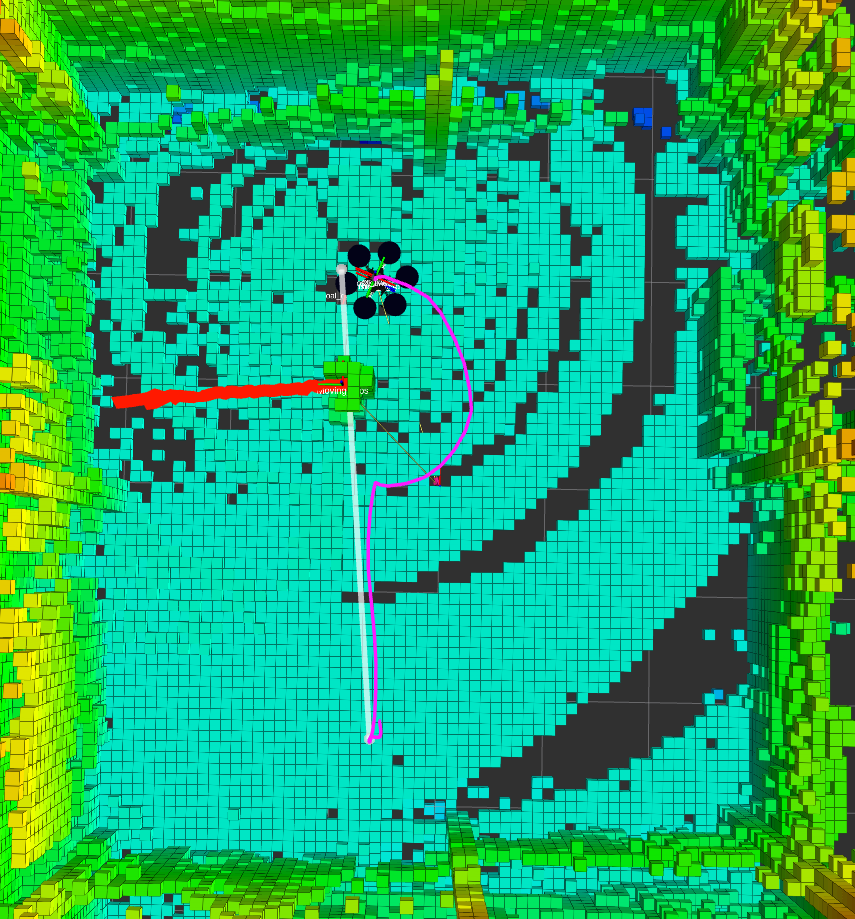}
        \caption{The UGV stops once it has interrupted the UAV path. The drone avoids  the new obstacle. {\color{blue}\href{https://www.youtube.com/watch?v=hhF1UDDtVf0}{VIDEO}}.}
        \label{fig:MovingObs1}
    \end{subfigure}
    \hfill
    \begin{subfigure}{0.5\columnwidth}
        \centering
        \includegraphics[height=3.85cm,keepaspectratio]{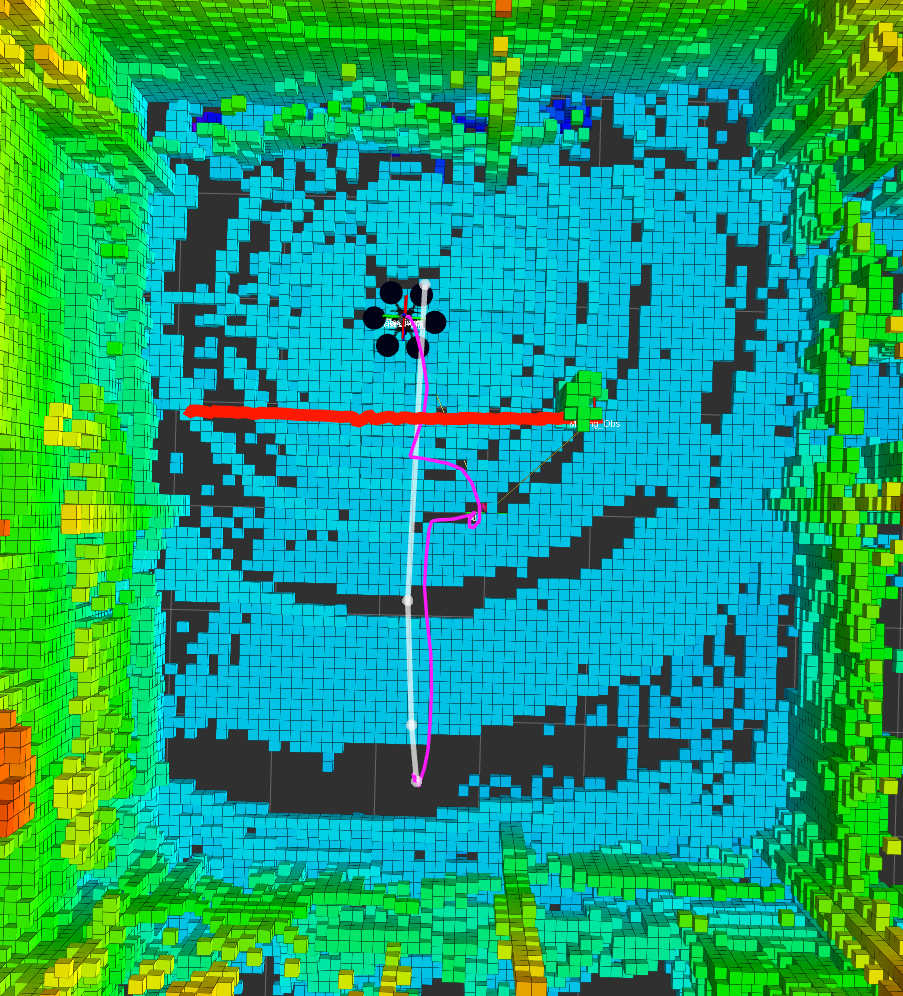}
        \caption{UGV keeps moving and blocks the UAV movement for a longer time until the room is opened again, then the drone keeps its path. {\color{blue}\href{https://www.youtube.com/watch?v=UpeWJ9jL2-w}{VIDEO}}.}
        \label{fig:MovingObs2}
    \end{subfigure}
    \hfill
    \caption{Trajectories executed by the UAV (purple) and global planner path (white) computed when the UGV is in the left part of the red trajectory shown, and so the room to the goal is free. }
    \label{fig:MovingObs}
\end{figure}

Finally, although DWA-3D is not tailored for highly dynamic scenarios, we want to explore the capability of dodging moving obstacles. It has been studied by interrupting the trajectory of the UAV with a TurtleBot 2 platform with a cylinder tied to its top plate. Initially, the UGV roomba remains static and allows the drone's global planner to compute a path. Once the hexarotor has a plan and starts following it, the roomba moves at a constant speed ($0.3 \frac{m}{s}$), enforcing the DWA-3D to react and avoid it, see \autoref{fig:MovingObs}. Two cases are represented, in the first one the UGV stops once it cuts the UAV trajectory, then avoiding  laterally the obstacle; while in the second one it (UGV) keeps moving forward, the drone tries to avoid it laterally as in the first case, but finally it finds room to pass  behind the UGV. A lateral avoidance is selected and $r_{search} = 1.5m$. During the experiments in this scenario it has been noticed that Octomap does not update the dynamic changes in the scene at the moment. As the occupancy map is updated regarding the occupancy probabilities given the sensor measurements and the previous information, those changes in the scene take some time to be considered as real and not only as sensor noise. This involves a noticeable delay while refreshing the position of the UGV, thus taking more time and maneuvers to avoid.

\subsection{Localization Estimation Error}
\label{sec:ErrorLocalization}
\begin{figure}[h]
\hspace*{-0.75cm}
    \centering
        \includegraphics[width=1.1\linewidth,keepaspectratio, trim = 4.5cm 1.5cm 5.5cm 0cm, clip]{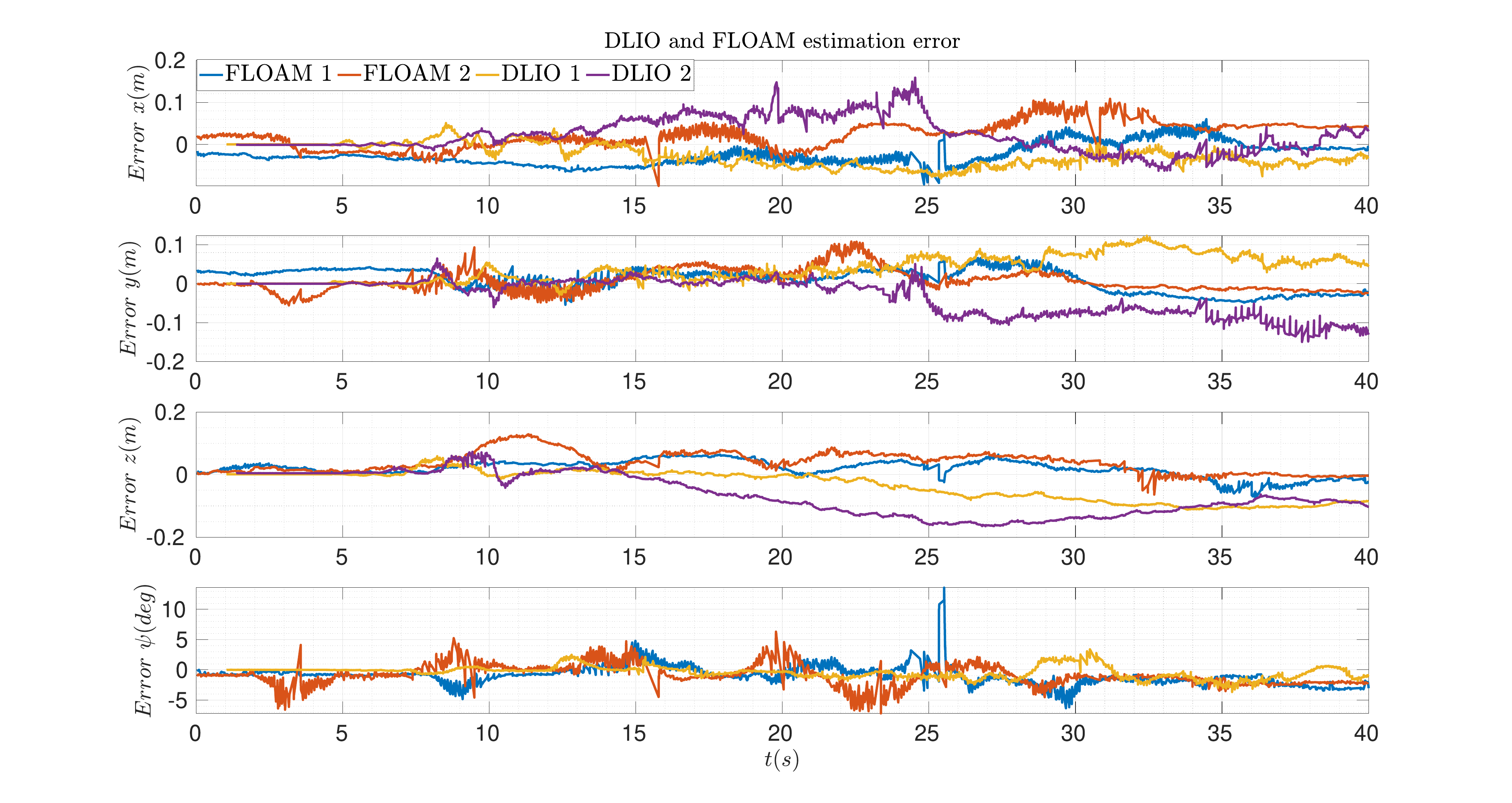}
    \caption{Localization estimation error committed by F-LOAM with our height estimate correction and DLIO with respect to the motion capture system in our facilities. Two consecutive flights in scenarios populated by obstacles were executed. Our "Drone Arena" involves an additional challenge for these localization solutions because of the sparsity and the lack of features that safety net presents.}
    \label{fig:ErrorFLOAM}
\end{figure}

Prior to testing in environments without localization, we evaluated F-LOAM (with our height estimation correction) and DLIO \cite{DLIO}, a LiDAR-Inertial approach, comparing their performance against a motion capture system. The error committed during two consecutive flights with obstacles is shown in \autoref{fig:ErrorFLOAM}. 
While the errors were initially similar, DLIO's error diverged more significantly from the motion capture system's localization towards the end, particularly in the $z$ coordinate, which is crucial for the drone's low level control for a stable flight. F-LOAM's $x$ and $y$ coordinate errors are bounded below the Octomap's voxel size, $0.1m$. Although its $z$ errors were slightly higher, they remained under $0.2m$ and tend to recover from the accumulated drift. Based on these results, we selected F-LOAM as a reliable and stable localization source for the UAV in uncontrolled environments. 

\subsection{Field Experiments}
\label{sec:FieldExperiments}
Once safe and robust navigation has been secured in scenarios where ground-truth localization is given, the system capabilities are demonstrated in larger and more complex scenarios where it must localize itself while navigating. Regarding the size of the scenarios now it is interesting to enable the global planner to compute new solutions at the same time that it discovers new regions. Additionally, the size-aware version of the global planner has been selected for those experiments for smoother trajectories. 

\subsubsection{Warehouse Scenario}
\begin{figure}[h!]
    \centering
    \includegraphics[width=\linewidth,keepaspectratio, trim = 0cm 5cm 0cm 0cm, clip]{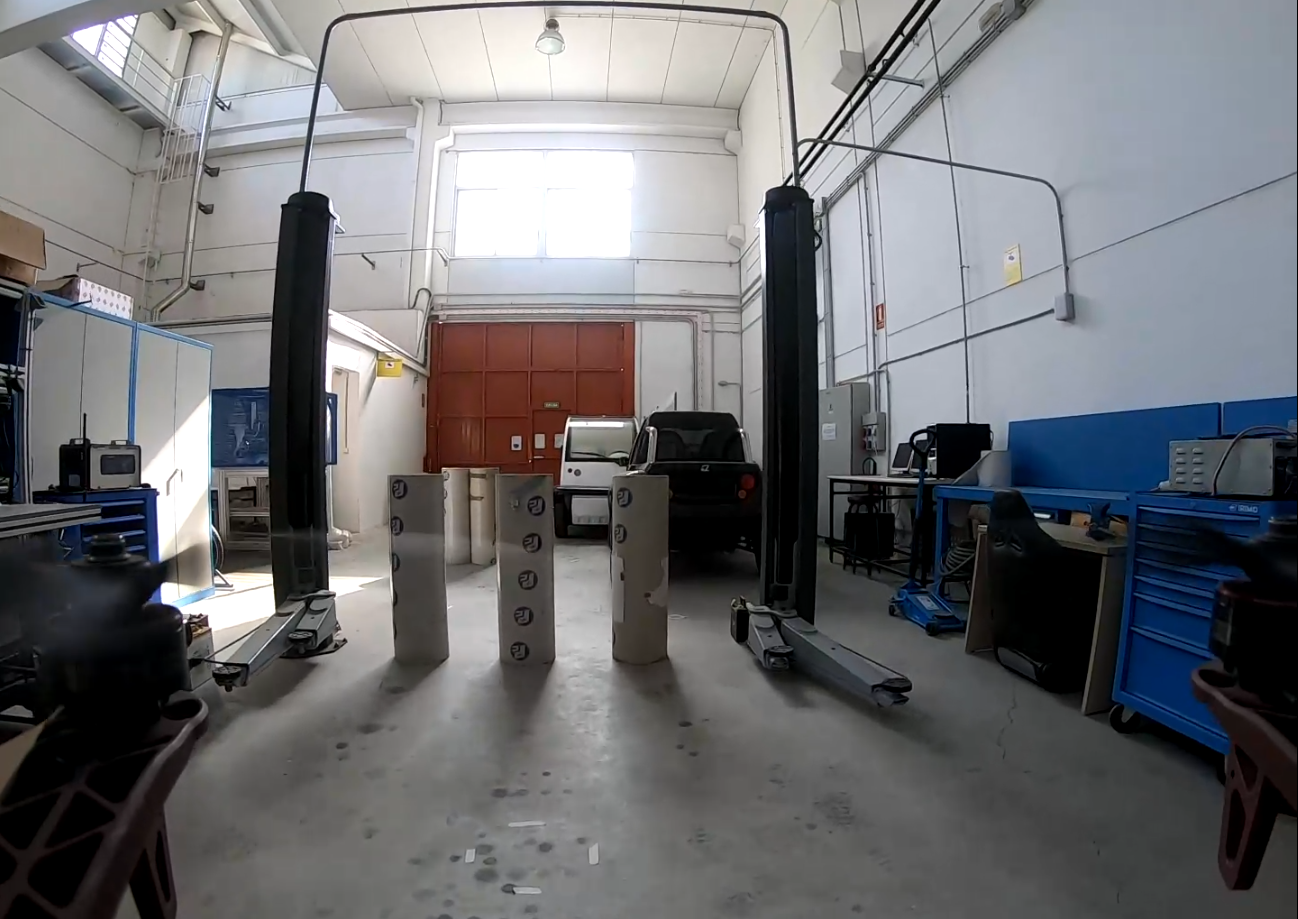}
    \caption{Drone's Point of View during a flight in the Warehouse. The frontal obstacles block the forward motion, thus they have to be avoided above.}
    \label{fig:Warehouse_FPV}
\end{figure}

\begin{figure}[h!]
    \centering
    \begin{subfigure}{\columnwidth}
        \centering
        \includegraphics[width=0.7\linewidth,keepaspectratio, trim=18cm 0cm 18cm 0cm, clip]{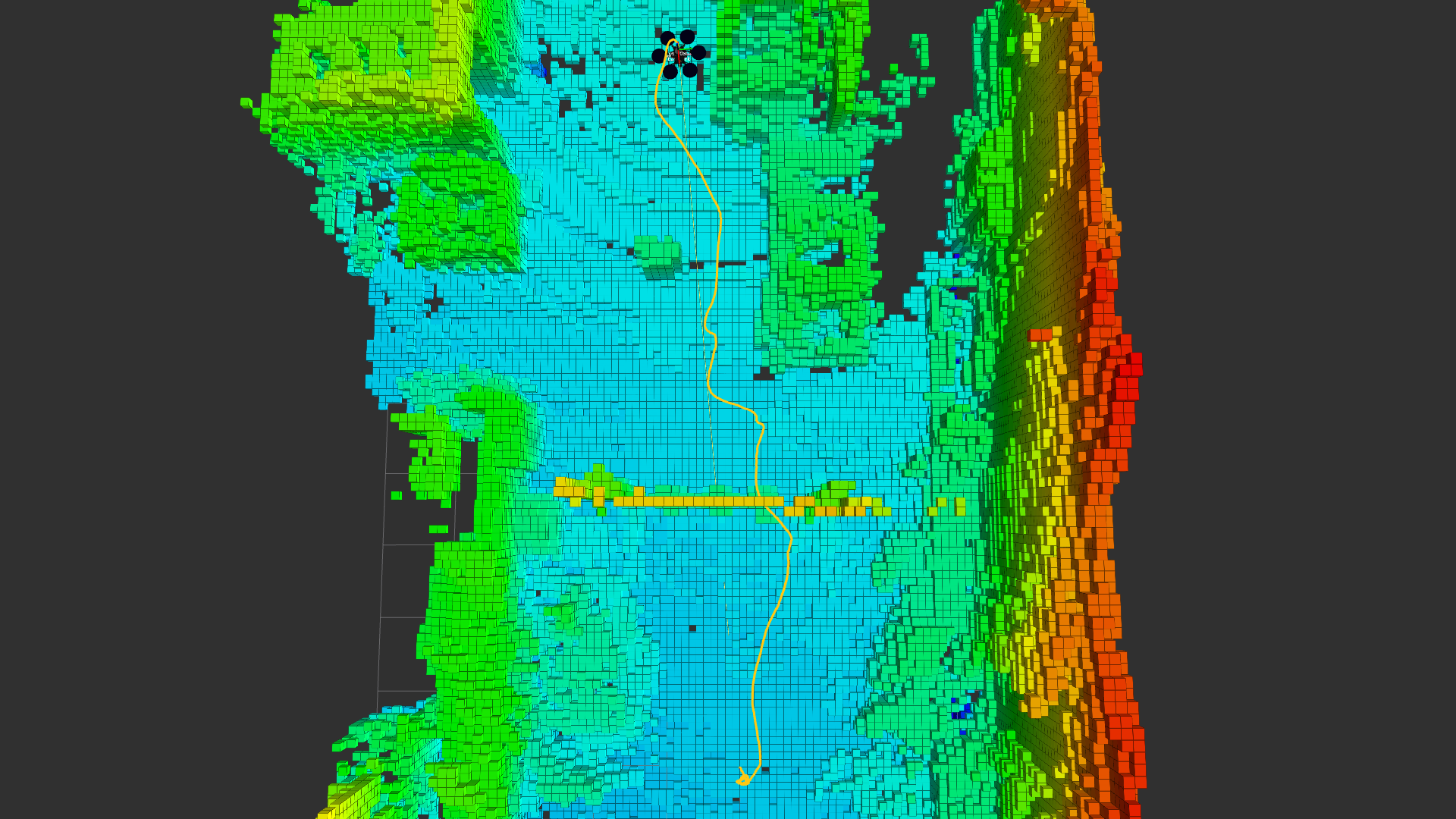}
        \caption{Performed trajectory (orange) in the \textit{Warehouse Scenario}. The first set of obstacles (wall) are sorted flying over them, while the rest are avoided laterally.}
        \label{fig:WarehouseTrajectory}
    \end{subfigure}

    \begin{subfigure}{\columnwidth}
        \centering
        \includegraphics[width=8cm,height=8cm,keepaspectratio, trim = 0cm 0cm 0cm 0cm, clip]{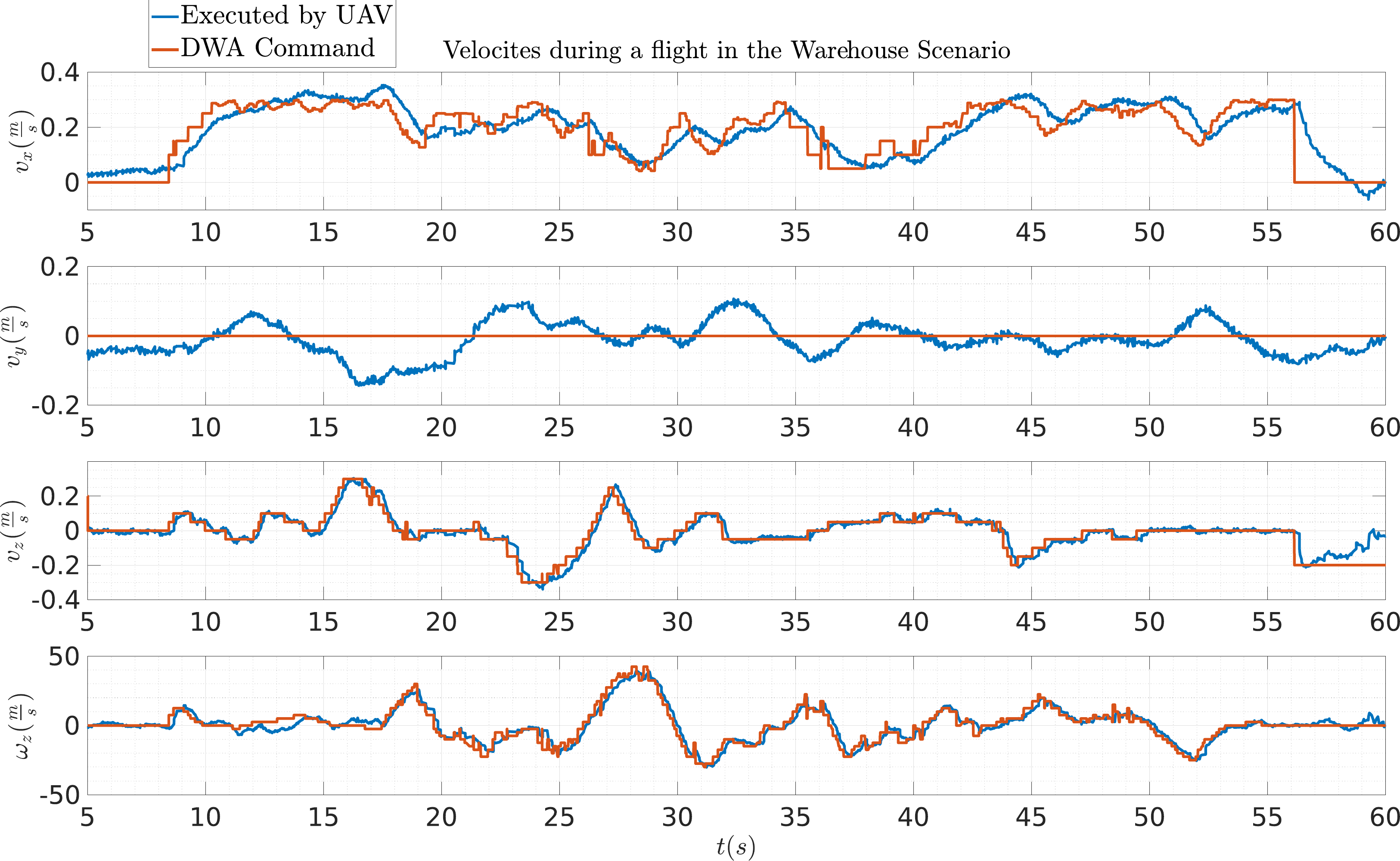}
        \caption{Velocities profiles during a flight in the \textit{Warehouse Scenario} with $v_{x}^{max} = 0.3 \frac{m}{s}$. In red, the one commanded by the local planner (DWA-3D); in blue the ones executed by the UAV.}
        \label{fig:WarehouseVelocities}
    \end{subfigure}
    \caption{Flight in the \textit{Warehouse Scenario} (top) and velocities (desired and executed) during it (bottom). {\color{blue}\href{https://youtu.be/pIDRoKn-rb0}{VIDEO}}}
    \label{fig:Warehouse}
\end{figure}

In this scenario, the UAV has to cross a warehouse, avoiding multiple obstacles that are in the way, see \autoref{fig:Warehouse_FPV}. Most of them are hidden at the beginning and must be managed once they are encountered. First, the drone must fly over a set of obstacles that block the forward motion and later it must cross thought a narrow gap in an obstacle-crowded area to reach its destination, see \autoref{fig:WarehouseTrajectory} for more information about the performed trajectory. The velocity profiles in \autoref{fig:WarehouseVelocities} show the avoidance flying over the first set of obstacles around $t \sim 17s$, followed by a return to the desired height, $t \sim 24s$. Then, the UAV performs a left turn, followed by another one to the right to face the narrow gap that it must traverse to avoid the last set of obstacles.  

\begin{figure}[t]
    \centering
    \begin{subfigure}{\columnwidth}
        \centering
        \includegraphics[width=0.9\linewidth]{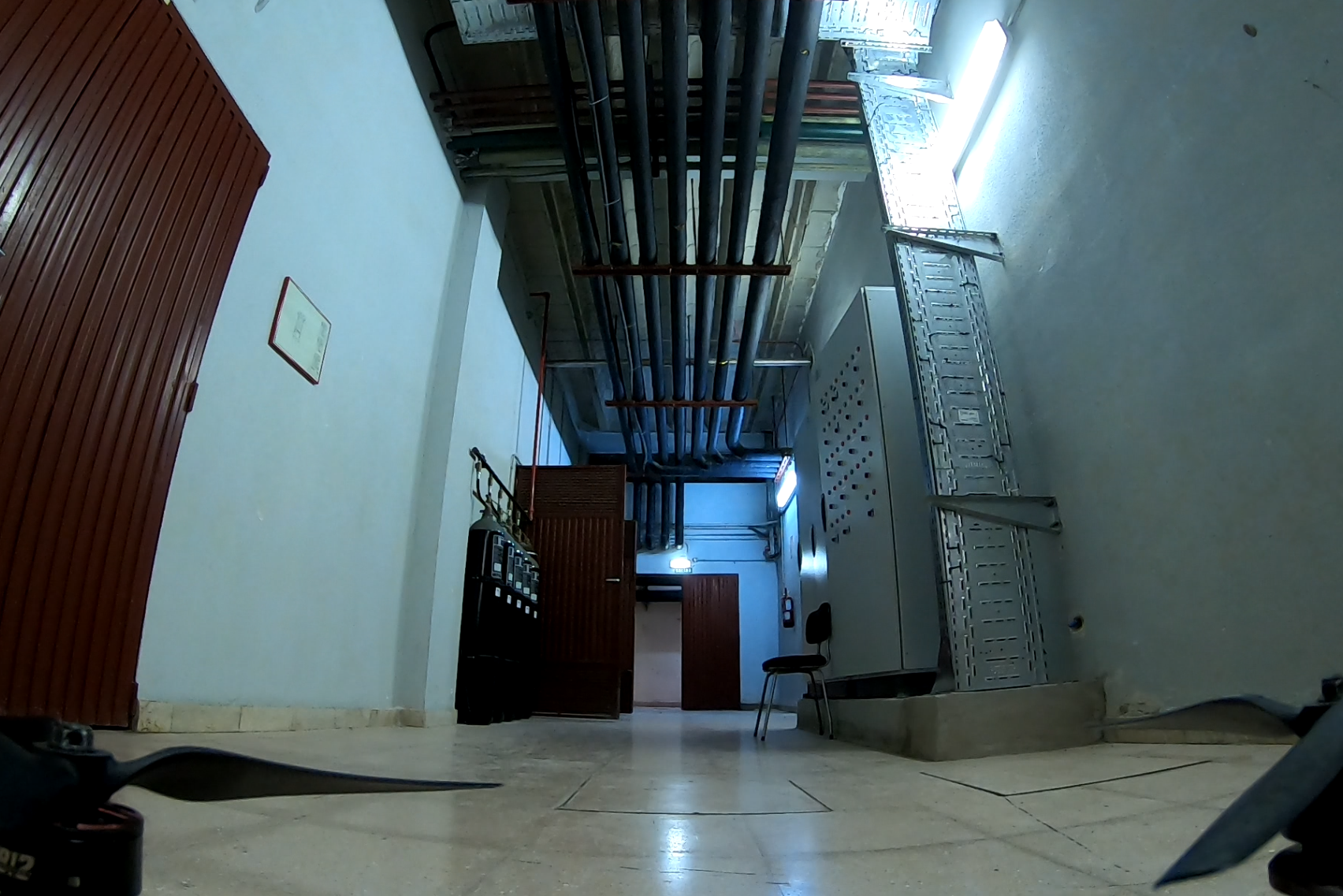}
        \caption{Before taking off in \textit{Gallery Scenario.} The left room entrance is occluded by the opened door.}
        \label{fig:BasementFPVInitial}
    \end{subfigure}
    \begin{subfigure}{\columnwidth}
        \centering
        \includegraphics[width=0.9\linewidth,keepaspectratio, trim=0cm 0cm 0cm 2.5cm, clip]{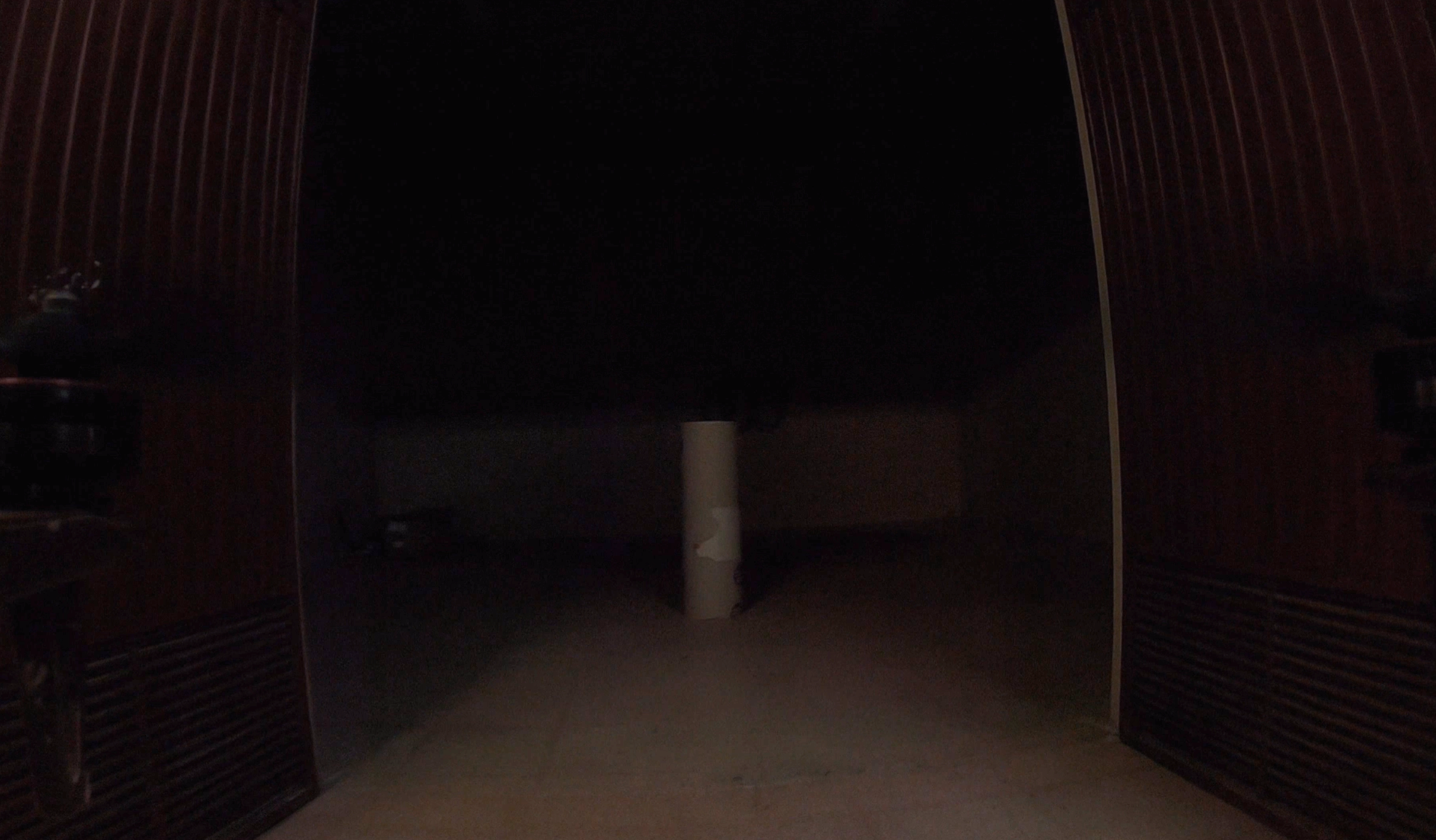}
        \caption{Entering the target room in \textit{Gallery Scenario.} The poor visibility conditions make impossible to rely on a visual system neither for localization nor obstacle detection.}
        \label{fig:BasementFPVSala}
    \end{subfigure}
    \hfill
    \caption{Drone's Point of View images captured in the \textit{Gallery Scenario}.}
    \label{fig:Basement_FPV}
\end{figure}

\begin{figure}[h!]
    \centering
    \begin{subfigure}{\columnwidth}
        \centering
        \includegraphics[width=0.9\linewidth,keepaspectratio, trim=0cm 2cm 0cm 0cm, clip]{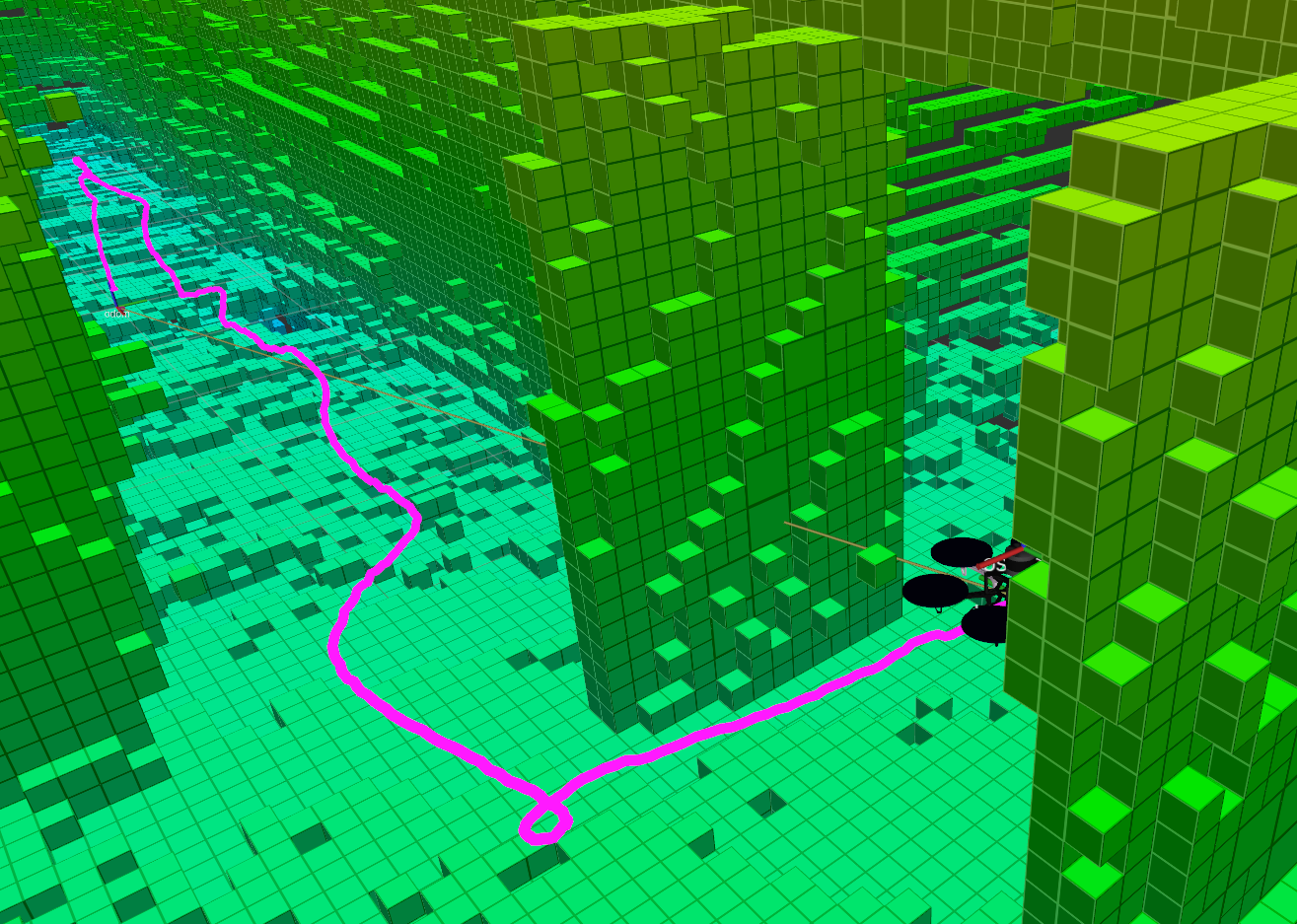}
        \caption{Performed trajectory (purple) during the approach to the room of interest during the gallery inspection task. Note that the problem exposed in \autoref{fig:BlockedDoor} due to $z$ drift is solved by the height estimation correction we added.}
        \label{fig:BasementTrajectoryPart1}
    \end{subfigure}
    \begin{subfigure}{\columnwidth}
        \centering
        \includegraphics[width=0.9\linewidth,keepaspectratio, trim=2cm 0cm 0cm 0cm, clip]{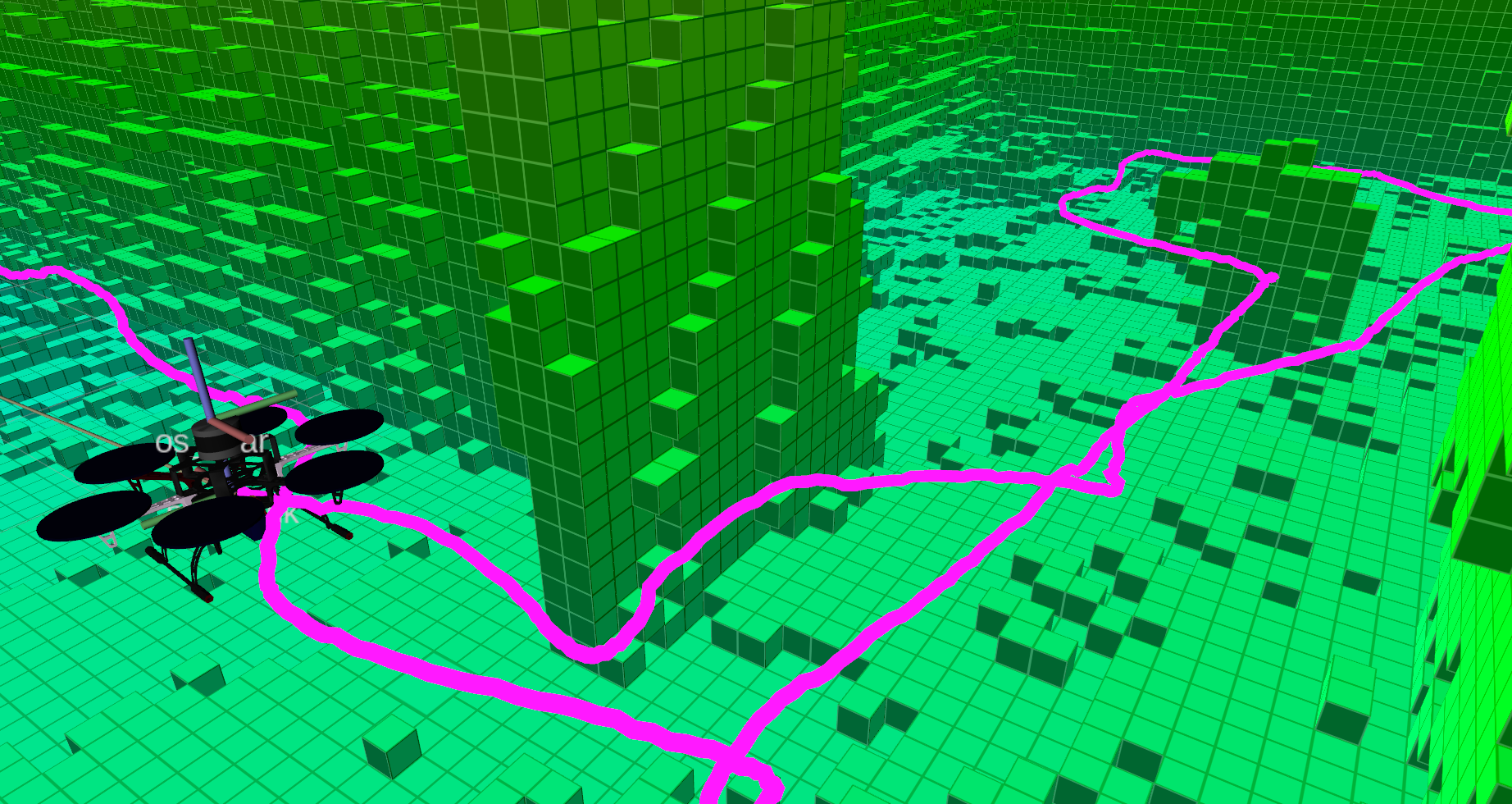}
        \caption{Performed trajectory (purple) to avoid the unexpected obstacle, inspect the room and leave it in the gallery mission.}
        \label{fig:BasementTrajectoryPart2}
    \end{subfigure}
    \begin{subfigure}{\columnwidth}
        \centering
        \includegraphics[width=8cm,height=8cm,keepaspectratio, trim = 5cm 0cm 5cm 0cm, clip]{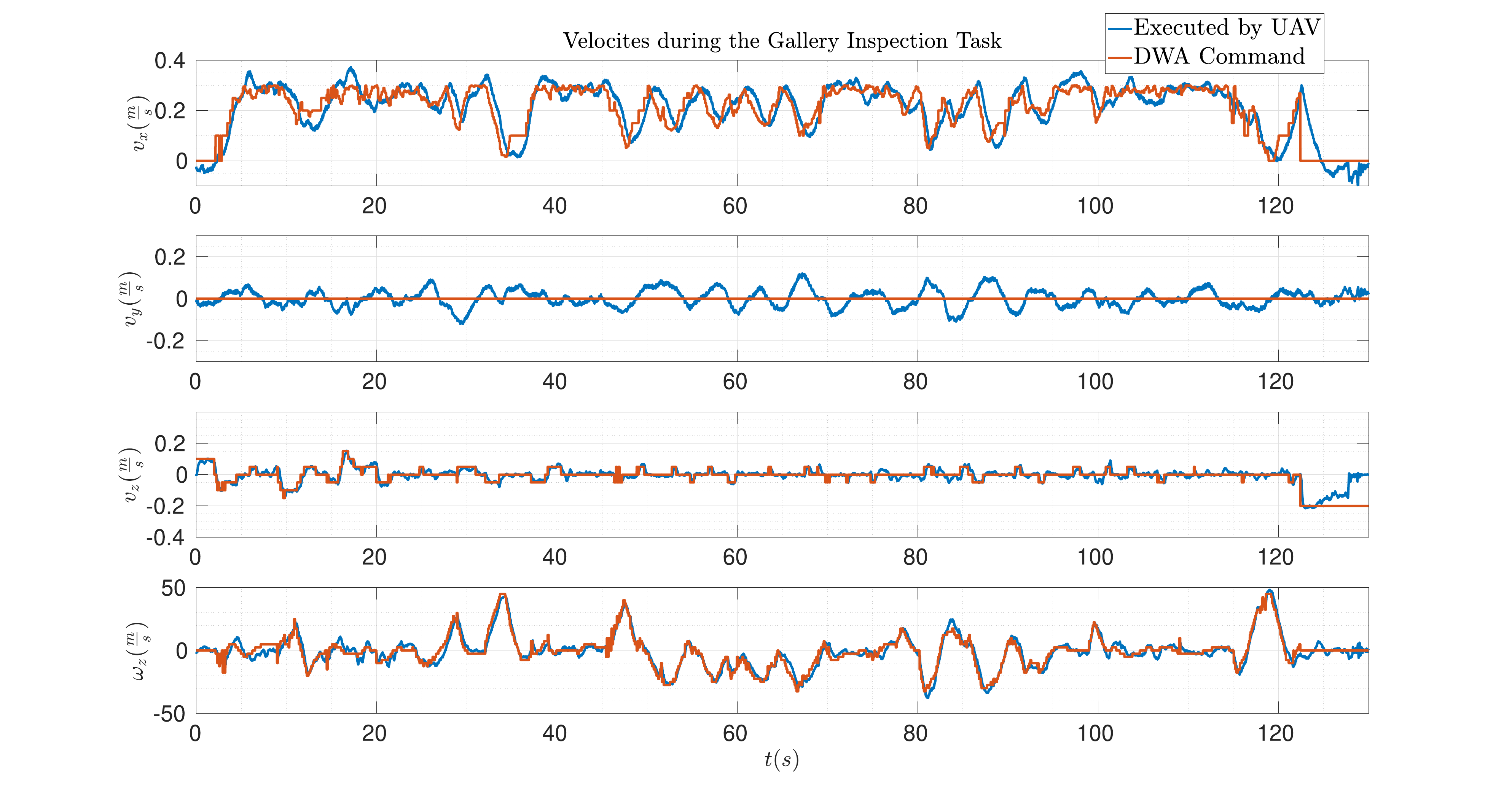}
        \caption{Velocities profiles during the gallery inspection task with $v_{x}^{max} = 0.3 \frac{m}{s}$. In red, the one commanded by DWA-3D; in blue the ones executed by the UAV.}
        \label{fig:BasementVelocities}
    \end{subfigure}
    \caption{Gallery inspection task trajectory (top and middle) and velocities (desired and executed) during it (bottom). {\color{blue}\href{https://www.youtube.com/watch?v=SmczpY4ZBwc}{VIDEO}}}
    \label{fig:Basement}
\end{figure}

\subsubsection{Facilities Gallery Scenario}
An inspection mission is emulated in a gallery, \autoref{fig:Basement_FPV}. The UAV must advance in a narrow corridor ($3m$ at its widest area) while dodging several dangers that appear along its way, turn left to enter a poorly illuminated room and reach its center while avoiding an unexpected obstacle. Later, it comes back to its takeoff point. This scenario involves added challenges as the entrance to the target room is initially occluded, \autoref{fig:BasementFPVInitial}, and its contents are unknown beforehand. Thus, it exhibits DWA-3D's navigation capabilities in a mostly hidden scenario with a constantly changing global plan. Additionally, the bad light conditions inside the room, /\autoref{fig:BasementFPVSala}, prohibit relying on a visual perception system, reinforcing the LiDAR-based selection. Key fragments of the trajectory are shown in \autoref{fig:BasementTrajectoryPart1} and \autoref{fig:BasementTrajectoryPart2}. The velocity profiles in \autoref{fig:BasementVelocities} reflect that the UAV executes a zig-zag maneuver to avoid the obstacles near the corridor walls while approaching the room. Around $t = 35s$, it orients towards the entrance performing a $90 ^{\circ}$ left rotation and accelerates to cross the door. Then, the room inspection, including navigation around the suddenly discovered obstacle, occurs between $t = 50s$ and $t = 80s$. Finally, the drone leaves the room and returns to the starting point, revisiting the corridor. Compared with previous experiments, $v_x$ and $\omega_z$ exhibit a more oscillatory behavior because the UAV must maneuver in a narrow space to avoid the obstacles along the corridor and within the room. Coupled with those maneuvers, slight $v_y$ drift appears.    


\subsubsection{Somport Tunnel Inspection}
\begin{figure}[h!]
    \centering
        \includegraphics[width = \linewidth, trim=0cm 0cm 0cm 0cm, clip,keepaspectratio]{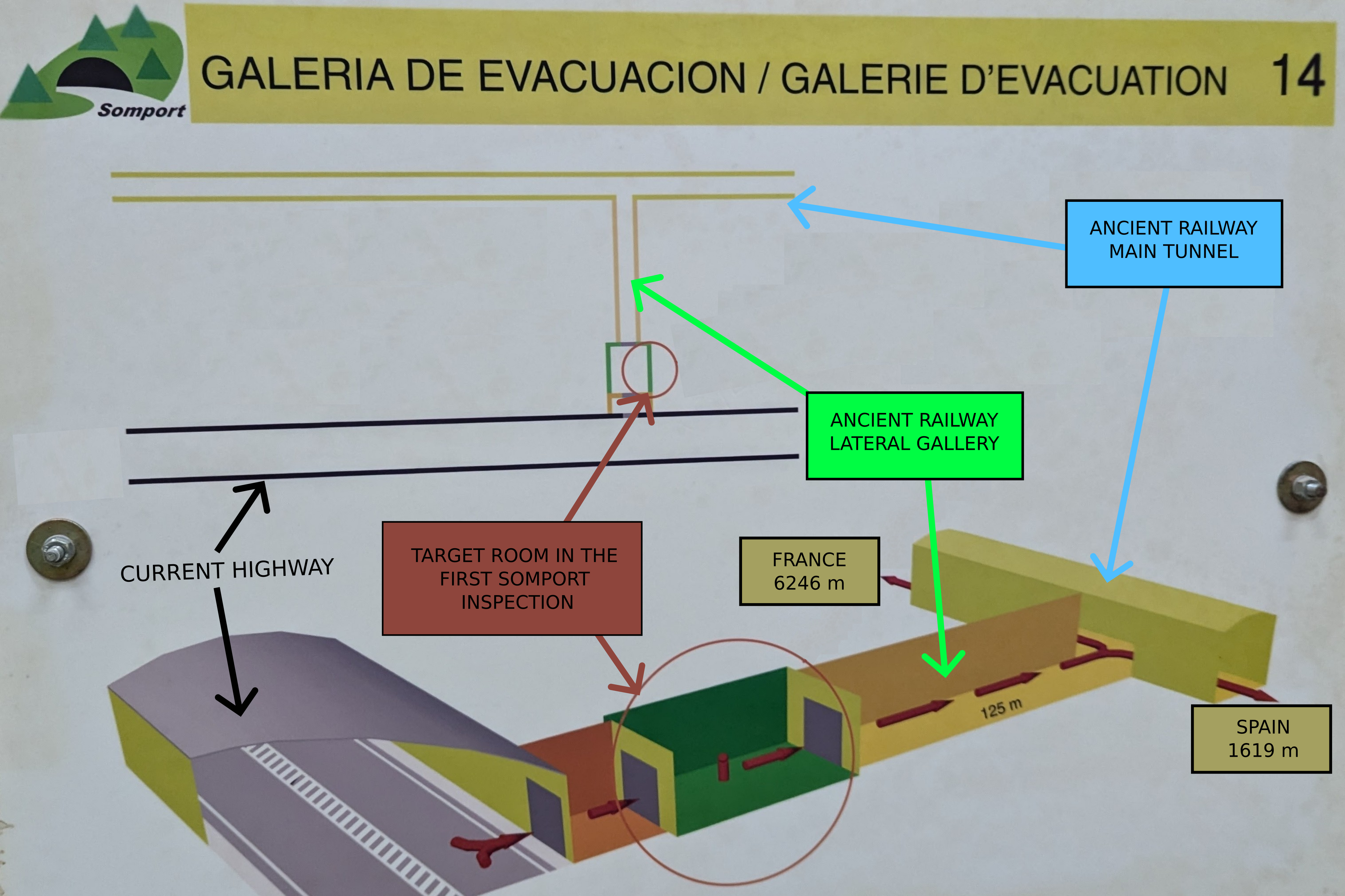}
        \caption{Highway evacuation plane located inside the inspected room at the end of one of the lateral galleries. The first inspection starts at the end of the lateral gallery, enters the target room and returns. The second inspection starts near the intersection between the auxiliary gallery and the main tunnel.}
        \label{fig:PlanoSomport}
\end{figure}

Finally, the system faced a complex environment, the Somport railway tunnel, an ancient railway tunnel that used to connect Spain and France through the Pyrenees, which is 7.7 km long. The main tunnel has 16 lateral auxiliary galleries distributed all along its way. Some of these galleries connect the ancient tunnel with the current highway tunnel between Spain and France, serving as evacuation ways, see \autoref{fig:PlanoSomport}. One of the tasks manually performed by the people in charge of the tunnel control and maintenance is the periodical inspection of the tunnel walls and facilities, to evaluate the state and safety of this infrastructure under a mountain, subjected to humidity, water, wind and dust. A clear improvement of this task in this long tunnel can be obtained by automating it by means of drones autonomously navigating, only supervised by the control center. The experiments were designed for an autonomous drone navigation with an onboard camera registering the scenario. The tunnel and galleries involve additional challenges to the ones that appear in more structured scenarios, such as dust and variable wind that disturb the perception and the flight. The experiments were considered as a concept proof for evaluating the robustness of the navigation and localization methods in this hostile scenario. 

\begin{figure}[h!]
    \centering
    \begin{subfigure}{\columnwidth}
    \centering
        \includegraphics[width=\linewidth, keepaspectratio, trim=0cm 0cm 0cm 3.4cm, clip]{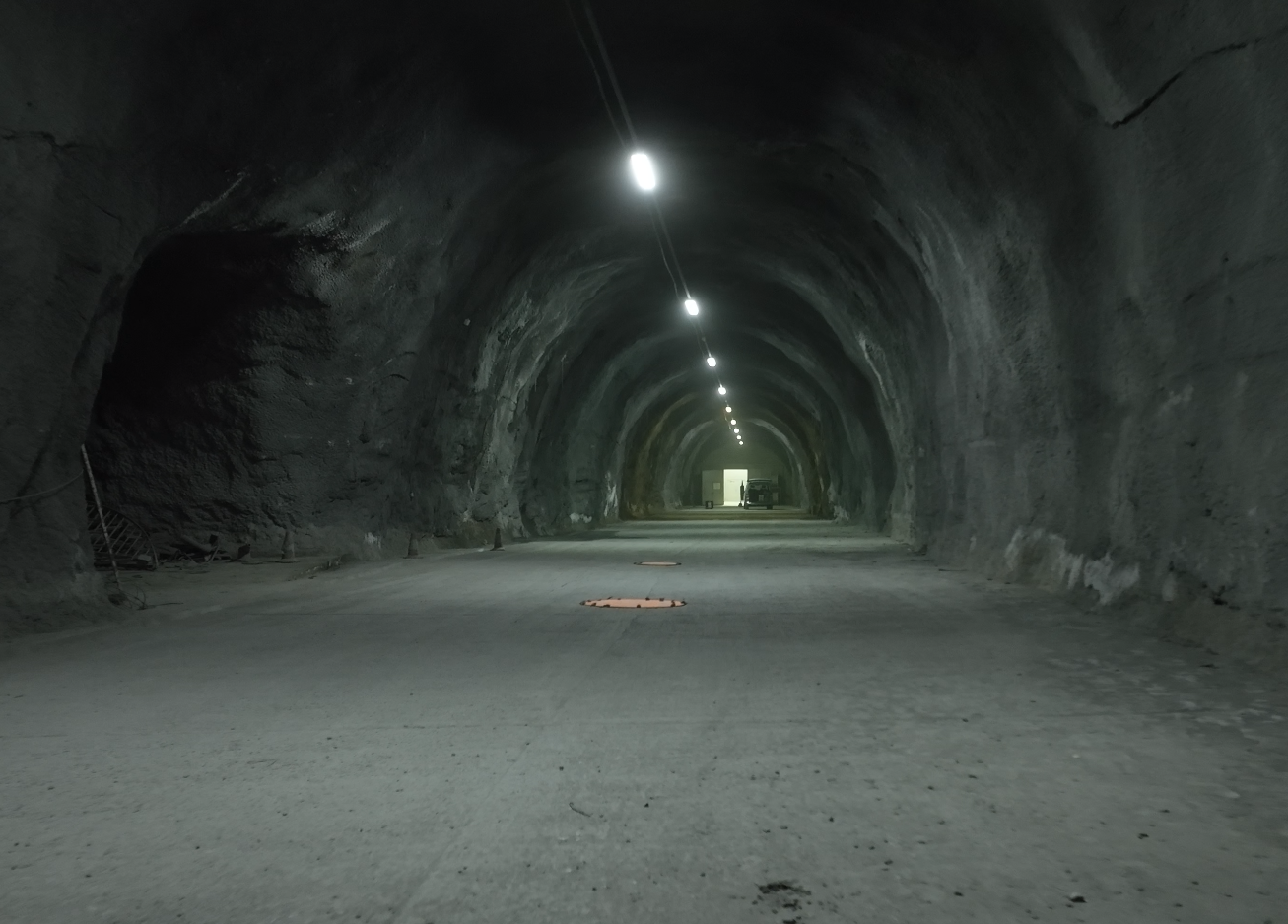}
        \caption{Somport lateral gallery 14. The target room of the first mission is at the background. The orange circle at the ground is the second inspection starting point.}
        \label{fig:SomportG14}
    \end{subfigure}
    \begin{subfigure}{\columnwidth}
    \centering
        \includegraphics[width=\linewidth,  keepaspectratio, trim=2.25cm 0cm 1cm 0cm, clip]{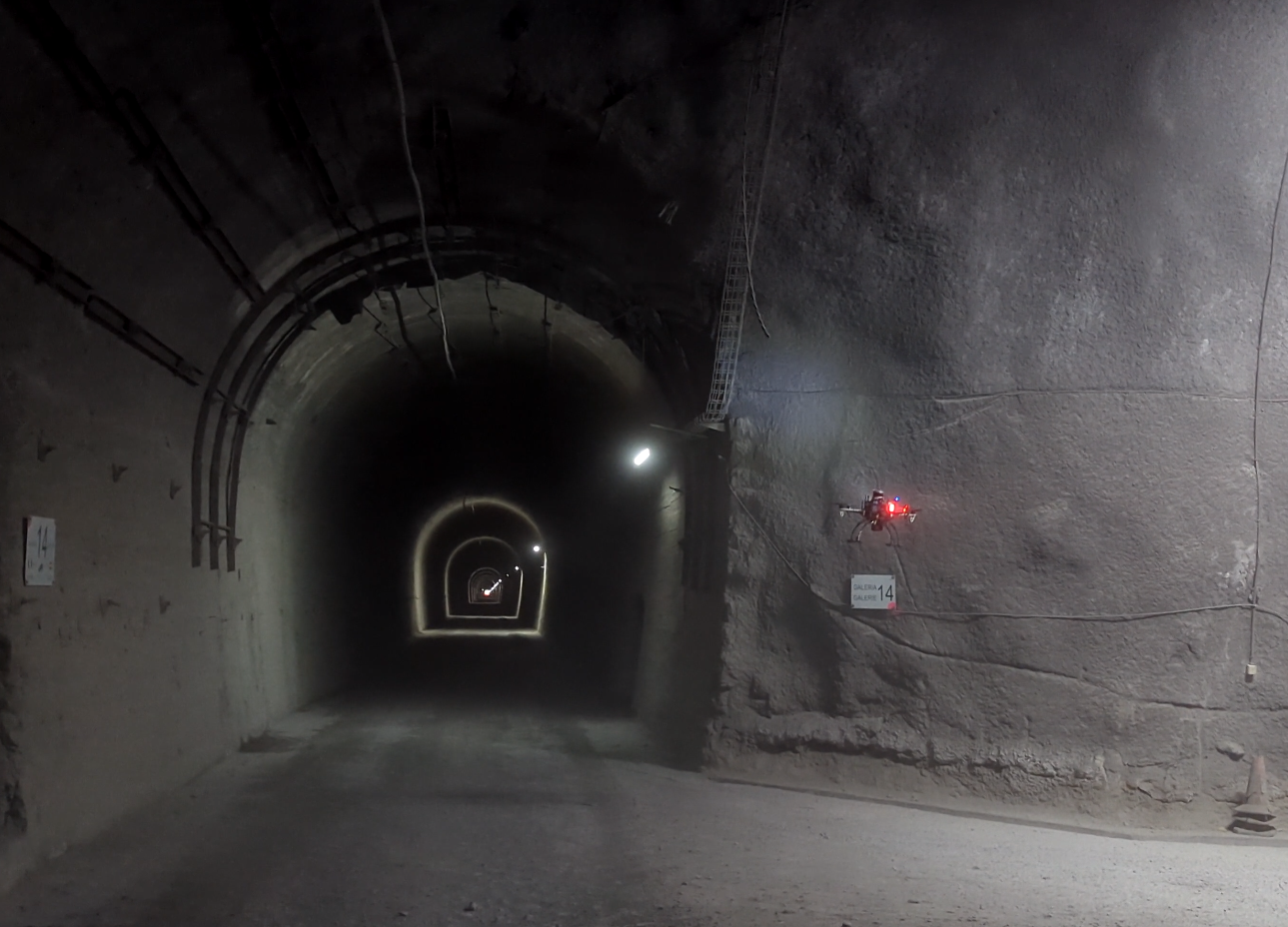}
        \caption{Intersection between the main tunnel and gallery 14. The main tunnel leads to Spain in the direction in which the image was taken. Strong variable winds blow coming from Spain to France. Both the wind and the UAV flight lift dust that would severely affect a visual-based perception system. Additionally, the main tunnel is poorly illuminated in several regions.}
        \label{fig:SomportInterseccion}
    \end{subfigure}
    \hfill
    \caption{Ancient Somport railway tunnel details. Both the downward slope of the gallery and the change of slope in the intersection involve an additional challenges to the localization system during the flights.}
    \label{fig:SomportDetalle}
\end{figure}

The drone performed two inspection missions, both starting at different points of gallery 14. Gallery 14 is shown in \autoref{fig:SomportG14} and the UAV entering its intersection with the main tunnel in \autoref{fig:SomportInterseccion}. Despite the additional challenges that this place involves, such as dust and variable winds that disturb the flight, the autonomous drone successfully navigated.

\begin{figure}[t]
    \centering
    \begin{subfigure}{\columnwidth}
    \centering
        \includegraphics[width=\linewidth, keepaspectratio, trim=0cm 0cm 0cm 3.4cm, clip]{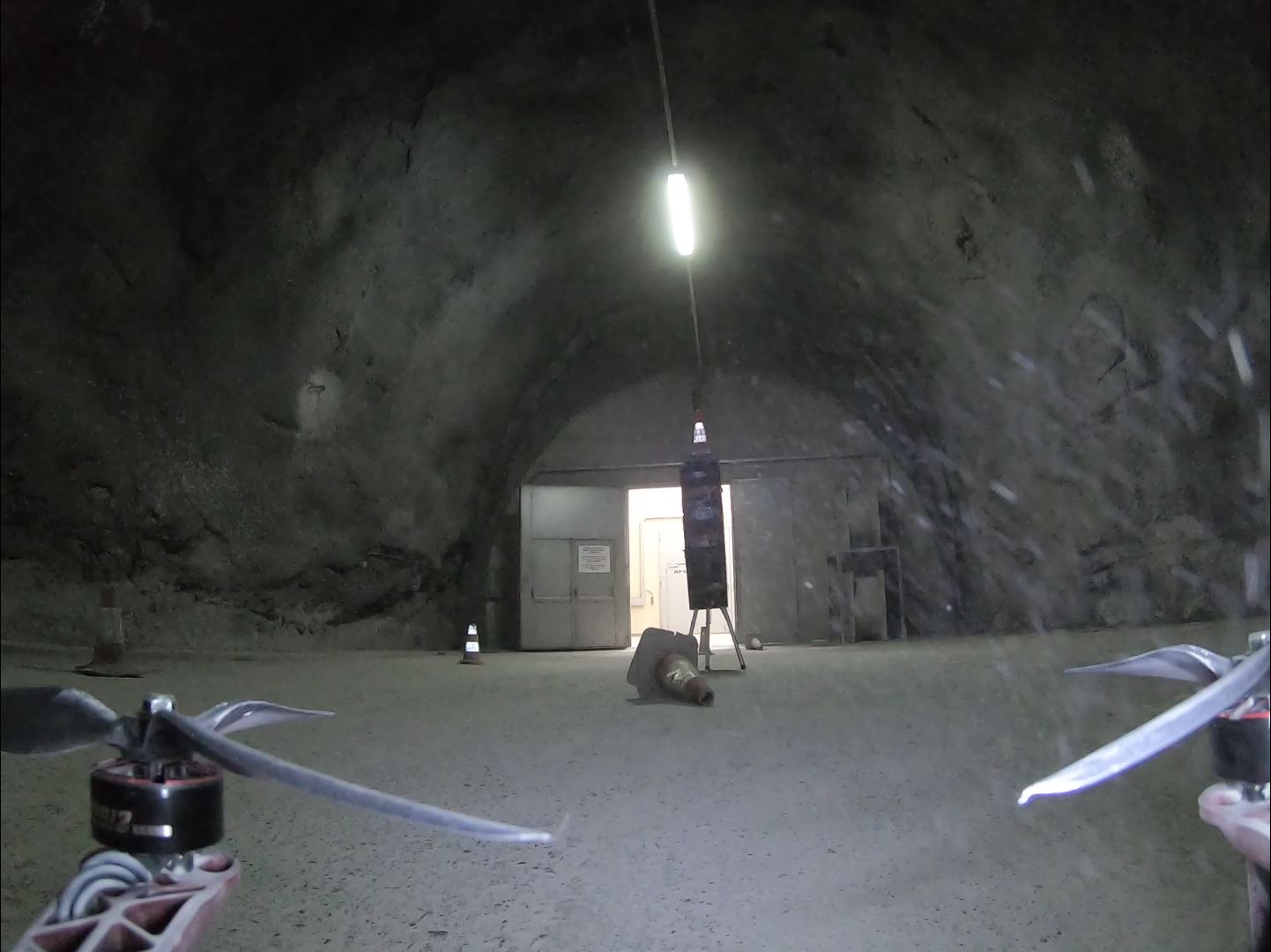}
        \caption{Drone's Point of View at takeoff point. The UAV must traverse a gallery with downward slope, inspect the room located at the background and return to the starting point. Diverse obstacles were intentionally placed to occlude the entrance.}
        \label{fig:Somport11Initial}
    \end{subfigure}
    \begin{subfigure}{\columnwidth}
    \centering
        \includegraphics[width=\linewidth,  keepaspectratio, trim=2.25cm 0cm 1cm 0cm, clip]{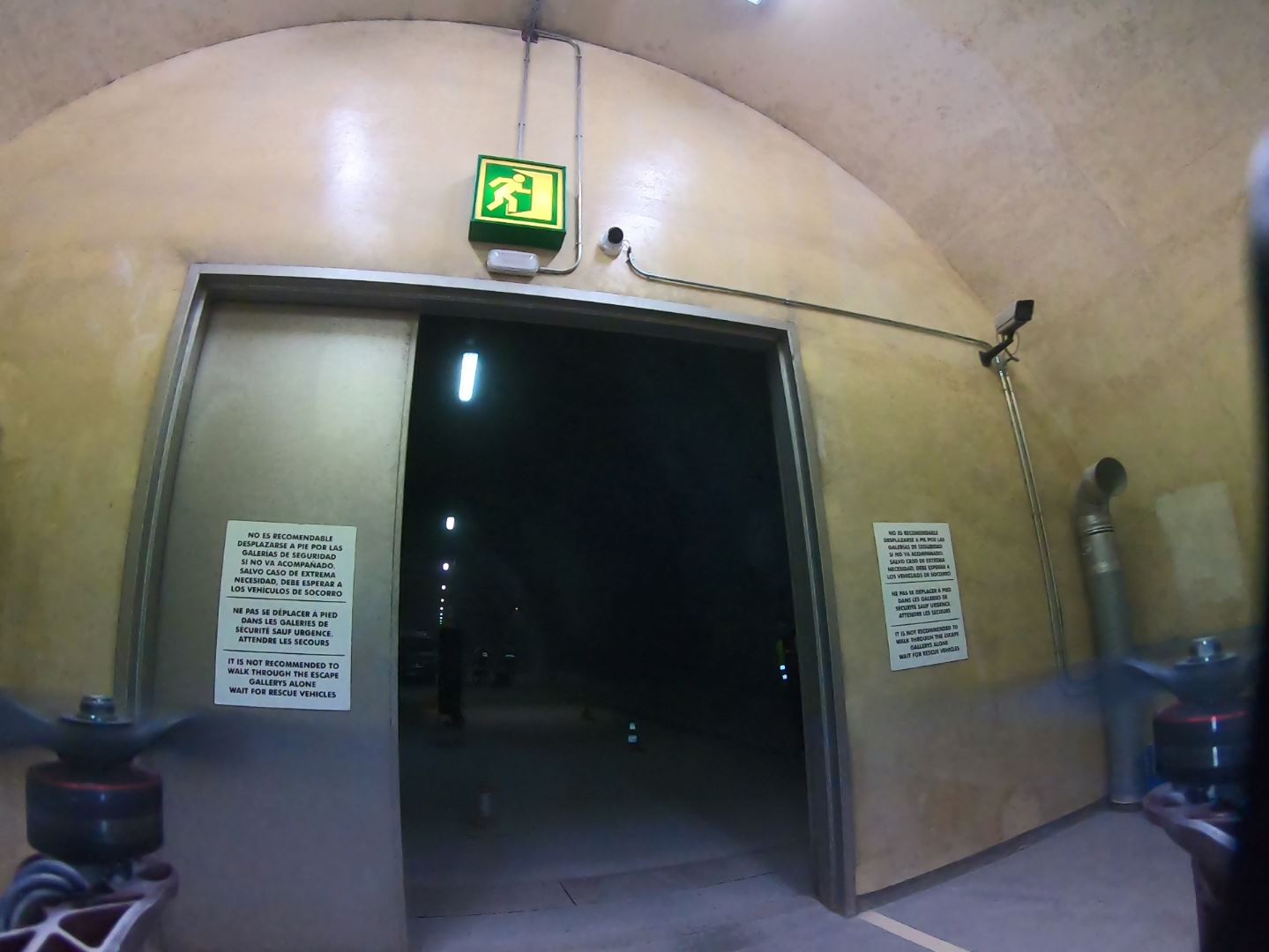}
        \caption{Room inspection has been performed and return is started. Note the significant lighting differences inside and outside the room.}
        \label{fig:Somport11Return}
    \end{subfigure}
    \hfill
    \caption{Different points during the first inspection in the Somport tunnel.}
    \label{fig:Vuelo11Somport}
\end{figure}

\begin{figure}[h!]
    \centering
    \begin{subfigure}{\columnwidth}
        \centering
        \includegraphics[width=\columnwidth,keepaspectratio, trim=2cm 0cm 0cm 0cm, clip]{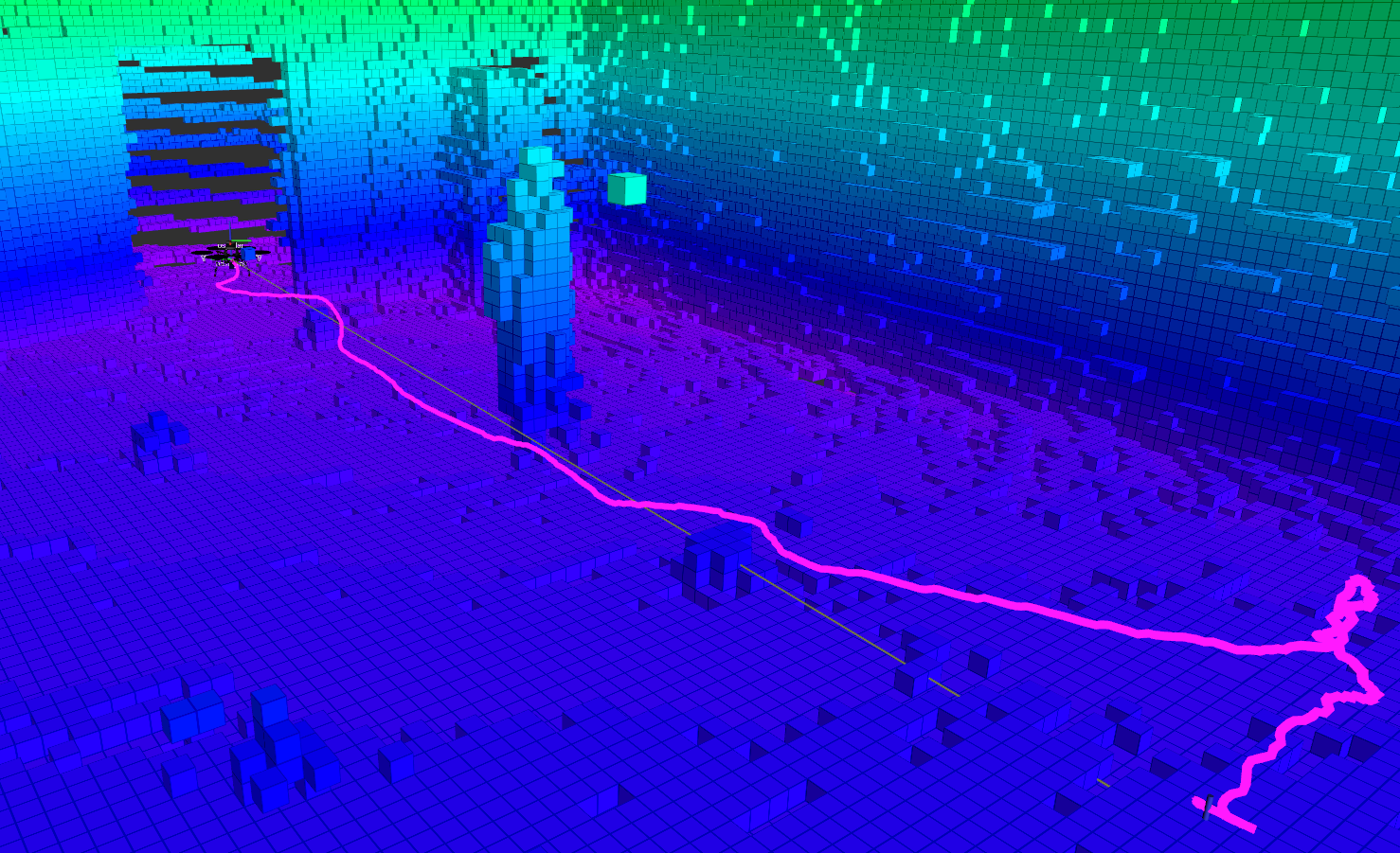}
        \caption{Performed trajectory (purple) to arrive the objective room during the first inspection task in the Somport Tunnel. Several obstacles were between the UAV initial location and the room entrance (see \autoref{fig:Somport11Initial}). The UAV executes a set of curve maneuvers to avoid them while approaching the door.}
        \label{fig:Somport11TrajectoryPart1}
    \end{subfigure}
    \begin{subfigure}{\columnwidth}
        \centering
        \includegraphics[width=\columnwidth,keepaspectratio, trim=0cm 1cm 0cm 0cm, clip]{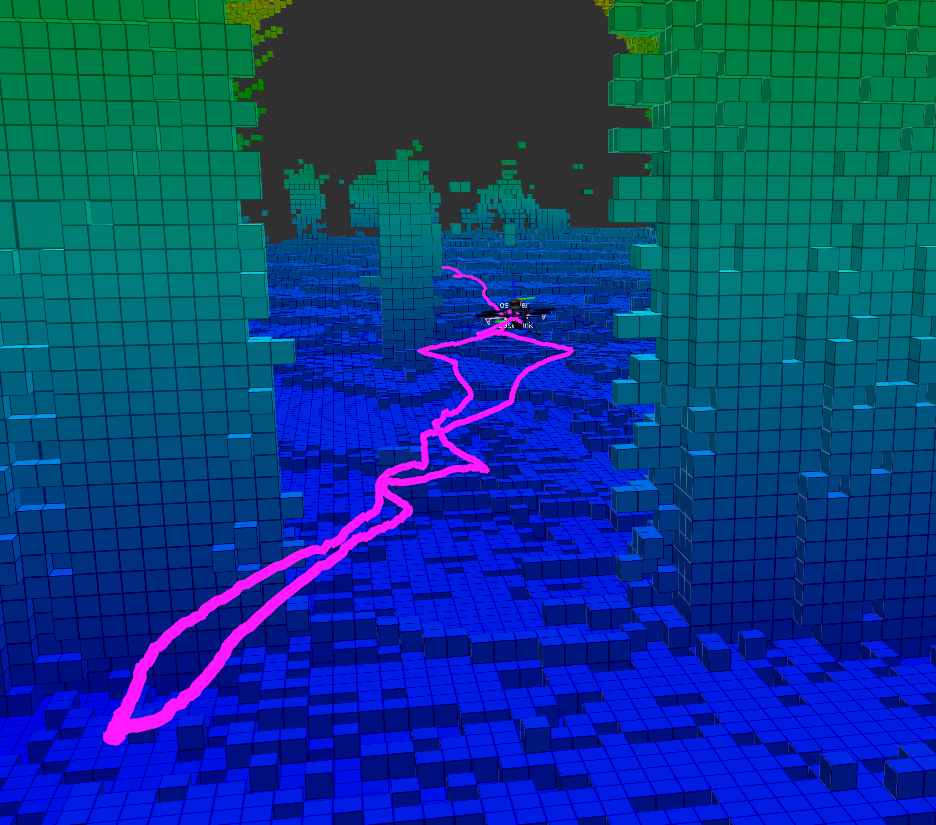}
        \caption{Performed trajectory (purple) to inspect and leave the room in the first inspection task in the Somport Tunnel. The maneuvers to properly orientate with the door frame before entering the room can be observed.}
        \label{fig:Somport11TrajectoryPart2}
    \end{subfigure}
    \begin{subfigure}{\columnwidth}
        \centering
        \includegraphics[width=\columnwidth,keepaspectratio, trim = 5cm 0cm 5cm 0cm, clip]{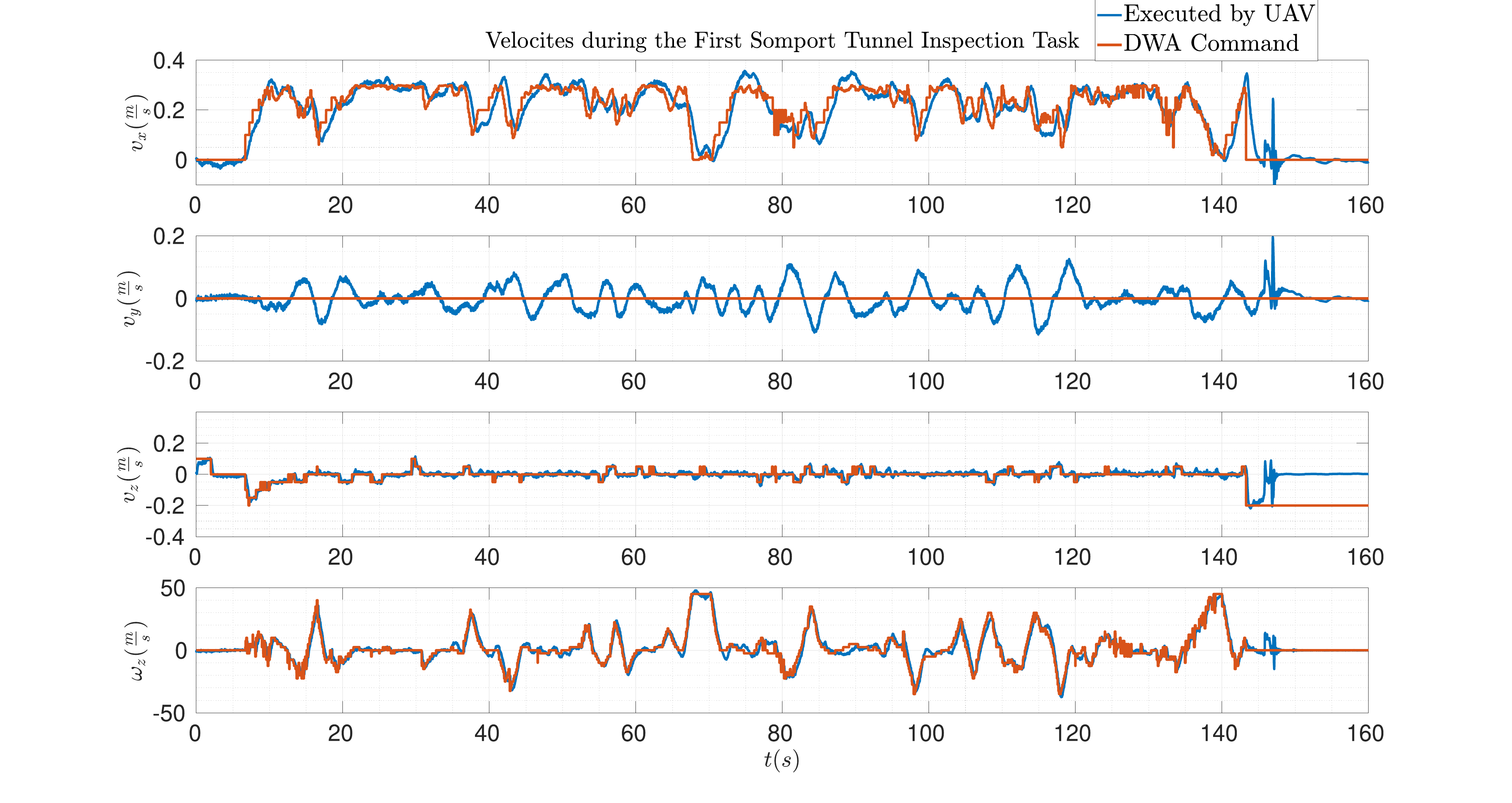}
        \caption{Velocities profiles during the first inspection task in the Somport Tunnel with $v_{x}^{max} = 0.3 \frac{m}{s}$. In red, the one commanded by DWA-3D; in blue the ones executed by the UAV.}
        \label{fig:Somport11Velocities}
    \end{subfigure}
    \caption{First inspection task in the Somport Tunnel (top and middle) and velocities (desired and executed) during it (bottom). {\color{blue}\href{https://www.youtube.com/watch?v=o5JKXsG04Z4}{VIDEO}}}
    \label{fig:Somport11}
\end{figure}

The initial inspection task was executed near the ending of the lateral gallery. The UAV was queried to advance through the downward-sloping gallery, enter a narrow room located at the end, with obstacles in the direct motion direction. It then returned to its initial location, ascending the slope without colliding. Obstacles were placed between the takeoff point and the room's entrance, intentionally out of the drone's initial field of view. Drone's point of view before takeoff and while exiting the room are shown in \autoref{fig:Somport11Initial} and \autoref{fig:Somport11Return}. \autoref{fig:Somport11TrajectoryPart1} and \autoref{fig:Somport11TrajectoryPart2} show the mission trajectory, both outbound and return. Note that the UAV maintained its altitude relative to the takeoff point despite the sloping terrain. The return maneuver starts around $t = 70s$, as it can be observed in the velocity profiles, \autoref{fig:Somport11Velocities}. DWA-3D commands braking plus a strong left rotation of $50 \frac{deg}{s}$ during $3s \sim 4s$ followed by a forward acceleration to recover the maximum velocity towards a new path that guides to the initial location.

\begin{figure}[t]
    \centering
    \begin{subfigure}{\columnwidth}
        \centering
        \includegraphics[width=\linewidth, keepaspectratio, trim=2cm 0cm 2cm 0cm, clip]{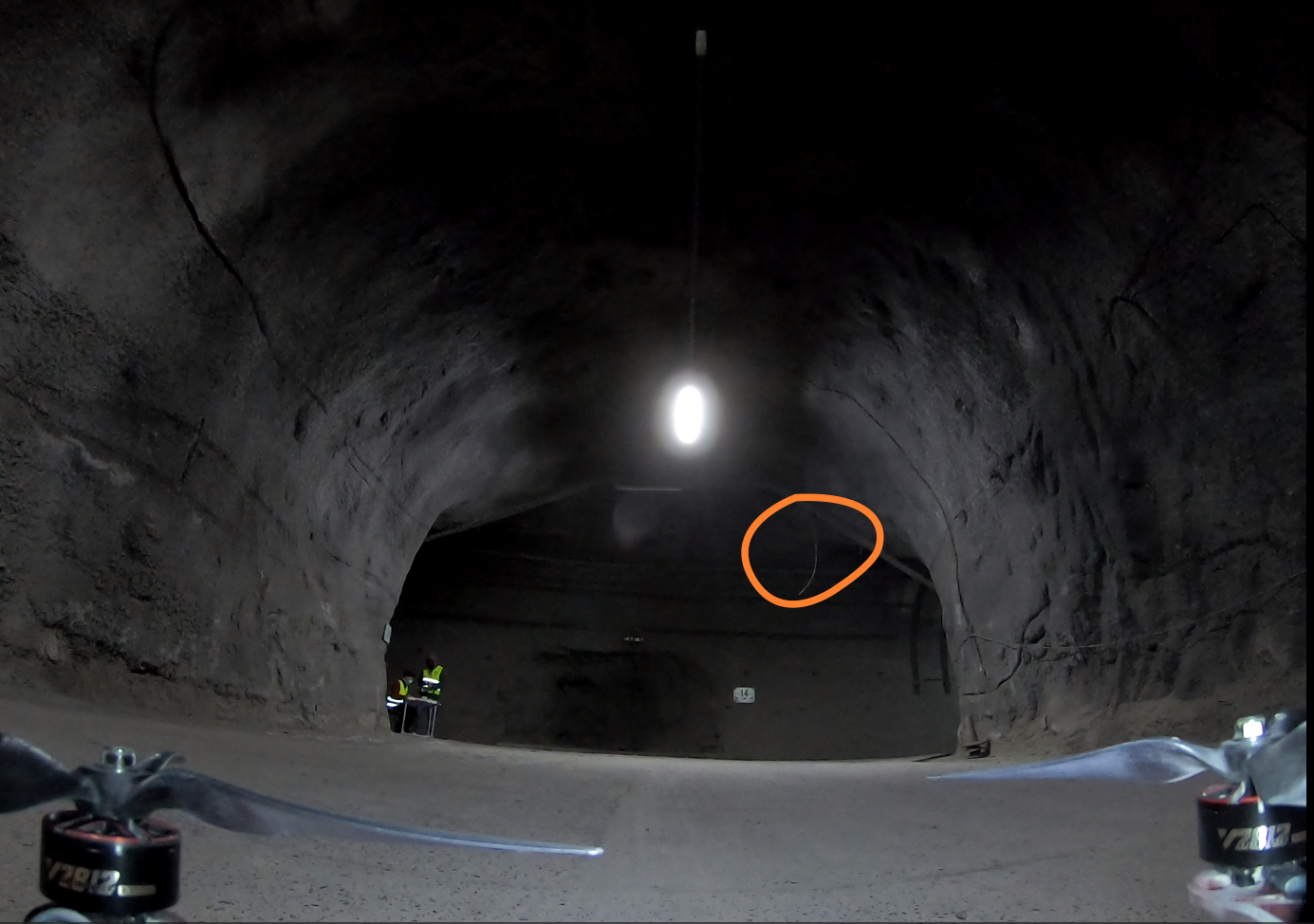}
        \caption{View at takeoff point. Notice the  thin wire at right hanging from the ceiling (circled in orange). The slope of the gallery until the intersection with the main tunnel can also be observed.}
        \label{fig:SomportVuelo13Ini}
    \end{subfigure}
    \begin{subfigure}{\columnwidth}
        \centering
        \includegraphics[width=\linewidth, keepaspectratio, trim=0cm 0cm 0cm 8.25cm, clip]{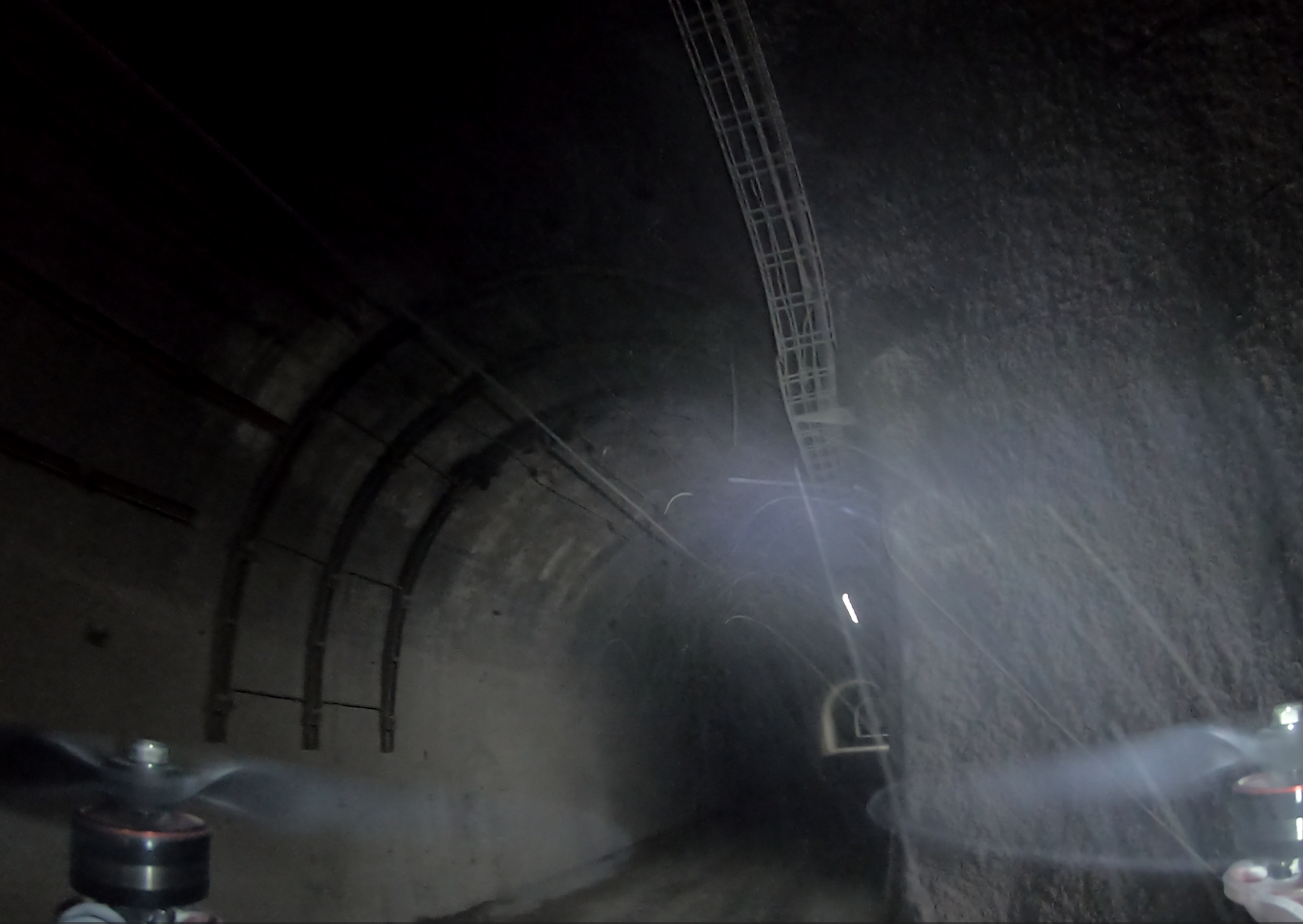}
        \caption{Frame captured while entering the main tunnel. The sudden wind blows dust towards the camera. Dark regions of the tunnel can be observed after turning around the corner.}
        \label{fig:Somport13Wind}
    \end{subfigure}
    \hfill
    \caption{FPV images taken from different points during the second inspection task in the Somport tunnel.}
    \label{fig:Vuelo13Somport}
\end{figure}

\begin{figure}[h!]
    \centering
    \begin{subfigure}{\columnwidth}
        \centering
        \includegraphics[width=\columnwidth,keepaspectratio, trim=0cm 2.25cm 0cm 2cm, clip]{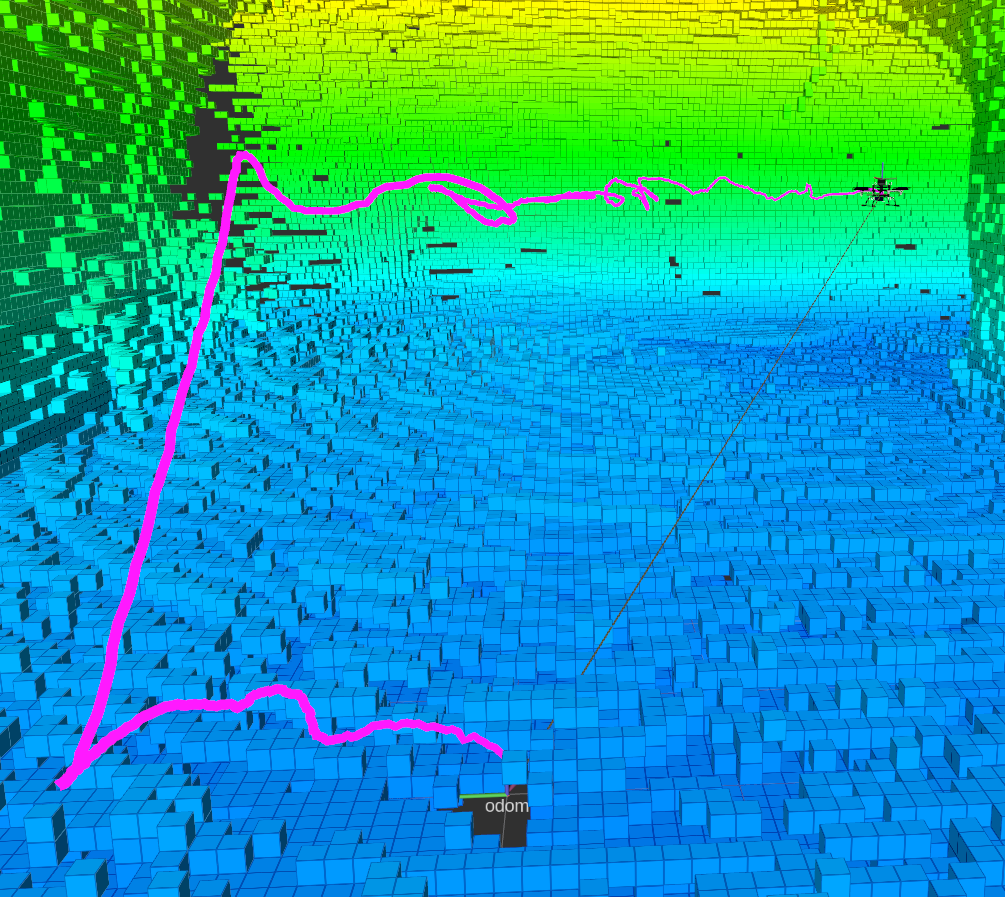}
        \caption{Performed trajectory (purple) to enter the main tunnel during the second inspection task in the Somport Tunnel. Initially, changing winds dragged the UAV to the left while it was idle waiting for the mission to be sent.}
        \label{fig:Somport13TrajectoryPart1}
    \end{subfigure}
    \begin{subfigure}{\columnwidth}
        \centering
        \includegraphics[width=\columnwidth,keepaspectratio, trim=0cm 0cm 0cm 0cm, clip]{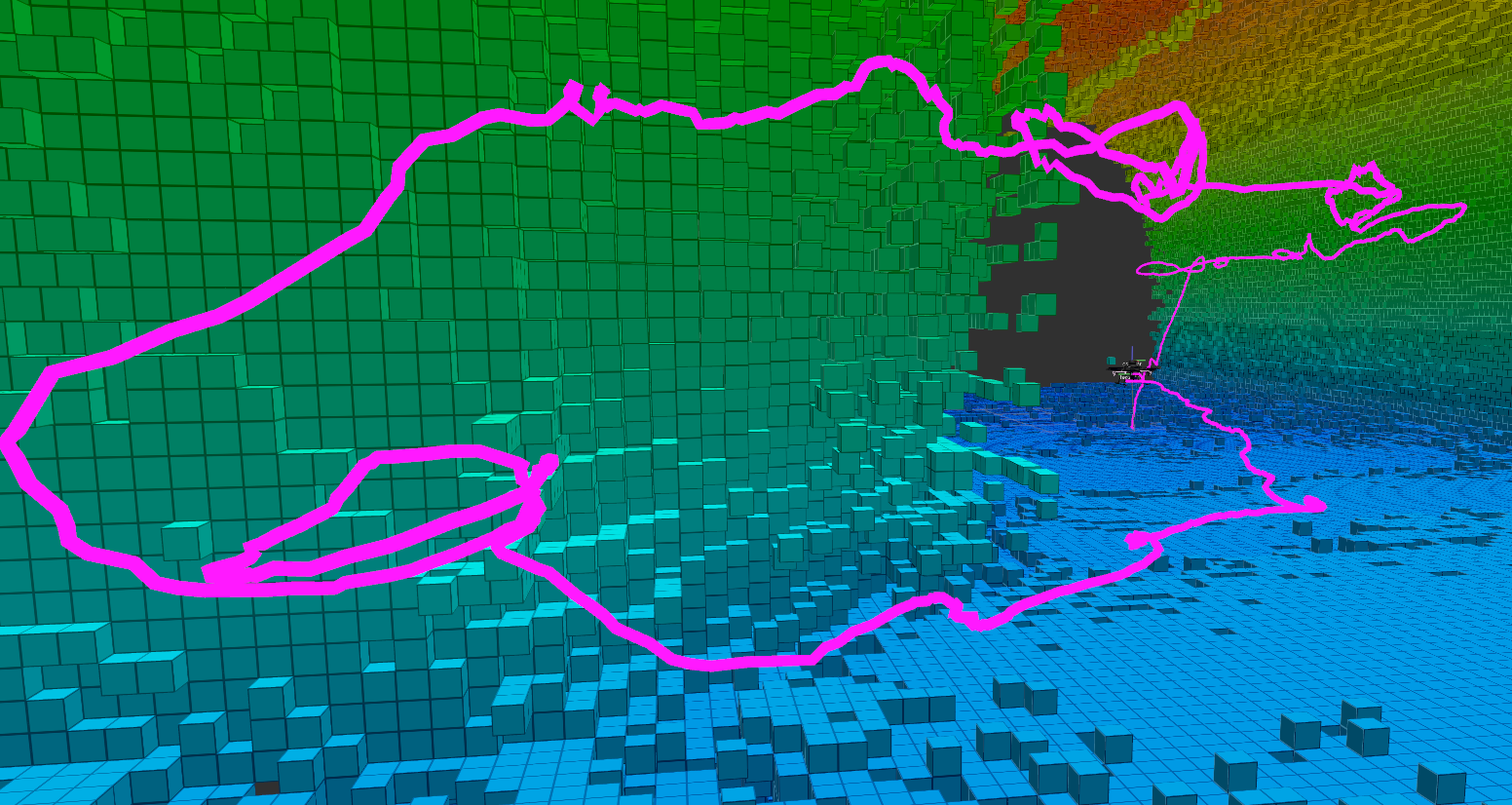}
        \caption{Performed trajectory (purple) to return the takeoff point during the second inspection task in the Somport Tunnel. Notice the different attitude in the outbound and return paths. The changing wind perturbations provoked oscillations during the flight, specially once the main tunnel is reached. Despite the disturbances, DWA-3D performs the maneuvers needed to keep the UAV away from the walls.}
        \label{fig:Somport13TrajectoryPart2}
    \end{subfigure}
    \begin{subfigure}{\columnwidth}
        \centering
        \includegraphics[width=\columnwidth,keepaspectratio, trim = 5cm 0cm 5cm 0cm, clip]{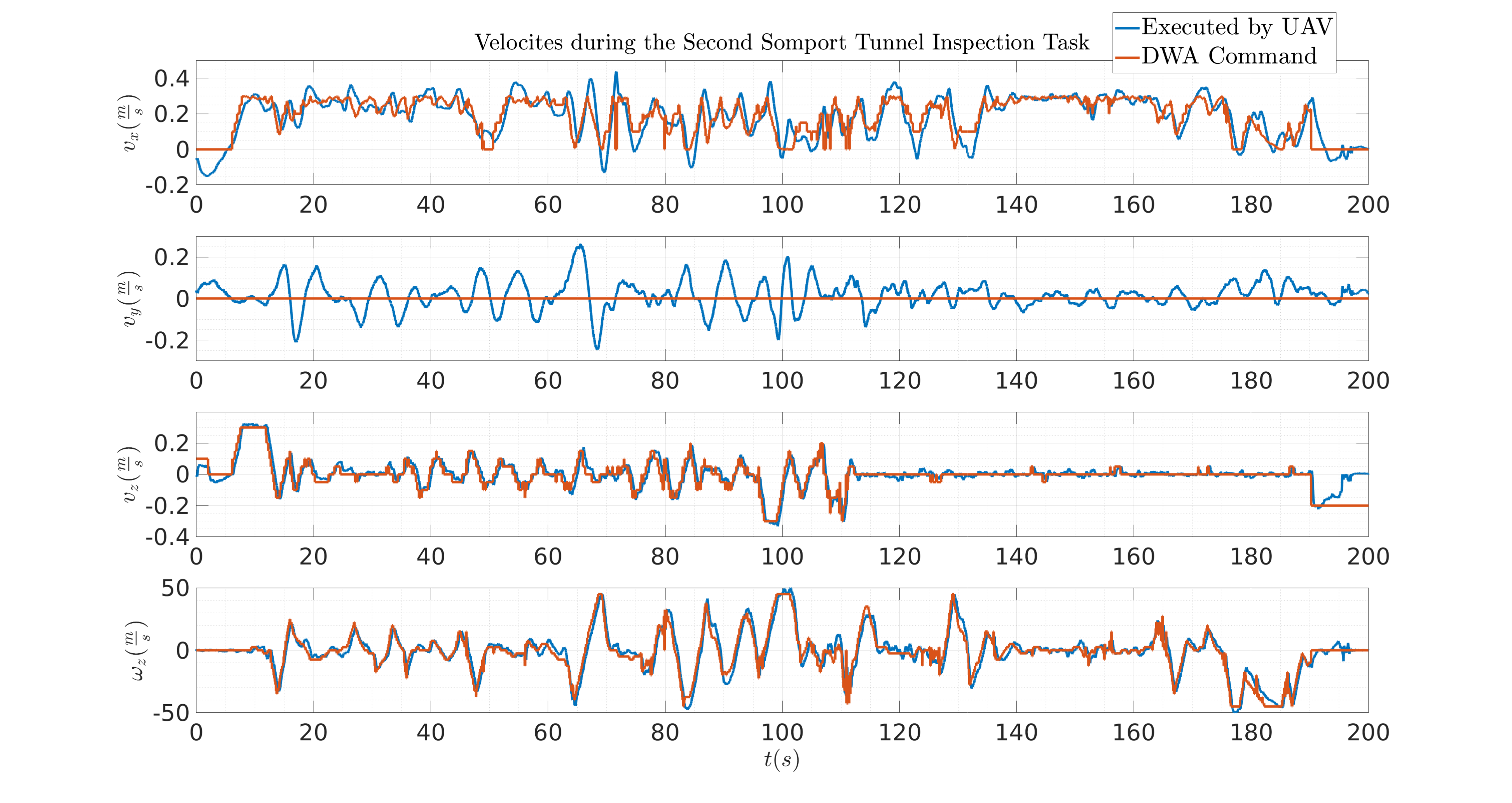}
        \caption{Velocities profiles during the second inspection task in the Somport Tunnel with $v_{x}^{max} = 0.3 \frac{m}{s}$. In red, the one commanded by the local planner (DWA-3D); in blue the ones executed by the UAV.}
        \label{fig:Somport13Velocities}
    \end{subfigure}
    \caption{Second inspection task in the Somport Tunnel trajectory (top and middle) and velocities (desired and executed) during it (bottom). {\color{blue}\href{https://www.youtube.com/watch?v=zmzGoHzXYsc}{VIDEO}}}
    \label{fig:Somport13}
\end{figure}

In the later inspection mission, the UAV was tasked with navigating the upward-sloping gallery, accessing the main tunnel, and returning to the takeoff location, but at a different altitude ($1m$) than its outbound path ($2.5$m). Upon entering the main tunnel, a thin wire hanging from the ceiling must be avoided, see \autoref{fig:SomportVuelo13Ini}. Additionally, strong winds are encountered when entering the main tunnel, so DWA-3D must recover from the unexpected perturbation and keep following the plan; notice the wind direction because of the dust blowing towards the camera in \autoref{fig:Somport13Wind}. DWA-3D control reacts to the disturbances and guides the UAV away from the nearby walls, enforcing a forward motion towards the goal. The resulting trajectory shown in \autoref{fig:Somport13TrajectoryPart1} and \autoref{fig:Somport13TrajectoryPart2} clearly display the wind perturbations along the flight, specially during main tunnel entry. This effect can also be observed in the velocity profiles shown in\autoref{fig:Somport13Velocities}; specifically at $t = 70s$ and $t = 85$ the UAV's $v_x$ exhibits negative values (despite DWA-3D never commanding $v_x < 0 $), while $v_y$ reaches peaks around $0.2 \sim 0.25 \frac{m}{s})$ due to the disturbances. Finally, $v_x$ suffers a slightly oscillatory behavior as the UAV approaches the ground for its return maneuver. This is caused by a sudden dust cloud, generated because of the proximity to the floor. DWA-3D perceives it as a new danger and decides to brake and avoid it.

\subsection{Comparison with FASTER}
\label{sec:ComparacionFASTER}
\begin{figure}[h]
    \centering
        \includegraphics[width = \linewidth, trim=5cm 0cm 5cm 0cm, clip,keepaspectratio]{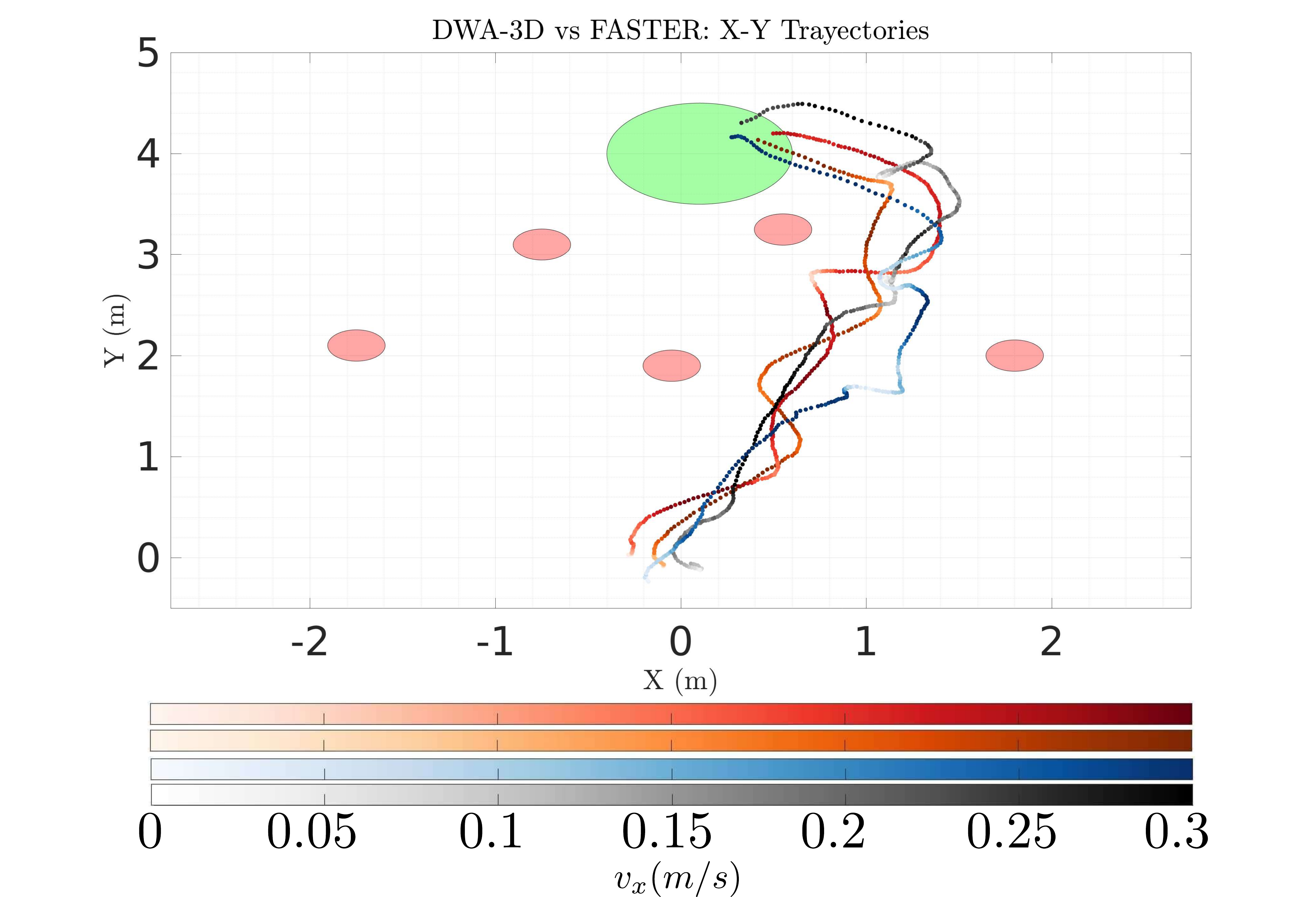}
        \caption{DWA-3D (red and orange) and FASTER (blue and black) trajectories performed sorting a set of six obstacles in our "Drone Arena". The darkness of each trace is related with the UAV velocity at each instant. Red circles represent the obstacles, while the green one is the goal.}
        \label{fig:DWA_vs_FASTER_sparse_traj}
\end{figure}

We compared our proposal with FASTER \cite{tordesillas2021faster}, the referenced work in \autoref{tab:SOTA_Comp} with downloadable code and  high maneuverability capabilities. They are evaluated in two new scenarios. In order to equalize the experimental conditions, both planners are integrated within the same architecture, using the onboard 3D-LiDAR for perception and F-LOAM for localization. In the first scenario they must traverse through the narrow gaps left between a set of six cylinders distributed all around our "Drone Arena". \autoref{fig:DWA_vs_FASTER_sparse_traj} shows the trajectories performed by DWA-3D (orange and red) and FASTER (black and blue) in two consecutive flights each. The darker the trace, the higher is the velocity in the XY plane. DWA-3D is able to maintain a higher velocity while maneuvering to avoid the obstacles. The results presented in \autoref{tab:dwa_vs_faster_sparse_tab} reinforce that observation, it contains the time needed for each flight, as well as the traversed distance and the medium velocity during them. Our approach is able to complete the mission in less time, travel less distance and maintain a higher mean velocity. 

\begin{table}[h!]
    \centering
    \begin{tabular}{|*{4}{c|}}
        \hline
        \textbf{Flight} & \textbf{T(s)} & \textbf{Dist.(m)} & \textbf{Mean Vel.(m/s)} \\
        \hline
        {\color{gray}FASTER 1} & 35.6s & 7.1m & 0.20 m/s \\
        \hline
        {\color{blue}FASTER 2} & 27.3s & 6.2m & 0.23 m/s \\
        \hline
        {\color{orange}DWA-3D 1} & 21.6s & 5.4m & 0.25 m/s \\
        \hline
        {\color{red}DWA-3D 2} & 27.2s & 5.8m & 0.21 m/s \\
        \hline
    \end{tabular}
    \caption{DWA-3D and FASTER results from sorting the set of six obstacles in our "Drone Arena".}
    \label{tab:dwa_vs_faster_sparse_tab}
\end{table}

\begin{figure}[h!]
    \centering
        \includegraphics[width = \linewidth, trim=5cm 0cm 5cm 0cm, clip,keepaspectratio]{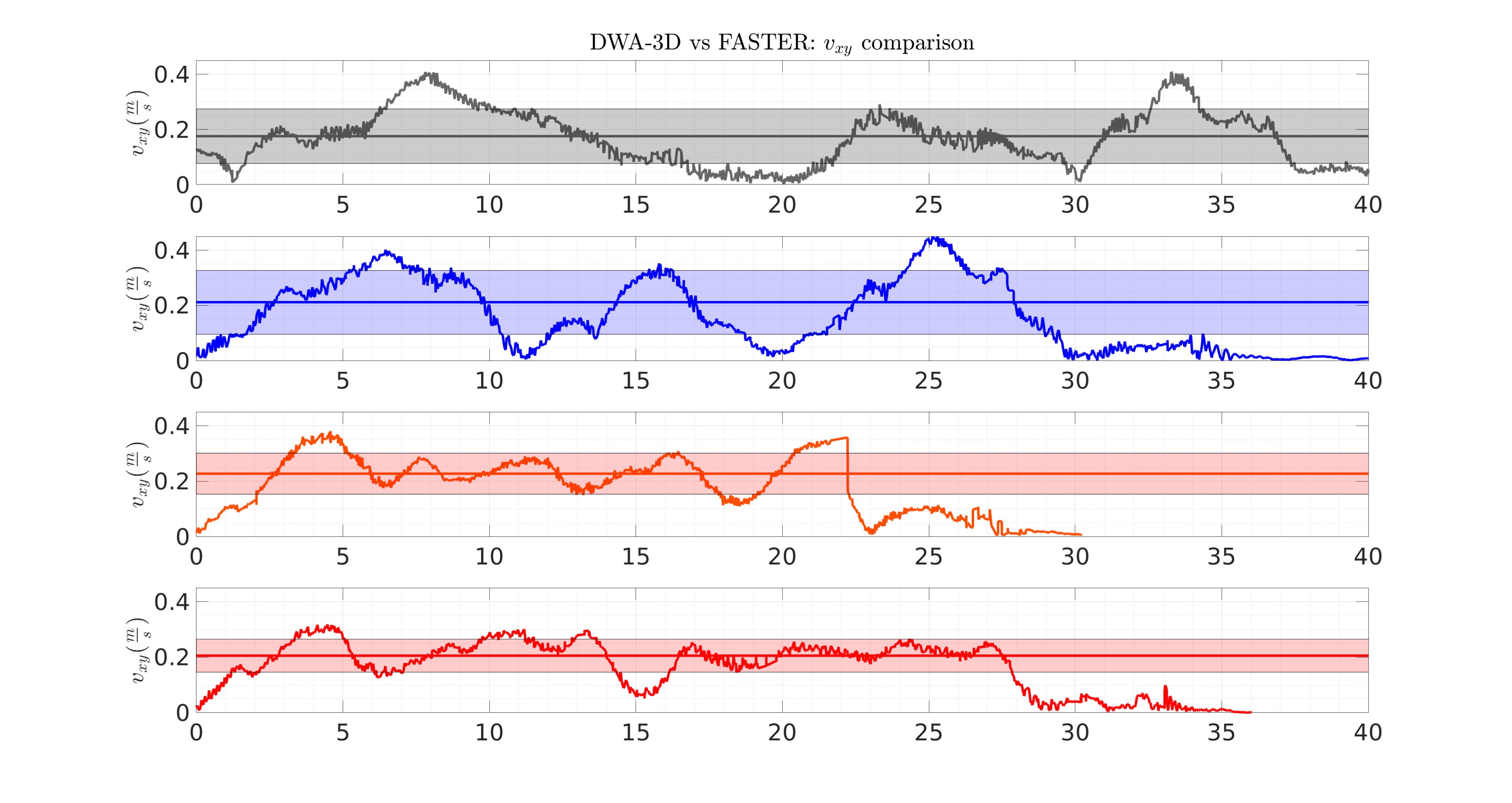}
        \caption{DWA-3D (red and orange) and FASTER (blue and black) $v_{xy}$ temporal evolution while sorting a set of six obstacles in our "Drone Arena". Mean value for each experiment is also represented. The shadowed areas represent mean $\pm$ standard deviation. Note that DWA needs less time, has a greater mean velocity and experiments lower standard deviation.}
        \label{fig:Dwa_vs_FASTER_vel_sparse}
\end{figure}

Additionally, by analyzing the temporal evolution of $v_{xy}$ for each experiment, \autoref{fig:Dwa_vs_FASTER_vel_sparse}, we can conclude that DWA-3D performs smoother flights and maintains a more continuous and stable velocity, needing to decelerate and accelerate less than FASTER. This softer behavior is translated into a lower standard deviation for DWA-3D. FASTER had to stop while replanning in several occasions because it does not compute a plan until the final location, it divides the whole trajectory into segments.

\begin{figure}[h!]
    \centering
        \includegraphics[width = \linewidth, trim=0cm 0cm 1cm 0cm, clip,keepaspectratio]{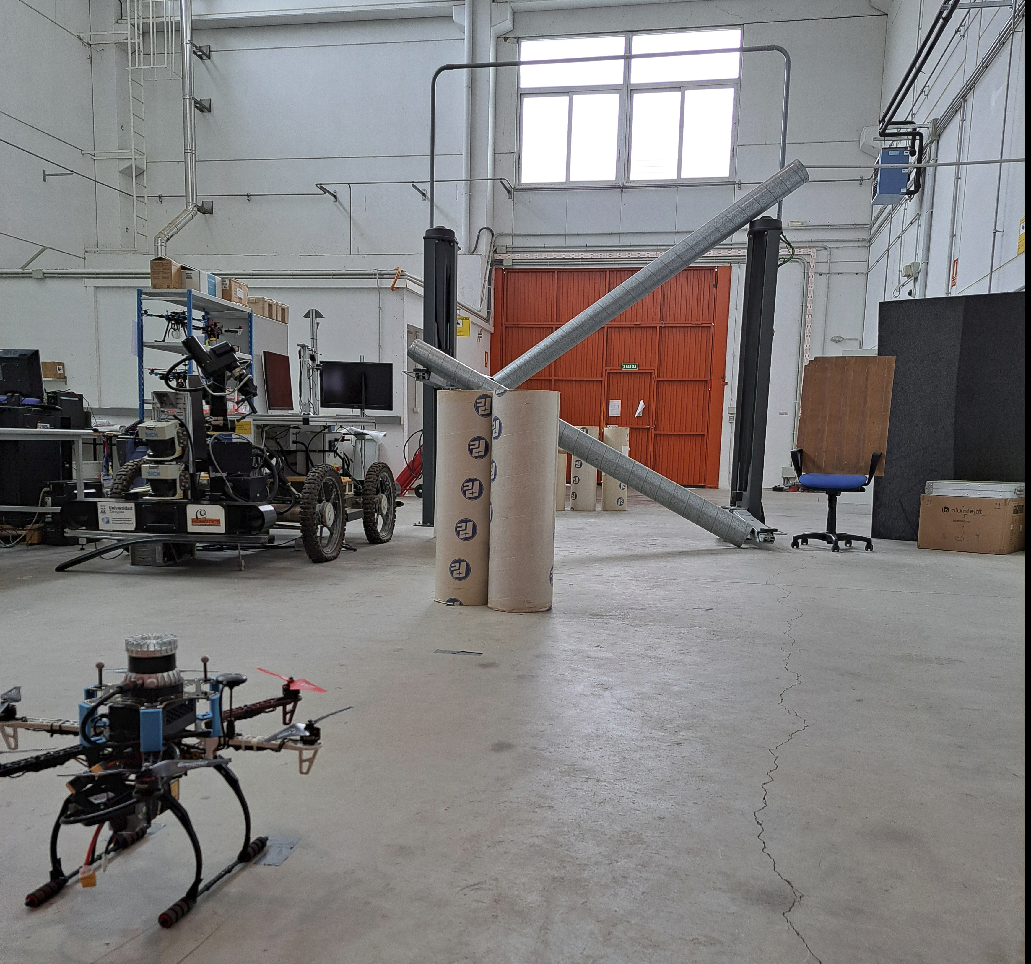}
        \caption{Scenario used for the second comparison between DWA-3D and FASTER. The first pair of cylinders must be avoided laterally, then the UAV has to fly over the first fallen pipe and under the second one; finally a set of three cylinders requires another lateral maneuver.}
        \label{fig:ComparacionNave}
\end{figure}

\begin{figure}[h!]
    \centering
        \includegraphics[width = \linewidth, trim=5cm 0cm 5cm 0cm, clip,keepaspectratio]{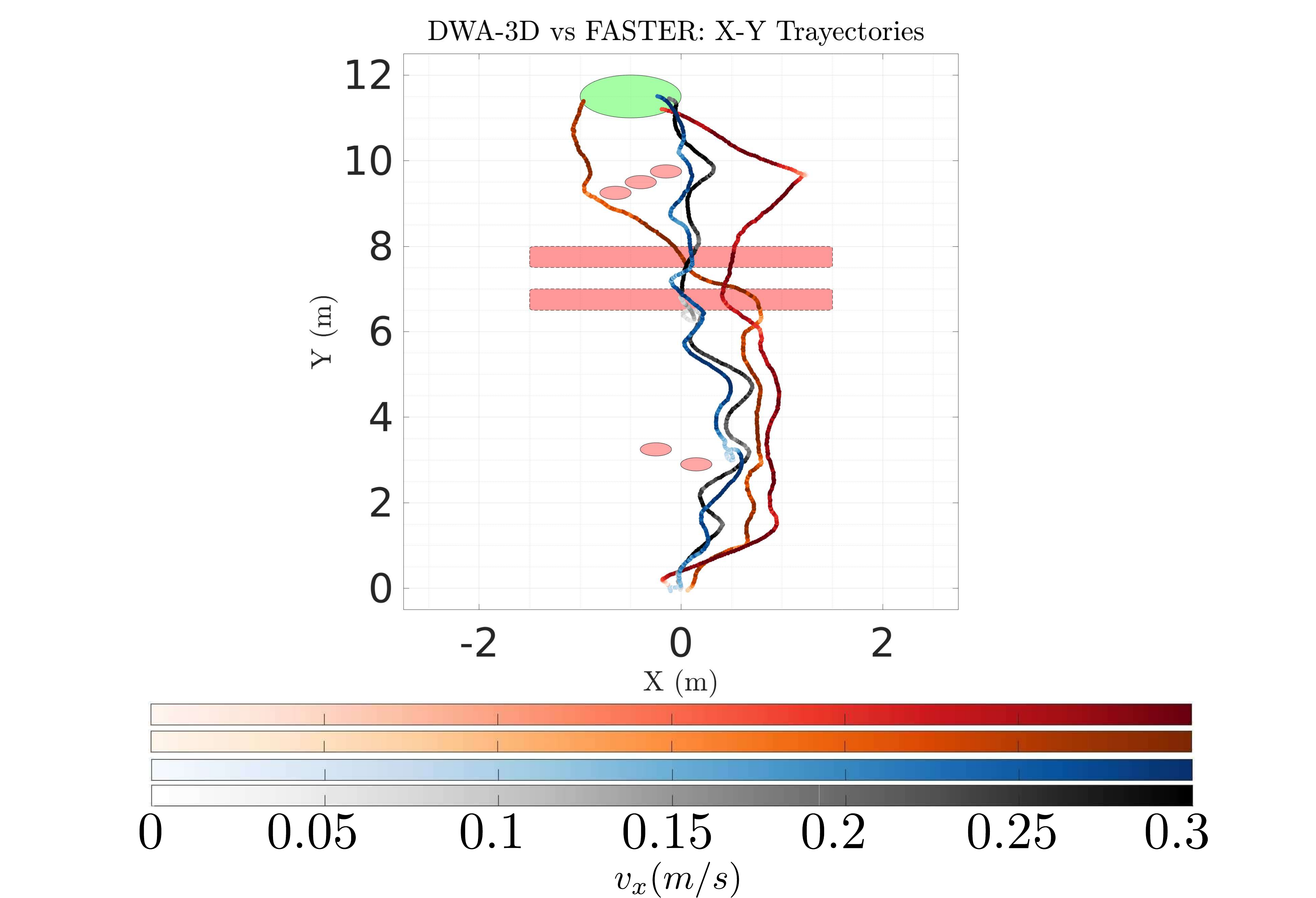}
        \caption{DWA-3D (red and orange) and FASTER (blue and black) trajectories performed during the 11 meters flights in our facilities. The darkness of each trace is related with the UAV velocity at each instant. Red shadowed regions represent the obstacles (see \autoref{fig:ComparacionNave}), while the green one is the goal.}
        \label{fig:DWA_vs_FASTER_nave_traj}
\end{figure}

\begin{figure}[h!]
    \centering
        \includegraphics[width = \linewidth, trim=5cm 0cm 5cm 0cm, clip,keepaspectratio]{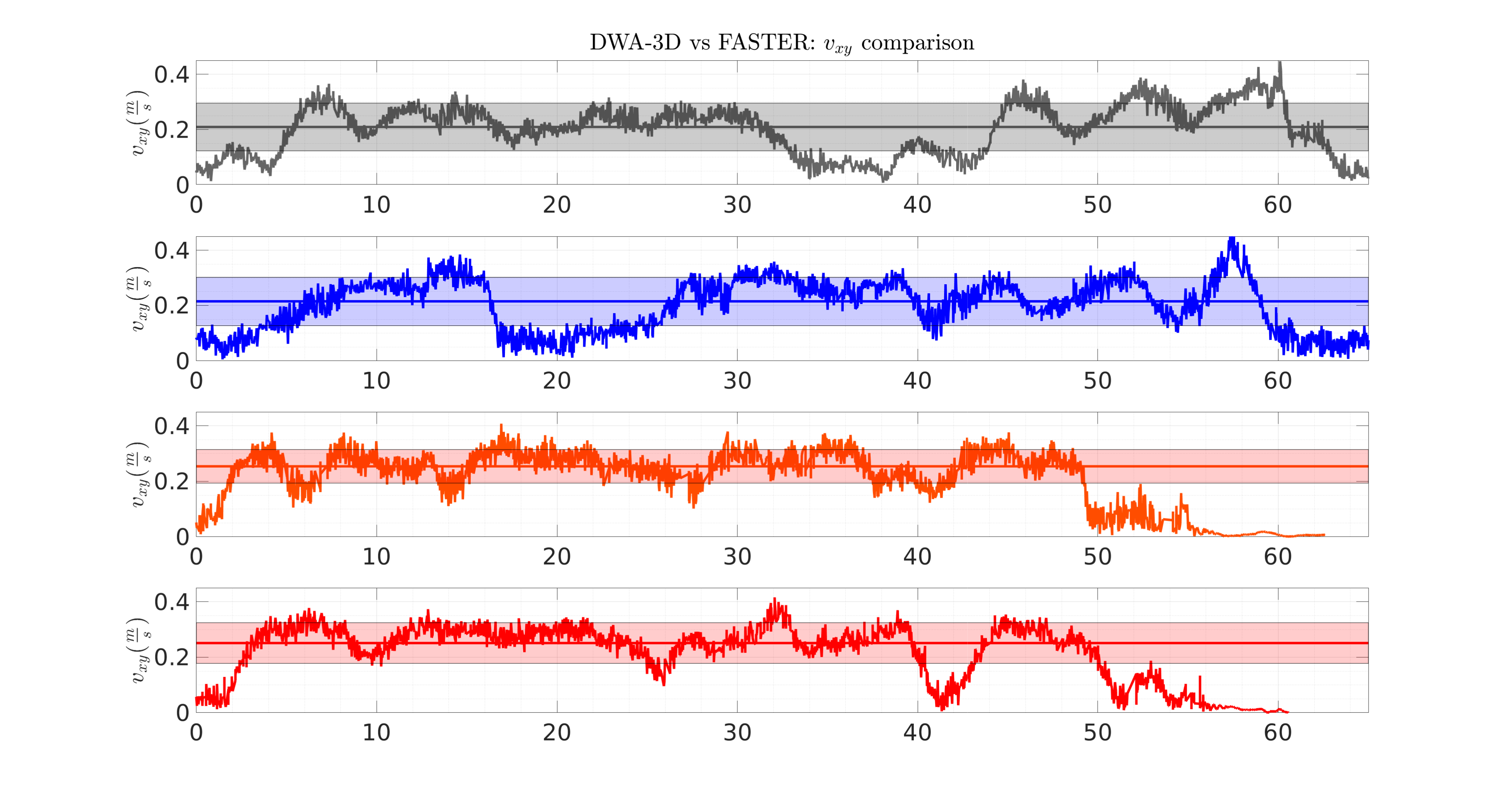}
        \caption{DWA-3D (red and orange) and FASTER (blue and black) $v_{xy}$ temporal evolution in the 11 meters flights in our facilities. Mean value for each experiment is also represented. The shadowed areas represent mean $\pm$ standard deviation. Note that DWA needs less time, has a greater mean velocity and experiments lower standard deviation. The brake maneuver in the second DWA-3D flight (red) corresponds with the avoidance of the last obstacle.}
        \label{fig:Dwa_vs_FASTER_vel_nave}
\end{figure}

In the second scenario the UAV must cross 11 meters in our facilities. The environment is populated by diverse obstacles that require both lateral and vertical avoidance maneuvers, see \autoref{fig:ComparacionNave}. The trajectories described by each planner at two consecutive experiments each one are represented in \autoref{fig:DWA_vs_FASTER_nave_traj}, where the darkness of each trace depends on the velocity of the UAV. \autoref{tab:dwa_vs_faster_nave_tab} resumes the performance of each flight. Again, DWA-3D needed to travel less time and  distance, while maintaining a higher mean velocity. Despite the larger maneuver to avoid the final obstacle, the second DWA-3D flight resulted in both a lower flight time and a lower total traveled distance compared to FASTER. This is because DWA-3D maintained a nearly straight trajectory for most of the flight, while FASTER's trajectory exhibited significant oscillations. The variations in DWA-3D's flights are attributed to the random solutions generated by RRT*. This is most noticeable when encountering the final obstacle, where one trajectory avoids it to the left and the other to the right.  The observations about the experiments in this scenario are reinforced by $v_{xy}$ temporal evolution, shown in \autoref{fig:Dwa_vs_FASTER_vel_nave}.

\begin{table}[h!]
    \centering
    \begin{tabular}{|*{4}{c|}}
        \hline
        \textbf{Flight} & \textbf{T(s)} & \textbf{Dist.(m)} & \textbf{Mean Vel.(m/s)} \\
        \hline
        {\color{gray}FASTER 1} & 61s & 13.2m & 0.22 m/s \\
        \hline
        {\color{blue}FASTER 2} & 59s & 13.1m & 0.22 m/s \\
        \hline
        {\color{orange}DWA-3D 1} & 49s & 12.5m & 0.26 m/s \\
        \hline
        {\color{red}DWA-3D 2} & 50s & 12.7m & 0.25 m/s \\
        \hline
    \end{tabular}
    \caption{DWA-3D and FASTER results from the 11 meters flights in our facilities.}
    \label{tab:dwa_vs_faster_nave_tab}
\end{table}

Our approach, DWA-3D, outperformed FASTER in terms of traveled distance, flight time, and mean velocity in both scenarios. For inspection tasks, DWA-3D exhibited smoother velocities profiles, crucial for accurate recording of the regions of interest. Furthermore, this stable velocity translates to fewer acceleration and braking maneuvers up to stopping several times, extending UAV autonomy. Finally, we provided intuitive guidelines for configuring our planner's parameters (\autoref{sec:ParamsConfig} and \hyperref[sec:AppendixA]{Appendix A}), contrasting with FASTER's absence of instructions and the complexity of its tunning. For instance, it involves setting up the polyhedras used for planning and the time allocation parameters.

\subsection{Discussion}
\label{sec:discussion}
Apart from the maximum velocities and acceleration dependent on the specific drone, a few set of parameters have to be tuned to change the desired behavior related with obstacle avoidance. Lower $r_{search}$ values allow passing through narrower gaps in exchange of assuming higher risks while avoiding obstacles, as minor safety distances are kept (frontal, vertical and lateral). The difference is even more noticeable when obstacles that were not observed during global planning are dodged. RRT* global planner may retrieve sub-optimal plans when the path traverses partially or fully unknown areas. 
Global and local planner integration along with providing UAV size awareness to the global planner allows performing smoother and more direct trajectories; combining their virtues and compensating their weaknesses. 

\begin{figure}[h!]
    \centering
    \begin{subfigure}{\columnwidth}
        \centering
        \includegraphics[width = \linewidth, keepaspectratio, trim=4.5cm 0cm 0cm 0cm, clip]{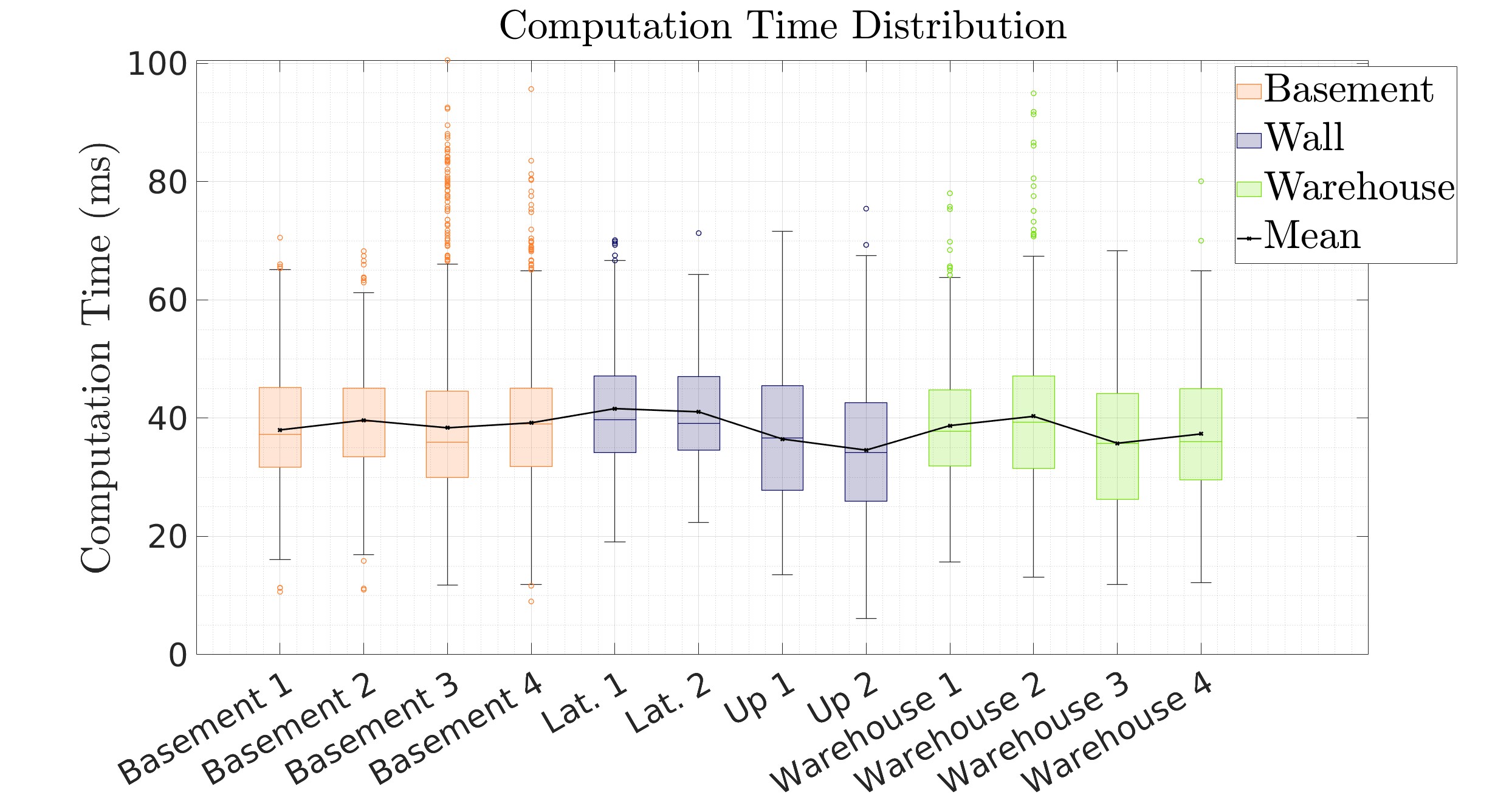}
        \caption{Computation time distribution for each individual flight.}
    \end{subfigure}
    \begin{subfigure}{\columnwidth}
        \centering
        \includegraphics[width = \linewidth, keepaspectratio, trim=4.5cm 0cm 0cm 0cm, clip]{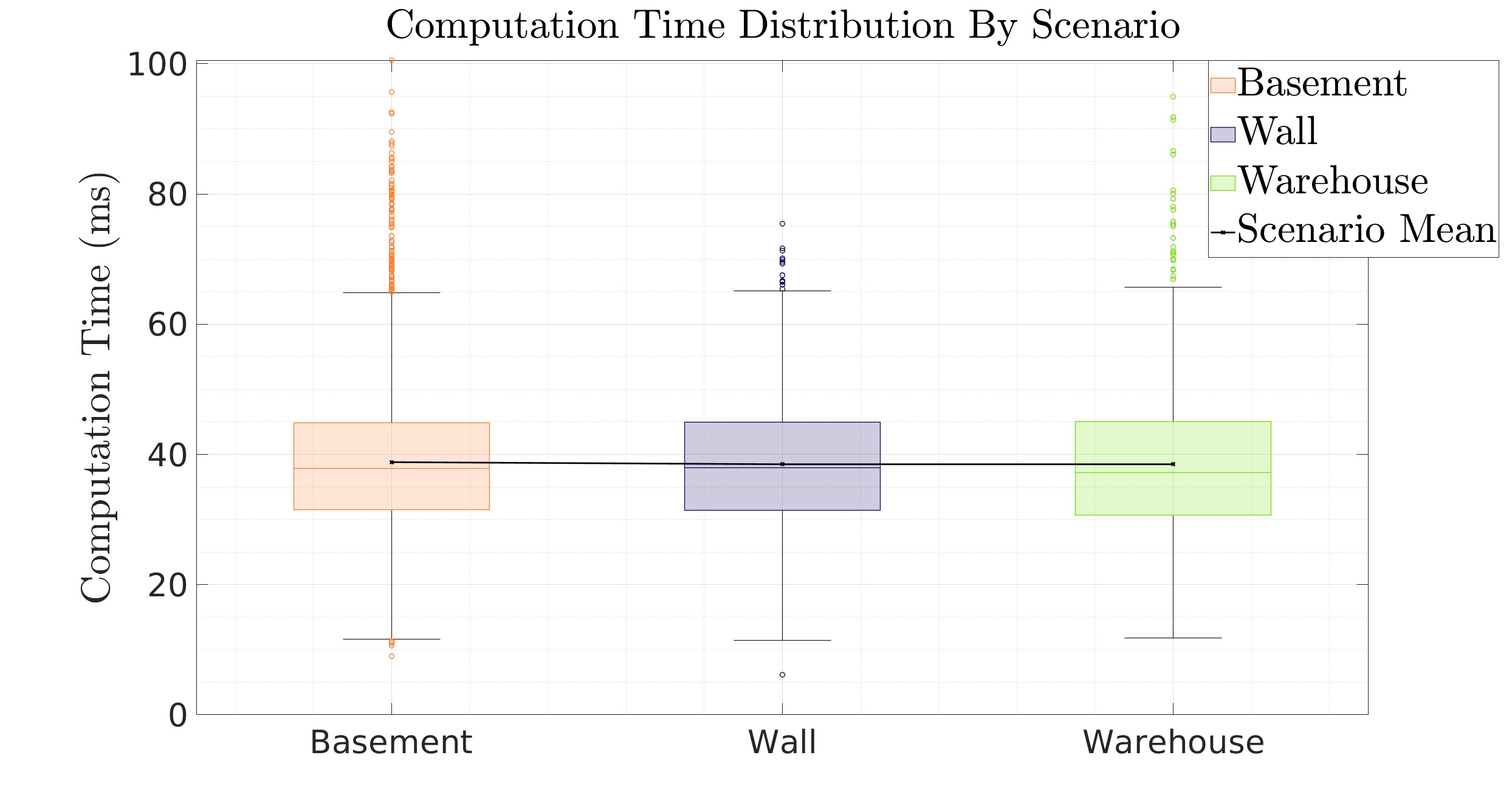}
        \caption{Computation time distribution grouped by scenario.}
    \end{subfigure}
    \caption{DWA computation time distribution in 3 scenarios (\textit{Basement}, \textit{Wall} and \textit{Warehouse}) for 4 consecutive flights in each one.}
    \label{fig:BoxCompTimes}
\end{figure}

A wide number of experiments have been performed, using the three planners, and different values of the weighting parameters 
in the objective function in DWA-3D local planner. From those experiments, we concluded that the values shown in \autoref{tab:DWA_parameters}, are the ones that best adapt to all the encountered situations, moreover having an intuitive meaning. Therefore it is very easy to tune or change to modify the desired avoidance behavior.

Regarding DWA-3D computation time, it has been measured in \textit{Basement}, \textit{Wall} and \textit{Warehouse} scenarios during the execution of four consecutive flights each, see \autoref{fig:BoxCompTimes}. It does not only remain bounded under $100 ms$, but also it has small variability, being stable about $40 ms$ (both mean and median) in every flight and scenario.

The code used for this research will be made available upon publication acceptance \url{https://github.com/JBesCar/Dwa3D}. This will enable researchers to reproduce the results and facilitate further development in this field.

\section{Conclusions and future work}\label{sec:conclusions}
In this work a new DWA-3D local planning algorithm has been developed for UAV navigation to obtain great maneuverability in narrow rooms among obstacles. It has been integrated along with a RRT* global planner. 
Avoidance preference can be adjusted to lateral or vertical evasion of the obstacles by balancing the values of two parameters $K_{\psi}$ and $K_z$ in the objective function optimization. The constraints for the parameters have been analytically analyzed, and a set of parameter values for proper configuration have been provided along with a well-tested and configurable code in ROS.  Real-world experimentation has been performed using a custom-built hexarotor equipped with a 3D LiDAR. Different scenarios have been built,  successfully navigating  between narrow gaps and in a lightly changing scenario. 

The navigation performance in the case of moving obstacles is limited by the Octomap refresh rate. Thus, a faster local map representation would be desirable to navigate in more complex and dynamic scenarios. Obviously this strongly affect to the global system performance, but not to the navigation method itself, the focus of this work.

During the experiments in the Somport Tunnel, one of the lessons learned was that dust was detected by the 3D LiDAR. Although it did not greatly affect the drone localization, it led to sparse small occupied areas in the Octomap that could obstruct sometimes its movement. 3D pointcloud filtering techniques must be considered when facing dusty environments like this in the future. Furthermore, certain regions of the tunnel lacked detail, making F-LOAM fail in detecting forward motion, so a more robust localization approach would be needed.  


Following with the work presented in this paper, an ongoing work we are achieving is centered in the inspection and monitoring tasks in tunnels and galleries, integrating exploration planners and heterogeneous robots, UGV and UAV, collaboration.   Although global or exploration planners have to be developed, a robust navigation technique as the developed here is needed to make the final maneuvering decisions against disturbances or bad perception due to external causes. In the experiments shown, it has been revealed that the technique is able to overcome these challenges.

\section*{Appendix A: Parameter Equations Development}
\label{sec:AppendixA}
An initial orientative value for $\alpha$, $\beta$ and $\gamma$ can be established by following a reasoned mathematical approach starting from \autoref{eq:G_expanded}. Given that the UAV is perfectly aligned both in height and orientation, it will score $1$ in the normalized \textit{Heading} term, so 

\begin{equation}
    \ G(\textbf{v}) = \alpha + \beta \cdot Dist(\textbf{v}) + \gamma \cdot Vel(\textbf{v})
\end{equation}
If we dismiss the \textit{Velocity} term, which could be feasible under certain circumstances,
\begin{equation}
    \ G(\textbf{v}) = \alpha + \beta \cdot Dist(\textbf{v})
\end{equation}
In order to be able to avoid an obstacle that appears in front of the UAV; a velocity that pulls the drone away from the path and the obstacle must be selected, thus
\begin{equation}
    \beta \cdot Dist(\textbf{v}) > \alpha
\end{equation}
So, if we want to be sure that the normalized \textit{Distance} term scores $1$ always it is possible, which ensures staying away from obstacles, 
\begin{equation}
    \beta > \alpha
    \label{eq:b>alpha}
\end{equation}
Note that motion direction could need to change even $180^{\circ}$ with respect to the goal to avoid certain dangers, this means even in the worst case for the $Head_{psi}$ term, the avoidance must be preferred if following the plan would lead to collision, thus considering that we are moving in a horizontal plane and combining \autoref{eq:Lij}, \autoref{eq:rho_dist} and \autoref{eq:Dist_Normalized}:
\begin{subequations}
    \begin{equation}
    \label{eq:4_5a}
        \beta \frac{r_{search} \cdot \rho_{\psi}^{90^{\circ}} \cdot \rho_{\theta}^{0^{\circ}}}{r_{search}} + \alpha \cdot \frac{\omega_{z}^{max} \cdot \Delta t}{\pi} < \beta
    \end{equation}
    \begin{equation}
            \beta \cdot (1 - \lambda_{\psi}) + \alpha \cdot \frac{\omega_{z}^{max} \cdot \Delta t}{\pi} < \beta
    \end{equation}
        \begin{equation}
            \beta \cdot \lambda_{\psi} > \alpha \cdot \frac{\omega_{z}^{max} \cdot \Delta t}{\pi}
            \label{eq:b*lambda>alpha}
    \end{equation}
\end{subequations}

that means that it must be preferred a velocity that avoids being penalized by the shortest horizontal ray (right side of \autoref{eq:4_5a}) rather than one that aligns the drone by an angle of $\omega_{z}^{max} \cdot \Delta t$ with the goal while touching the obstacle with that aforementioned ray (left side of \autoref{eq:4_5a}). Note that $R_{drone}$ has not been taken into account for the sake of simplicity, the committed error is accepted because we are computing a tentative value.  
Regarding $\gamma$ value, it must be considered that higher speeds are desirable always that they do not involve a greater risk of collision nor worsen the preferred alignment:
\begin{equation}
    \ \beta > \gamma
    \label{eq:b>gamma}
\end{equation}
\begin{equation}
    \ \alpha \cdot max(K_z, K_{\psi}) > \gamma
    \label{eq:alpha>gamma}
\end{equation}

The values in \autoref{tab:DWA_parameters}  fit these constraints, which were considered in the adjustment process.

Regarding maximum accelerations, both rotors thrust capabilities ($T_i$) and system mass ($m$) must be considered. From \cite{moussid2015dynamic}, $\dot{v}_x$ and $\dot{v}_z$ can be expressed as:

\begin{subequations}
    \begin{equation}
    \dot{v}_x = -\frac{K_{ftx}}{m} v_x + \frac{1}{m}u_{x}\sum^{6}_{i=1} T_i
    \end{equation}
    \begin{equation}
    u_x = cos(\phi) cos(\psi) sin(\theta) + sin(\phi) sin(\psi),
    \end{equation}
\end{subequations}
\begin{equation}
 \ \dot{v}_z = -\frac{K_{ftz}}{m} v_z - g + \frac{cos(\phi) cos(\theta)}{m}\sum^{6}_{i=1} T_i
\end{equation}
where $K_{ftx}$ and $K_{ftz}$ are air resistance coefficients and $g$ the gravity. To compute $\dot{v}_x^{max}$ we can impose that $\phi = 0$, $\psi = 0$ and dismiss $K_{ftx}$ and $K_{ftz}$,

\begin{equation}
\label{eq:ax}
    \ \dot{v}_x^{max} = \frac{\sum^{6}_{i=1} T_i sin(\theta)}{m}
\end{equation}
\begin{equation}
\label{eq:az}
    \ \dot{v}_z^{max} = -g + \frac{\sum^{6}_{i=1} T_i cos(\theta)}{m}
\end{equation}

If we suppose that $\dot{v}_z = 0$ and limit the maximum pitch to $\theta = 25 ^{\circ}$, the throttle that each motor will be,
\begin{equation}
    \ T_i (\theta = 25^{\circ}) = \frac{g \cdot m}{6 cos(\theta)} = 6.57 N
\end{equation}
according to the motors' manufacturer this thrust is achieved with around $50 \%$ of throttle, which ensures a safe working point. Note that all motors are considered to be performing the same thrust because only forward motion is considered here ($\omega_z = 0$).
Thus, the $\dot{v}_x^{max}$ computation is straightforward,
\vspace{-3mm}
\begin{equation}
  \ \dot{v}_x^{max} (\theta = 25 ^{\circ}) = \frac{6 \cdot T_i \cdot sin(\theta)}{m} = 4.5 \frac{m}{s^2}
\end{equation}
so the restriction imposed because of the lab size fulfills this limit. 

Under the previous scenario ($\theta \leq 25 ^{\circ}$), supposing the UAV is accelerating with $\dot{v}_x = \dot{v}_z = 1 \frac{m}{s^2}$, and working with \autoref{eq:ax} and \autoref{eq:az} the pitch angle will be, 
\vspace{-3mm}
\begin{equation}
    \ \theta(\dot{v}_x = \dot{v}_z = 1 m/s^2) = atan\bigg(\frac{1}{1 + g}\bigg) = 5.3 ^{\circ}
\end{equation}
 In this situation it is required a thrust of 
 \begin{equation}
    \ T_i (\theta = 5.3^{\circ}) = \frac{m}{6 sin(\theta)} = 6.58 N
\end{equation}
Then, the imposed acceleration limits are achievable with the available hardware.

\section*{Declaration of generative AI and AI-assisted technologies in the writing process.}
During the preparation of this work the authors used Gemini in order to improve text readability. After using this tool, the authors reviewed and edited the content as needed and take full responsibility for the content of the published article.

\section*{Acknowledgments}
This work was partially supported by the Spanish projects PID2022-139615OB-I00/MCIN/AEI/10.13039/501100011033/FEDER-UE, Investigo-111-68-D and DGA-T45-23R.

\bibliographystyle{elsarticle-num}
\balance
\bibliography{bibliography}

\end{document}